\documentclass[preprint,12pt]{elsarticle}

\usepackage{amssymb}
\usepackage{amsmath}
\usepackage{booktabs}
\usepackage{array}
\usepackage{multirow}
\usepackage{manyfoot}
\usepackage{pdflscape}
\usepackage{rotating}
\usepackage{threeparttable}
\usepackage[utf8]{inputenc}
\usepackage[colorlinks=true,      % false: boxed links; true: colored links
    linkcolor=black,      % color of internal links (change box color with linkbordercolor)
    citecolor=black,      % color of links to bibliography
    filecolor=black,      % color of file links
    urlcolor=black]{hyperref}

\usepackage[table]{xcolor}
\usepackage{longtable}
\usepackage{adjustbox}
\usepackage{nomencl}
\usepackage{framed}
\usepackage{float}
\usepackage{afterpage}
\usepackage{soul}
\usepackage{tikz}
\usepackage{colortbl}
\usepackage{makecell}
\usetikzlibrary{shapes.geometric, arrows, positioning, fit}
\usepackage{lscape}
\usepackage{geometry}
\usepackage{tabularx}

\definecolor{lightblue}{RGB}{173,216,230}
\definecolor{skyblue}{RGB}{135,206,235}
\definecolor{lightgreen}{RGB}{144,238,144}
\definecolor{lightyellow}{RGB}{255,255,224}
\definecolor{lightorange}{RGB}{255,228,196}
\definecolor{lightpurple}{RGB}{230,230,250}
\definecolor{lightpink}{RGB}{255,182,193}
\definecolor{lightcyan}{RGB}{224,255,255}

\definecolor{deteriorateColor}{HTML}{d8b365}
\definecolor{neutralColor}{HTML}{f5f5f5}
\definecolor{improveColor}{HTML}{5ab4ac}

\newcommand{\cellcolorgrad}[1]{%
  \ifnum#1>20
    \cellcolor{improveColor!100}%
  \else\ifnum#1>15
    \cellcolor{improveColor!75}%
  \else\ifnum#1>10
    \cellcolor{improveColor!50}%
  \else\ifnum#1>5
    \cellcolor{improveColor!25}%
  \else\ifnum#1>0
    \cellcolor{improveColor!10}%
  \else\ifnum#1<-20
    \cellcolor{deteriorateColor!100}%
  \else\ifnum#1<-15
    \cellcolor{deteriorateColor!75}%
  \else\ifnum#1<-10
    \cellcolor{deteriorateColor!50}%
  \else\ifnum#1<-5
    \cellcolor{deteriorateColor!25}%
  \else\ifnum#1<0
    \cellcolor{deteriorateColor!10}%
  \else
    \cellcolor{neutralColor}%
  \fi\fi\fi\fi\fi\fi\fi\fi\fi\fi
}

\newcommand\su[1]{\textbf{\textcolor{purple}{SR: #1}}}

\makenomenclature

\makeatletter
\def\ps@pprintTitle{%
 \let\@oddhead\@empty
 \let\@evenhead\@empty
 \def\@oddfoot{\reset@font\hfil}
 \let\@evenfoot\@oddfoot}
\makeatother

\begin{document}

                                        % Frontmatter
\newpage

\begin{frontmatter}
    %% Title, authors and addresses

                %% use the tnoteref command within \title for footnotes;
                %% use the tnotetext command for theassociated footnote;
                %% use the fnref command within \author or \affiliation for footnotes;
                %% use the fntext command for theassociated footnote;
                %% use the corref command within \author for corresponding author footnotes;
                %% use the cortext command for theassociated footnote;
                %% use the ead command for the email address,
                %% and the form \ead[url] for the home page:

\title{Transfer Learning on Transformers for Building Energy Consumption Forecasting - A Comparative Study\tnoteref{label1}}
\tnotetext[label1]{This research was supported by Massey University.}

\author[mc]{Robert Spencer}
\author[mc]{Surangika Ranathunga\corref{cor1}}
\ead{S.Ranathunga@massey.ac.nz}
\author[be]{Mikael Boulic}
\author[be]{Andries (Hennie) van Heerden}
\author[mc]{Teo Susnjak}

\cortext[cor1]{Corresponding author}
\affiliation[mc]{organization={School of Mathematical and Computational Sciences}}

\affiliation[be]{organization={School of Built Environment, Massey University}, 
           % addressline={Albany}, 
            city={Auckland}, 
            postcode={0632}, 
            %state={Auckland}, 
            country={New Zealand}}

                                        %% Abstract
\sloppy
\begin{abstract}
%% Text of abstract
This study investigates the application of Transfer Learning (TL) on Transformer architectures to enhance building energy consumption forecasting. Transformers are a relatively new deep learning architecture, which has served as the foundation for groundbreaking technologies such as ChatGPT. While TL has been studied in the past, prior studies considered either one data-centric TL strategy or used older deep learning models such as Recurrent Neural Networks or Convolutional Neural Networks. Here, we carry out an extensive empirical study on six different data-centric TL strategies and analyse their performance under varying feature spaces. In addition to the vanilla Transformer architecture, we also experiment with Informer and PatchTST, specifically designed for time series forecasting. We use 16 datasets from the Building Data Genome Project 2 to create building energy consumption forecasting models.  Experimental results reveal that while TL is generally beneficial, especially when the target domain has no data, careful selection of the exact TL strategy should be made to gain the maximum benefit. This decision largely depends on the feature space properties such as the recorded ``weather features''. We also note that PatchTST outperforms the other two Transformer variants (vanilla Transformer and Informer). Our findings advance the building energy consumption forecasting using advanced approaches like TL and Transformer architectures.  %Amidst growing environmental concerns and the pressing need for optimised energy management in buildings, traditional forecasting methods often fall short due to their inability to handle large, diverse datasets effectively. Our approach leverages the robust performance of Transformers in sequence modelling tasks to improve predictive accuracy and model adaptability across various building types and climatic conditions. Utilising the comprehensive Building Data Genome Project 2 dataset, which includes diverse energy consumption profiles across geographies and building characteristics, we trained and fine-tuned 996 Transformer-based models. The results demonstrate significant improvements in forecasting accuracy compared to traditional models, showcasing the potential of TL to enhance model precision and address challenges posed by limited data scenarios commonly encountered in building energy forecasting. These findings have important practical implications, suggesting pathways for integrating AI-driven tools into building management systems to achieve greater energy efficiency and sustainability. This study lays the groundwork for future explorations into scalable, efficient AI solutions for real-world energy management challenges, aligning with global sustainability goals.
\end{abstract}
\fussy

%%Graphical abstract NOT MANDATORY
%%\begin{graphicalabstract}
%%\includegraphics{grabs}
%%\end{graphicalabstract}

                                        %% Research highlights MANDATORY
\sloppy

\fussy

%% Keywords
\begin{keyword}
Neural Networks \sep Time Series Forecasting \sep Machine Learning \sep Deep Learning \sep Model Fine-Tuning \sep Building Energy Planning Models
\end{keyword}

\end{frontmatter}

                                        % INTRODUCTION
\begin{framed}                                 
\printnomenclature  
\nomenclature{TL}{Transfer Learning}
\nomenclature{DL}{Deep Learning}
\nomenclature{ML}{Machine Learning}
\nomenclature{BDGP2}{Building Data Genome Project 2}
\nomenclature{ASHRAE}{American Society of Heating, Refrigerating and Air-Conditioning Engineers}
\nomenclature{MAE}{Mean Absolute Error}
\nomenclature{MSE}{Mean Squared Error}
\nomenclature{RNN}{Recurrent Neural Networks}
\nomenclature{LSTM}{Long-short Term Memory}
\nomenclature{GRU}{Gated Recurrent Unit}
\nomenclature{DANN}{Domain Adversarial Neural Network}
\nomenclature{CNN}{Convolutional Neural Network}
\nomenclature{MLP}{Multi-layer Perceptron}
\nomenclature{FF}{Feed Forward}
\nomenclature{BPNN}{Back-Propagation Neural Network}
\nomenclature{TFT}{Temporal Fusion Transformer}
\nomenclature{ELM}{Extreme Learning Machine}
\nomenclature{FEDformer}{Frequency Enhanced Decomposed Transform}
\nomenclature{PatchTST}{Patch Time Series Transformer}
\end{framed}

\section{Introduction}\label{sec1}
\subsection{Background}
Driven by the need to mitigate the impact of climate change, there is a global focus on reducing carbon emissions in the building sector~\cite{en16093748, 10.1007/978-3-319-94965-9_17}. Previous studies have emphasised the significant role of the construction industry in energy usage, with buildings accounting for around one-third of global greenhouse gas emissions \cite{CHRISTOPHER2023136942}. %Moreover, buildings account for approximately 40\% of the total energy consumption worldwide \cite{KHALEDMOHAMMAD2023389}. 
Starting from 2012, there has been a consistent annual increase of 1.5\% in energy consumption within buildings in nations belonging to the Organisation for Economic Cooperation and Development (OECD) such as Australia, New Zealand, the United Kingdom, and the USA. In addition, nations outside the OECD have experienced a higher increase of 2.1\% in their energy consumption levels \cite{BUI2020116370}. Therefore, balancing economic viability, environmental sustainability, and occupant comfort, health, and safety is crucial for all stakeholders involved in the building sector \cite{luo2022controlling}. Being able to make accurate forecasting of a building's energy consumption is a crucial requirement in this context.

Approaches for building energy consumption forecasting are typically categorised into three primary groups: 1) engineering calculations, 2) numerical simulations, and 3) data-driven modelling \cite{BUI2020116370}. The first two methods rely heavily on physical laws or physics-based simulations to estimate energy usage. Engineering calculations are best suited for quick initial evaluations, offering straightforward estimations based on standard formulas. Numerical simulations, on the other hand, provide a more detailed analysis by modelling the complex interactions within buildings but demand significant computational resources and time, particularly as the complexity of a project increases \cite{BUI2020116370}.

The evolution of data-driven modelling and its rise in usage for forecasting building energy consumption and assessments is due to a significant shift away from traditional statistical models such as ARIMA and SARIMA to advanced Machine Learning (ML) techniques \cite{DEB2017902, GASSAR2020110238}. ML methods, notably Deep Learning (DL) techniques, are better suited for the dynamic and intricate energy usage patterns, offering significant improvements in forecasting accuracy and applicability across different temporal and spatial scales in energy planning models \cite{DEBNATH2018297}. 

Recurrent Neural Networks (RNNs) and their advanced variants Long-Short Memory Networks (LSTMs) and Gated Recurrent Units (GRUs) have been the most commonly used DL architectures for building energy forecasting. Some researchers have also used Convolutional Neural Networks (CNNs)~\cite{chandrasekaran2024advances}. However, both RNNs and CNNs have their drawbacks with respect to time series forecasting~\cite{wang2019multiple}. As an alternative, Transformer models, a sophisticated type of neural network architecture, have emerged as exceptionally potent in processing complex data sequences \cite{NIPS2017_3f5ee243}. Renowned for their effectiveness in domains such as Natural Language Processing~\cite{devlin2019bert} and Computer Vision~\cite{dosovitskiy2020image}, Transformers can be used for interpreting the intricate interrelations in time series data that affect building energy usage. While the basic Transformer architecture (which we term the \textit{vanilla Transformer})  has shown to be very effective~\cite{hertel2023transformer}, several recent research has already demonstrated that advanced Transformer variants such as Informer~\cite{zhou2021informer}, PatchTST~\cite{nie2022time} and Temporal Fusion Transformers~\cite{lim2021temporal} have already outperformed their older DL counterparts~\cite{hertel2023transformer,ni2024study}.  

Despite their promise, all DL techniques - CNNs, RNNs and Transformers alike, have one major limitation - they rely on vast amounts of training data to make accurate forecasting. However, such large amounts of energy consumption data may not exist for some buildings, due to practical reasons such as the building being newly constructed, not having the facilities to record energy consumption data in a timely manner, privacy concerns, data collection costs, data ownership, and the sheer diversity of buildings \cite{10.1007/978-981-10-0557-2_90, GUERRASANTIN2015189}. On the other hand, there are multiple publicly available datasets that contain energy consumption data from different buildings around the world, such as the  Building Data Genome Project 2 (BDGP2)~\cite{Miller_2020, doi:10.1080/14786451.2015.1100196, rodrigues2018load}. 

\begin{figure}[H]
    \centering
    \includegraphics[width=1\linewidth]{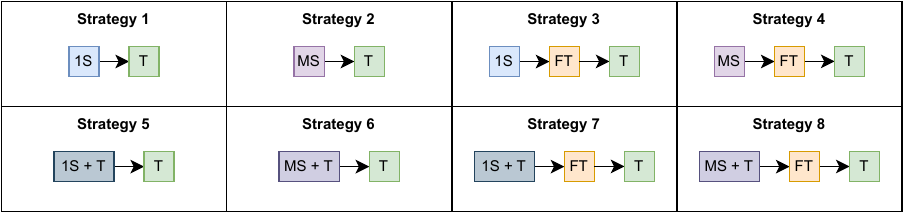}
    \caption{Data-centric TL strategies. 1S- one-source, MS- multi-source, FT- fine-tune with target, T- target. Green coloured box indicates the testing phase.}
    \label{fig:Data-centric TL strategies}
\end{figure}

Transfer Learning (TL) can be broadly defined as `the ability of a system to 
recognise and apply knowledge and skills learned in
previous tasks to novel tasks'~\cite{pan2009survey}. In other words, TL utilises insights derived from a more documented dataset (the source) to bolster the predictive accuracy of models applied to a new, data-sparse context (the target). TL is an excellent way to make use of these existing datasets to build energy consumption models for buildings that have limited or no data of their own. Our survey of existing literature  revealed that the TL techniques used in the context of building energy consumption forecasting can be categorised based on how the \textit{source} and \textit{target} datasets are being used, as shown in Figure~\ref{fig:Data-centric TL strategies} and outlined below\footnote{Note that the first two data-centric TL strategies above can be identified as the zero-shot setup, as the model does not see any target data during training time.}. We term these \textit{data-centric TL strategies}.

\begin{itemize}
    \item \textbf{Strategy 1:} Train a model with one source (S) $\rightarrow$ test with the target (T) - (1S $\rightarrow$ T)
    \item \textbf{Strategy 2:} Train a model with multiple sources (MS) $\rightarrow$  test with the target - (MS $\rightarrow$ T)
    \item \textbf{Strategy 3:}Train a model with one source $\rightarrow$  further train with the target (FT)$\rightarrow$ test with the target - (1S $\rightarrow$ FT $\rightarrow$ T)
    \item \textbf{Strategy 4:} Train a model with multiple sources $\rightarrow$ further train with the target $\rightarrow$ test with the target - (MS $\rightarrow$ FT $\rightarrow$ T)
    \item \textbf{Strategy 5:} Train a model with one source and target $\rightarrow$ test with the target - (1S + T $\rightarrow$ T)
    \item \textbf{Strategy 6:} Train a model with multiple sources and target $\rightarrow$ test with the target - (MS + T $\rightarrow$ T)
    \item \textbf{Strategy 7:} Train a model with one source and target $\rightarrow$ further train with the target $\rightarrow$ test with the target -  (1S + T $\rightarrow$ FT $\rightarrow$ T)
    \item \textbf{Strategy 8:} Train a model with multiple sources and target $\rightarrow$ further train with target $\rightarrow$ test with the target - (MS + T $\rightarrow$ FT $\rightarrow$ T)
\end{itemize}

\subsection{Problem and Motivation}
The application of Transformers to building energy consumption forecasting is new, with only a few known studies employing TL techniques on this architecture~\cite{10.1145/3600100.3626635, LOPEZSANTOS2023113164}. These studies, similar to most TL-based research in this domain using other DL methods (e.g., LSTM, CNN), explored only a single data-centric TL strategy (see Table~\ref{table:tl_techniques_summary}). The work by ~\citet{electronics13101996} stands out for investigating four data-centric TL strategies, but their experiment was limited in scale, involving only three buildings, and they did not consider the Transformer architecture. Similarly,~\citet{LU2023109024} experimented with two Strategies on the LSTM model. Consequently, the potential of combining various TL techniques with Transformers for building energy consumption forecasting remains largely unexplored.

\subsection{Aims of the Study}
This study aims to comprehensively investigate the effectiveness of various data-centric TL strategies when applied to Transformer architectures for building energy consumption forecasting. Specifically, we seek to answer the following Research Questions (RQs):

\begin{itemize}
    % \item RQ 1: What is the comparative performance of different data-centric TL strategies in the context of the vanilla Transformer architecture?
    % \item RQ 2: What dataset characteristics influence the effectiveness of different data-centric TL strategies implemented on the vanilla Transformer architecture?
    % \item RQ 3: How do different data-centric TL strategies perform on advanced Transformer architectures designed for time series prediction?
    \item RQ1: How do different data-centric TL strategies compare in performance when applied to the vanilla Transformer architecture for building energy consumption forecasting?
    \item RQ2: What specific characteristics of building energy datasets (e.g. ambient weather features, climate zone, data volume) influence the effectiveness of different data-centric TL strategies when implemented on the vanilla Transformer architecture?
    \item RQ3: How does the performance of various data-centric TL strategies differ when applied to advanced Transformer architectures specifically designed for time series forecasting, compared to the vanilla Transformer?
  
\end{itemize}

\subsection{Contribution}
% In order to answer the three RQs listed above, we conducted a large-scale experiment involving 16 datasets from the Building Data Genome Project 2 (BDGP2) \cite{Miller_2020}. We conducted experiments for all eight data-centric transfer strategies, making this the most comprehensive study on the use of TL for building energy prediction (\rs{we only did 6 actually, 1,2,5,6,7,8}). We analysed the impact of the following features in the dataset: climate zone \su{what else?}.

% \su{TODO: describe findings}

% We also experimented with Informer, \su{what else and what did we find}

In order to answer the three RQs listed above, we conducted a large-scale experiment involving 16 datasets from BDGP2 \cite{Miller_2020}. Our study makes several key contributions to the field of building energy forecasting:

\begin{enumerate}
    \item Comprehensive analysis of TL Strategies: We conducted experiments for six out of eight data-centric transfer learning strategies (1, 2, 5, 6, 7, and 8), making this one of the most comprehensive studies on the use of TL for building energy forecasting. This broad approach allows for a nuanced understanding of how different TL techniques perform in various scenarios.

    \item Impact analysis of dataset characteristics: We analysed the impact of the following features in the dataset: climate zone, weather features, data volume and temporal range.
   % \begin{itemize}
      %  \item Climate zone: We examined how models trained on data from one climate zone perform when applied to buildings in different climate zones.
      %  \item Data volume: We studied how the amount of training data affects model performance and transferability.
      %  \item Weather features: We assessed the importance of various weather-related inputs on prediction accuracy.
      %  \item Temporal range: We explored how models perform when trained on data from different time periods.
    %\end{itemize}

    \item Large-Scale Modelling with Advanced Transformer Variants: In addition to the vanilla Transformer, we extended our experiments to include two advanced Transformer variants: Informer and PatchTST.
   % \begin{itemize}
      %  \item Informer: We implemented the Informer architecture, known for its efficiency in handling long sequences, to assess its performance in building energy forecasting.
      %  \item PatchTST: We utilised the PatchTST model, which is specifically designed for time series forecasting tasks, to compare its effectiveness against the vanilla Transformer and Informer.
  %  \end{itemize}

   % \item Extensive Model Training: We trained and evaluated a total of 996 Transformer-based models across various configurations, providing a robust foundation for our findings.

   % \item Multi-Horizon Forecasting: We assessed model performance for both 24-hour and 96-hour forecast horizons, offering insights into short-term and medium-term prediction capabilities.

  %  \item Zero-Shot Learning Evaluation: We conducted a comprehensive assessment of zero-shot learning performance, providing valuable insights into model generalisation across different buildings and climates without specific training data.
\end{enumerate}

By combining these elements, our study provides a holistic view of TL applications on the Transformer architectures for the task of building energy forecasting and offers practical insights for researchers and practitioners in the field of building energy consumption forecasting.

\section{Related Work}\label{sec2}

\subsection{Transfer Learning}
Transfer Learning (TL) is an ML technique that makes use of the knowledge acquired from a \textit{source} model that has been trained with sufficient datasets to improve the performance of a new \textit{target} model, where there is insufficient or no data. The conceptual backbone of TL is structured around two key elements: domains and tasks. A domain is defined by a feature space $\mathcal X$ and a corresponding marginal probability distribution $P(X)$, where $X = \{x_1, x_2, \ldots, x_n\} \in \mathcal X$. 
A task consists of a label space $ \mathcal Y$ and an objective predictive function $f(\cdot)$. Together, these define the specific regression or classification task at hand.

In the field of building energy consumption forecasting, three types of input data can be identified: ambient environmental data, historical data, and time data~\cite{AHN2022111717}. Ambient environmental data is usually obtained from the Meteorological Agency or by measuring it in the field. Examples of environment data types include outdoor temperature, humidity, and solar radiation~\cite{AHN2022111717}. Historical data refers to the building energy consumption data that has already been recorded. Time data refers to the time of the year.~\citet{huy2022short} showed that special days such as New Year and Labour Day have an impact on the prediction accuracy of TL techniques. Time data is particularly important in countries where distinct seasons are present, and some researchers refer to such data as seasonal data~\cite{RIBEIRO2018352}.   

These types of data recorded in the context of a building form the feature space $\mathcal X$ of the TL problem.~\citet{LU2021119322} argue that every building is personalised, with their thermal performance and occupant behaviour being different. Therefore it is safe to consider each building as a separate domain~\cite{8848386}. If the source and target domains have the same feature space, it is termed \textit{homogeneous}, otherwise \textit{heterogeneous}.  

Based on the nature of source and target tasks and domains, TL can be categorised into three primary types — inductive, transductive, and unsupervised \cite{5288526}. Inductive TL refers to the case where the target task is different from the source task (source and target domains can be similar or not), while transductive TL refers to the case where the source and target domains are different, but the tasks are the same. Transductive TL is also referred to as `domain adaptation'. If the label information is unknown for both domains, this is considered unsupervised TL. However, for time series forecasting tasks, the use of unsupervised TL is rather uncommon\footnote{We did not find any research that claims to use unsupervised TL for building energy consumption forecasting.}.

Approaches for TL can be categorised into 4 groups: instance-based, feature-based, parameter-based (aka model-based), and relational-based approaches. Table~\ref{table:tl_techniques_summary} shows a non-exhaustive list of past research on building energy consumption forecasting with TL and DL techniques. Most of this research can be identified as parameter-based techniques, with the exception of~\citet{FANG2021119208,LI2024122357, li2024comprehensive} and some experiments of~\citet{LI2022112461}, which are feature-based techniques.

Table~\ref{table:tl_techniques_summary} also lists the data-centric TL strategy employed by the previous research. This confirms that the past research, except~\citet{electronics13101996} and \citet{LU2023109024}, experimented with only one data-centric TL strategy. Moreover, out of the 38 papers we surveyed, 44\% used Strategy 3, followed by Strategy 4, which was used by 23\%. Strategies 6, 5, 1 and 8 were applied by 16\%, 13\%, 8\% and 5\%, respectively. No research experimented with strategies 2 and 7.

% Further training a model that has already been trained is termed as `fine-tuning'. When fine-tuning a DL model that consists of multiple layers, which layers to fine-tune is an important factor. As shown in Table \ref{table:tl_techniques_summary}, related research suggests several options. Maybe the simplest option is not to fine-tune any layers of the pre-trained model. In other words, all the layers in the pre-trained model are frozen, and the model is simply used for inferencing (i.e.~test with the target - this refers to as zero-shot learning and is useful when the target has no data for model training). Fine-tuning all the layers of the pre-trained model with target data is termed as full fine-tuning or weight initialisation. On the other hand, fine-tune only the last layer of the pre-trained model with the target data is termed as feature extraction. It is always possible to fine-tune a custom set of layers of the pre-trained model with the target data. Which layers to fine-tune is important to consider as the performance of TL depends on the similarity between the source and target domains. The more layers that are fine-tuned, the closer the model resembles the target domain. 

Further training a model that has already been trained is termed `fine-tuning'. When fine-tuning a deep learning model that consists of multiple layers, which layers to fine-tune is an important factor. As shown in Table \ref{table:tl_techniques_summary}, related research suggests several options. The simplest option is not to fine-tune any layers of the pre-trained model. In other words, all the layers in the pre-trained model are frozen, and the model is simply used for inferencing (i.e., test with the target -- this refers to zero-shot learning and is useful when the target has no data for model training). Fine-tuning all the layers of the pre-trained model with target data is termed `full fine-tuning' or `weight initialisation'. On the other hand, fine-tuning only the last layer of the pre-trained model with target data is termed `feature extraction'. It is always possible to fine-tune a custom set of layers of the pre-trained model with target data. Which layers to fine-tune is important to consider, as the performance of transfer learning depends on the similarity between the source and target domains. The more layers that are fine-tuned, the closer the model resembles the target domain.

\begin{table}[p]
\centering
\caption{Summary of TL Techniques in Building Energy Consumption Forecasting}
\vspace{0.2cm}
\label{table:tl_techniques_summary}
\fontsize{8}{10}\selectfont
\begin{adjustbox}{max width=1\textwidth,center}
\begin{tabular}{@{}p{4cm}ccc@{}}
\toprule
\textbf{Authors} & \textbf{DL} & \textbf{Strategy} & \textbf{Fine-tuning} \\
\midrule
\rowcolor{gray!10}
Ribeiro et al. \cite{RIBEIRO2018352} & MLP & Strategy 6 & N/A \\
Voß et al. \cite{VOS} & CNN & Strategy 6 & N/A \\
\rowcolor{gray!10}
Tian et al. \cite{8848386} & RNN & Strategy 3 & Full \\
Hooshmand \& Sharma \cite{10.1145/3307772.3328284} & CNN & Strategy 4  & Last FC layer \\
\rowcolor{gray!10}
Fan et al. \cite{FAN2020114499} & LSTM & Strategy 4 & Last layer, full \\
Gao et al. \cite{Gao} & CNN \& LSTM & Strategy 3 & Top dense layers \\
\rowcolor{gray!10}
Jung et al. \cite{su12166364} & FF & Strategy 4 & Full \\
Ma et al. \cite{Ma} & LSTM & Strategy 3 & Bottom layers \\
\rowcolor{gray!10}
Hu et al. \cite{10.1145/3340531.3412155} & DRN & Strategy 3 & not mentioned \\
Lee et al. \cite{9330546} & LSTM & Strategy 4 & Full \\
\rowcolor{gray!10}
Li et al. \cite{li2021development} & BPNN & Strategy 3 & Full \\
Jain et al. \cite{osti_1869802} & FF & Strategy 3 & Last two layers \\
\rowcolor{gray!10}
Fang et al. \cite{FANG2021119208} & LSTM-DANN & Strategy 5 & N/A \\
Lu et al. \cite{LU2021119322} & LSTM & Strategy 3 & Full \\
\rowcolor{gray!10}
Park et al. \cite{PARK2022109060} & LSTM, CNN & Strategy 3 & Partial \\
Peng et al. \cite{PENG2022117194} & LSTM & Strategy 3 & Last layer \\
\rowcolor{gray!10}
Ahn et al. \cite{AHN2022111717} & LSTM & Strategy 3 & Top/bottom layers \\
Li et al. \cite{LI2022112461} & LSTM & Strategy 3 & Last layer \\
\rowcolor{gray!10}
Yan et al. \cite{YAN2023108127} & bi-LSTM & Strategy 8 & Last layer \\
Kim et al. \cite{su15032340} & LSTM & Strategy 3 & No/Full/Partial \\
\rowcolor{gray!10}
Lu et al. \cite{LU2023109024} & LSTM & Str 3, 4 & Partial layers \\
Tzortzis et al. \cite{math12010019} & MLP & Strategy 4 & Full \\
\rowcolor{gray!10}
Yuan et al. \cite{YUAN2023126878} & CNN-LSTM & Strategy 1 & N/A \\
Yang et al. \cite{Yang} & LSTM & Strategy 4 & New layers only \\
\rowcolor{gray!10}
Zhou et al. \cite{ZHOU2023107618} & TAB-LSTM & Strategy 5 & N/A \\
Fang et al. \cite{FANG2023112968} & LSTM & Strategy 5 & not mentioned \\
\rowcolor{gray!10}
Gokhale et al. \cite{10.1145/3600100.3626635} & TFT & Strategy 4 & Full \\
Laitsos et al. \cite{10345954} & CNN, GRU, CNN-GRU & Strategy 1 & N/A \\
\rowcolor{gray!10}
Santos et al. \cite{LOPEZSANTOS2023113164} & TFT & Strategy 3 & Last/Partial/Full \\
Laitsos et al. \cite{electronics13101996} & MLP, CNN, ELM & Str 1,3,4,6 & Full \\
\rowcolor{gray!10}
Li et al. \cite{li2024improved} & LSTM-DANN & Strategy 6 & N/A \\
Kim et al. \cite{KIM2024123500} & LSTM & Strategy 3 & Partial \\
\rowcolor{gray!10}
Xiao et al. \cite{XIAO2024111627} & LSTM, GRU & Strategy 3 & Last layer \\
Wei et al. \cite{WEI2024122087} & LSTM & Strategy 8 & not mentioned \\
\rowcolor{gray!10}
Xing et al. \cite{XING2024123276} & LSTM, vanilla Transformer & Strategy 3 & Last layer \\
Li et al. \cite{LI2024122357} & LSTM & Strategy 5 & N/A \\
\rowcolor{gray!10}
Li et al. \cite{li2024comprehensive} & LSTM-DANN & Strategy 6 & N/A \\
Wei et al. \cite{WEI2024108713} & LSTM & Strategy 6 & N/A \\
\midrule[\heavyrulewidth]
\end{tabular}
\end{adjustbox}
\vspace{0.2cm}
\begin{minipage}{\linewidth}
\scriptsize
\textbf{Abbreviations:} 1S: One-source, MS: Multi-source, T: Target, DRN: Deep Residual Network, FF: Feedforward Neural Network, MLP: Multi-Layer Perceptron, BPNN: Back-Propagation Neural Network, TFT: Temporal Fusion Transformer, DANN: Domain Adversarial Neural Network, FC: Fully Connected, ELM: Extreme Learning Machine. N/A fine-tuning is not applicable.
\end{minipage}
\end{table}

Relatedly, various similarity measurement indexes have been used to identify source domain(s) similar to the target domain.~\citet{su12166364} used the `Pearson Correlation' for this task, while~\citet{LU2021119322} developed their own correlation measurement, which they termed `Similarity Measurement Index'.~\citet{PENG2022117194} and~\citet{li2024comprehensive} used `Dynamic Time Warping', combined with `Euclidean distance'. Other techniques include `Maximum Mean Dispersion'~\cite{ZHOU2023107618}, `Variational Mode Decomposition'~\cite{XING2024123276} and a combination of `Wasserstein Distance' and `Maximal Information Coefficient'~\cite{WEI2024122087}.

In order to further explain results variations when using different source domains, some studies evaluated the impact of source domain data volume size~\cite{li2024improved, LI2024122357}, seasonality~\cite{RIBEIRO2018352}, and climate zone~\cite{AHN2022111717}.~\citet{li2024comprehensive} investigated the impact of training data volumes, seasonal information, building type and location, while~\citet{PARK2022109060} considered seasonality and occupation level.

\subsection{Transformers \& their Application in Building Energy Consumption Forecasting}
The Transformer model (i.e., the vanilla Transformer), introduced by Vaswani et al. \cite{NIPS2017_3f5ee243} in 2017, represents a significant advancement in neural network design for processing data sequences. Unlike other neural networks that rely on recurrent or convolutional layers, the Transformer uses a self-attention mechanism to compute outputs based on the entirety of input sequences directly. This approach allows it to efficiently handle long-range dependencies within the data.

% More recently, several advanced Transformer architectures have been introduced for the general problem of time series prediction. Some of these architectures are Fedformer, Informer, Patch Time Series Transformer (Patch-TST) and Temporal Fusion Transformer (TFT).\su{add a bit more description on these architectures}

More recently, several advanced Transformer architectures have been introduced for the general problem of time series forecasting. Some of these architectures are Frequency Enhanced Decomposed Transformer (FEDformer), Informer, Patch Time Series Transformer (PatchTST), and Temporal Fusion Transformer (TFT).

FEDformer model combines frequency domain analysis with the Transformer architecture. It decomposes time series data into different frequency components and applies attention mechanisms in both time and frequency domains, potentially capturing both short-term and long-term patterns more effectively. Informer is designed to address the quadratic computational complexity of standard Transformers. It uses a ProbSparse self-attention mechanism. This allows it to handle longer sequences more efficiently, making it particularly suitable for long-term time series forecasting tasks. PatchTST architecture adapts ideas from vision Transformers to time series data. It divides the input time series into patches, similar to how image patches are used in vision Transformers. This approach can capture local patterns within patches while maintaining the ability to model long-range dependencies. TFT is specifically designed for multi-horizon forecasting with multiple related time series. It incorporates specialised components for processing static covariates, known future inputs and observed inputs, making it well-suited for complex forecasting scenarios with multiple input types.

Out of these, TFT, which is relatively older has been experimented with by several researchers for the task of energy consumption forecasting~\cite{giacomazzi2023short,huy2022short,ye2024short,saadipour2023deep,ji2023multi,ni2024study}, and one has used FEDformer~\cite{liu2022short}.~\citet{hertel2022evaluation} evaluated multiple time series Transformer architectures: vanilla Transformers, convolutional self-attention Transformer, and Informer, and reported that the latter two are superior.
The TL research that used Transformer models has primarily focused on the TFT architecture~\cite{10.1145/3600100.3626635, LOPEZSANTOS2023113164}. Gokhale et al.~\cite{10.1145/3600100.3626635} applied TFT with TL for demand forecasting in home energy management systems, while Santos et al.~\cite{LOPEZSANTOS2023113164} explored various fine-tuning strategies using TFT for short-term load forecasting in data-poor buildings within local energy communities. These studies demonstrate the growing interest in leveraging advanced Transformer architectures.

% Out of these, the TFT has been experimented by several researchers for the task of energy consumption prediction~\cite{giacomazzi2023short,huy2022short,ye2024short,saadipour2023deep,ji2023multi,ni2024study}, and some have used FEDformer~\cite{liu2022short}.~\citet{hertel2022evaluation} evaluated multiple time series Transformer architectures: vanilla Transformers, convolutional self-attention Transformer and Informer, and reported that latter two are superior. The TL research that used Transformer models are all based on TFT~\cite{10.1145/3600100.3626635, LOPEZSANTOS2023113164}\rs{could use clarification}
\section{Methodology}\label{sec3}
\begin{figure}
    \centering
    \includegraphics[width=0.7\linewidth]{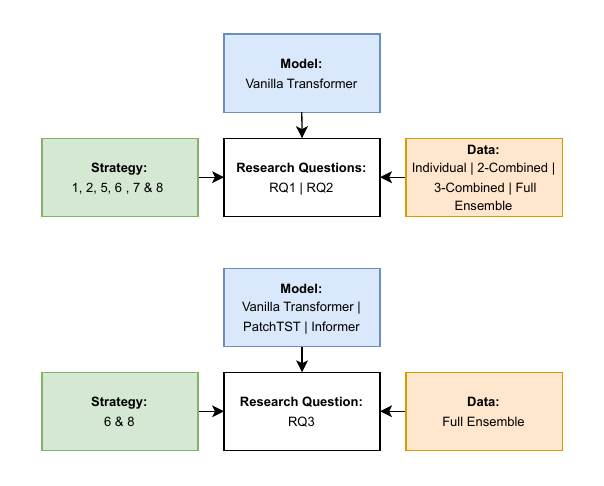}
    \caption{Conceptual diagram of the research framework}
    \label{fig:ConceptualDiagram2}
\end{figure}

In order to answer the research questions, we experimented with different data-centric TL strategies, using different dataset combinations mentioned above and three different Transformer variants. Figure \ref{fig:ConceptualDiagram2} depicts the conceptual research framework.

\begin{figure}[!t]
\begin{framed}
\fontsize{7.5}{8.5}\selectfont
\begin{itemize}
\item \colorbox[HTML]{fbb4ae}{Unmodified individual}: Original, unprocessed datasets from BDGP2.
\item \colorbox[HTML]{b3cde3}{Unmodified combined}: Combinations of two or more original datasets without modifications (meaning that these original datasets may vary with respect to climate zone, building count, etc).
\item \colorbox[HTML]{ccebc5}{Uniform}: Combinations standardised across all features for consistent evaluation (meaning that these original datasets have the same values with respect to climate zone, building count, etc.).
\item \colorbox[HTML]{decbe4}{Climate-variant}: Combined datasets vary only by the geographical location.
\item \colorbox[HTML]{fed9a6}{Building count-variant}: Combined datasets vary only by the number of buildings included in the dataset.
\item \colorbox[HTML]{ffffcc}{Weather feature-variant}: Combined datasets vary only with respect to the weather features.
\item \colorbox[HTML]{e5d8bd}{Temporal range-variant}: These combinations align datasets with different original time spans.
\end{itemize}
\end{framed}
\caption{Colour coding scheme that describes the variation in feature spaces of the combined datasets}
\label{fig:colour_code}
\end{figure}

%\newgeometry{bottom=4cm}  % Adjust the bottom margin
\begin{table}[!h]
\centering
\caption{Temporal Range-Variant and Truncated Dataset Combinations}
\vspace{0.2cm}
\label{table:clarified-temporal-range-variant}
\scriptsize % Reduce font size; you can also try \footnotesize
\setlength{\tabcolsep}{3pt} % Reduce horizontal padding (default is 6pt)
\renewcommand{\arraystretch}{0.9} % Reduce vertical padding (default is 1)
\begin{tabularx}{\textwidth}{>{\raggedright\arraybackslash}p{0.17\textwidth} >{\raggedright\arraybackslash}p{0.22\textwidth} X}
\toprule
\textbf{Dataset} & \textbf{Composition} & \textbf{Key Modifications} \\
\midrule
\rowcolor[HTML]{ffffcc}
\textbf{Peacock + Wolftruncated1} &
36 Buildings each, Climate Zone 5A &
\textbf{Wolftruncated1} is derived from the original Wolf dataset by reducing its number of weather features from 6 to 3 to match the Peacock dataset. %Both datasets now have 36 buildings and 3 weather features in Climate Zone 5A. This allows for analysis of model performance when datasets are identical in all aspects except the original number of weather features. 
\\
\midrule
\rowcolor[HTML]{fed9a6}
\textbf{Eagletruncated1 + Robin} &
50 Buildings each, Climate Zone 4A &
\textbf{Eagletruncated1} is a version of the Eagle dataset where the number of buildings is reduced from 87 to 50 to match the Robin dataset. %Both datasets now have 50 buildings and 5 weather features in Climate Zone 4A. This facilitates a fair comparison in model training and evaluation with identical dataset sizes. 
\\
\midrule
\rowcolor[HTML]{decbe4}
\textbf{Bear + Foxtruncated1} &
73 Buildings each, Bear: Climate Zone 3C, Foxtruncated1: Climate Zone 2B &
\textbf{Foxtruncated1} is created from the Fox dataset by reducing the number of buildings from 127 to 73 to match the Bear dataset.% Both datasets have 73 buildings and 5 weather features but are from different climate zones. This combination allows exploration of the impact of climate zone differences on model performance with identical dataset sizes. 
\\
\midrule
\rowcolor[HTML]{decbe4}
\textbf{Bobcat + Moosetruncated1} &
7 Buildings each, Bobcat: Climate Zone 5B, Moosetruncated1: Climate Zone 6A &
\textbf{Moosetruncated1} is derived from the Moose dataset by reducing the number of buildings from 9 to 7 to match the Bobcat dataset. %Both datasets have 7 buildings and 5 weather features but are from different climate zones. This allows analysis of model performance when datasets are identical in size but differ in climate zone. 
\\
\midrule
\rowcolor[HTML]{ffffcc}
\textbf{Lamb + Robintruncated1} &
41 Buildings each, Climate Zone 4A &
\textbf{Robintruncated1} is created by reducing the number of weather features in the Robin dataset from 5 to 4 to match the Lamb dataset. %Both datasets now have 41 buildings and 4 weather features in Climate Zone 4A. This enables the study of model performance with identical datasets except for the number of weather features. 
\\
\midrule
\rowcolor[HTML]{fed9a6}
\textbf{Bulltruncated1 + Gator} &
Bulltruncated1: 41 Buildings, Gator: 29 Buildings, Climate Zone 2A &
\textbf{Bulltruncated1} is derived from the Bull dataset by reducing the number of weather features from 3 to 0 to match the Gator dataset, which has no weather features. %Both datasets are from Climate Zone 2A but differ in the number of buildings. This combination explores the effects of varying building counts and weather feature availability on model performance. 
\\
\midrule
\rowcolor[HTML]{e5d8bd}
\textbf{Hogtruncated1 + Moosetruncated2} &
9 Buildings each, Climate Zone 6A &
\textbf{Hogtruncated1} is derived from the Hog dataset by retaining only 35\% of the data for training, excluding 35\% by zero-padding in the middle, and using the remaining 30\% for validation and testing. \textbf{Moosetruncated2} is obtained by excluding the first 35\% of data through zero-padding, using the next 35\% for training, and the final 30\% for validation and testing. %This combination investigates the impact of different temporal truncation positions on model training and performance. Both datasets are from Climate Zone 6A. 
\\
\midrule
\rowcolor[HTML]{e5d8bd}
\textbf{Hogtruncated2 + Moosetruncated2} &
9 Buildings each, Climate Zone 6A &
\textbf{Hogtruncated2} is created from the Hog dataset without zero-padding, using a standard 70\% training and 30\% validation/testing split. \textbf{Moosetruncated2} is the same as in the previous combination.% Comparing this combination with the previous one helps assess the effects of zero-padding and data exclusion on model outcomes when combined with a consistently truncated dataset. 
\\
\bottomrule
\end{tabularx}
\end{table}
%\restoregeometry

%\newgeometry{bottom=4cm}  % Adjust bottom margin to push down page number
\begin{table}[p]
\centering
\caption{Dataset Characteristics and Experiment Categories}
\vspace{0.2cm}
\label{table:dataset_characteristics}
\scriptsize
\begin{adjustbox}{max width=\textwidth,center}
\begin{tabular}{>{\raggedright\arraybackslash}p{40mm}>{\raggedleft\arraybackslash}p{13.5mm}>{\raggedleft\arraybackslash}p{5mm}>{\raggedleft\arraybackslash}p{11mm}>{\raggedleft\arraybackslash}p{11mm}>{\raggedleft\arraybackslash}p{11mm}>{\raggedleft\arraybackslash}p{11mm}>{\raggedleft\arraybackslash}p{11mm}>{\raggedleft\arraybackslash}p{11mm}}
\toprule
\textbf{Dataset} & \textbf{B} & \textbf{CZ} & \textbf{AT} & \textbf{DT} & \textbf{SLP} & \textbf{WD} & \textbf{WS} & \textbf{CC}\\
\midrule
\rowcolor[HTML]{fbb4ae}
Bear & 73 & 3C & 1 & 1 & 1 & 1 & 1 & 0 \\
\rowcolor[HTML]{fbb4ae}
Bobcat & 7 & 5B & 1 & 1 & 1 & 1 & 1 & 0 \\
\rowcolor[HTML]{fbb4ae}
Bull & 41 & 2A & 1 & 1 & 1 & 0 & 0 & 0 \\
\rowcolor[HTML]{fbb4ae}
Cockatoo & 1 & 6A & 1 & 1 & 1 & 1 & 1 & 0 \\
\rowcolor[HTML]{fbb4ae}
Crow & 4 & 6A & 1 & 1 & 1 & 1 & 1 & 0 \\
\rowcolor[HTML]{fbb4ae}
Eagle & 87 & 4A & 1 & 1 & 1 & 1 & 1 & 0 \\
\rowcolor[HTML]{fbb4ae}
Fox & 127 & 2B & 1 & 1 & 1 & 1 & 1 & 0 \\
\rowcolor[HTML]{fbb4ae}
Gator & 29 & 2A & 0 & 0 & 0 & 0 & 0 & 0 \\
\rowcolor[HTML]{fbb4ae}
Hog & 24 & 6A & 1 & 1 & 1 & 1 & 1 & 0 \\
\rowcolor[HTML]{fbb4ae}
Lamb & 41 & 4A & 1 & 1 & 0 & 1 & 1 & 0 \\
\rowcolor[HTML]{fbb4ae}
Moose & 9 & 6A & 1 & 1 & 1 & 1 & 1 & 0 \\
\rowcolor[HTML]{fbb4ae}
Mouse & 3 & 4A & 1 & 1 & 1 & 1 & 1 & 0 \\
\rowcolor[HTML]{fbb4ae}
Peacock & 36 & 5A & 1 & 1 & 1 & 0 & 0 & 0 \\
\rowcolor[HTML]{fbb4ae}
Rat & 251 & 4A & 1 & 1 & 1 & 1 & 1 & 0 \\
\rowcolor[HTML]{fbb4ae}
Robin & 50 & 4A & 1 & 1 & 1 & 1 & 1 & 0 \\
\rowcolor[HTML]{fbb4ae}
Wolf & 36 & 5A & 1 & 1 & 1 & 1 & 1 & 1 \\
\midrule
\rowcolor[HTML]{b3cde3}
Bull + Cockatoo + Hog & 66 & MC & 3×1 & 3×1 & 3×1 & 2×1,1×0 & 2×1,1×0 & 3×0 \\
\rowcolor[HTML]{b3cde3}
Bear + Fox & 200 & MC & 2×1 & 2×1 & 2×1 & 2×1 & 2×1 & 2×0 \\
\rowcolor[HTML]{b3cde3}
Bobcat + Moose & 16 & MC & 2×1 & 2×1 & 2×1 & 2×1 & 2×1 & 2×0 \\
\rowcolor[HTML]{b3cde3}
Bull + Gator & 70 & 2A & 1×1,1×0 & 1×1,1×0 & 1×1,1×0 & 2×0 & 2×0 & 2×0 \\
\rowcolor[HTML]{b3cde3}
Crow + Robin & 54 & MC & 2×1 & 2×1 & 2×1 & 2×1 & 2×1 & 2×0 \\
\rowcolor[HTML]{b3cde3}
Hog + Moose & 33 & 6A & 2×1 & 2×1 & 2×1 & 2×1 & 2×1 & 2×0 \\
\rowcolor[HTML]{b3cde3}
Lamb + Robin & 91 & 4A & 2×1 & 2×1 & 1×1,1×0 & 2×1 & 2×1 & 2×0 \\
\rowcolor[HTML]{b3cde3}
Mouse + Rat & 254 & 4A & 2×1 & 2×1 & 2×1 & 2×1 & 2×1 & 2×0 \\
\midrule
\rowcolor[HTML]{ccebc5}
Peacock + Wolftruncated1 & 36×2 & 5A & 2×1 & 2×1 & 2×1 & 2×0 & 2×0 & 2×0 \\
\rowcolor[HTML]{ccebc5}
Eagletruncated1 + Robin & 50×2 & 4A & 2×1 & 2×1 & 2×1 & 2×1 & 2×1 & 2×0 \\
\midrule
\rowcolor[HTML]{decbe4}
Bear + Foxtruncated1 & 73×2 & MC & 2×1 & 2×1 & 2×1 & 2×1 & 2×1 & 2×0 \\
\rowcolor[HTML]{decbe4}
Bobcat + Moosetruncated1 & 7x2 & MC & 2×1 & 2×1 & 2×1 & 2×1 & 2×1 & 2×0 \\
\midrule
\rowcolor[HTML]{fed9a6}
Eagle + Robin & 87×1,50×1 & 4A & 2×1 & 2×1 & 2×1 & 2×1 & 2×1 & 2×0 \\
\rowcolor[HTML]{fed9a6}
Bulltruncated1 + Gator & 41×1,29×1 & 2A & 2×0 & 2×0 & 2×0 & 2×0 & 2×0 & 2×0 \\
\midrule
\rowcolor[HTML]{ffffcc}
Peacock + Wolf & 36×2 & 5A & 2×1 & 2×1 & 2×1 & 1×1,1×0 & 1×1,1×0 & 1×1,1×0 \\
\rowcolor[HTML]{ffffcc}
Lamb + Robintruncated1 & 41×2 & 4A & 2×1 & 2×1 & 1×1,1×0 & 2×1 & 2×1 & 2×0 \\
\midrule
\rowcolor[HTML]{e5d8bd}
Hogtruncated1 + Moosetruncated2 & 9×2 & 6A & 2×1 & 2×1 & 2×1 & 2×1 & 2×1 & 2×0 \\
\rowcolor[HTML]{e5d8bd}
Hogtruncated2 + Moosetruncated2 & 9×2 & 6A & 2×1 & 2×1 & 2×1 & 2×1 & 2×1 & 2×0 \\
\midrule
\rowcolor[HTML]{fddaec}
Full Ensemble & 819 & MC & 15×1,1×0 & 15×1,1×0 & 14×1,2×0 & 13×1,3×0 & 13×1,3×0 & 1×1,15×0 \\
\bottomrule
\end{tabular}
\end{adjustbox}
\vspace{0.2cm}
\par\noindent
\begin{minipage}{\linewidth}
\raggedright
\tiny
\textbf{Abbreviations:} B: Number of buildings, CZ: Climate Zone, MC: Mixed Climates, AT: Air Temperature, DT: Dew Temperature, SLP: Sea Level Pressure, WD: Wind Direction, WS: Wind Speed, CC: Cloud Coverage, MC: different climate zones.

\textbf{Colour coding:}
\colorbox[HTML]{fbb4ae}{Unmodified individual}, 
\colorbox[HTML]{b3cde3}{Unmodified combined}, 
\colorbox[HTML]{ccebc5}{Uniform}, 
\colorbox[HTML]{decbe4}{Climate-variant}, 
\colorbox[HTML]{fed9a6}{Building count-variant}, 
\colorbox[HTML]{ffffcc}{Weather feature-variant}, 
\colorbox[HTML]{e5d8bd}{Temporal range-variant}, 
\colorbox[HTML]{fddaec}{Full Ensemble}.

For combined datasets, entries show the count of each feature value (e.g., 2×1,1×0 means two '1' values and one '0' value). Full Ensemble includes all unmodified individual datasets. Truncation (indicated by 'truncated1' or 'truncated2' appended to dataset names) refers to reducing one or more of these: the number of buildings, the number of weather features, or the temporal range of the dataset.
\end{minipage}
\end{table}
%\restoregeometry

\subsection{Performance Comparison of Different TL Strategies with Vanilla Transformers}

For the vanilla Transformer, we implemented six of the eight data-centric TL strategies (1-2 and 5-8). Strategies 3 and 4 were not considered because if target data is available, fine-tuning a model by combining that target data with the source data (strategies 5-8) is always beneficial.%, creating a total of 600 models:

%\paragraph{Baseline (114 models)}
\paragraph{Baseline}
We trained vanilla Transformer models with the training split of each of the dataset, and test with the tested split.
%We conducted baseline experiments for 19 datasets (16 original + 3 truncated datasets).  %Each experiment was repeated three times with different random seeds (1, 50, and 100), resulting in:
\iffalse
\begin{equation*}
\begin{split}
\text{Total models} &= \text{Number of datasets (19)} \times \text{Forecast horizons (2)} \times \text{Seeds (3)} \\
&= 19 \times 2 \times 3 = 114 \text{ models}
\end{split}
\end{equation*}
\fi
%The three truncated datasets are Hogtruncated1, Hogtruncated2 and Moosetruncated2.

\paragraph{Strategy 1 (Single Source Zero-shot)}

We used the baseline models for zero-shot testing on other datasets. No additional models were trained for this strategy.

%\paragraph{Strategy 2 (Multi-source Zero-shot Learning, 132 models)}
\paragraph{Strategy 2 (Multi-source Zero-shot)}
We trained models on combinations of datasets and tested them on the test splits of other datasets (excluding those used in training). The combinations included two-dataset combinations and one three-dataset combination.
\iffalse
\begin{itemize}
    \item \textbf{Two-dataset combinations}: 40 experiments
    \item \textbf{Three-dataset combinations}: 2 experiments
    \item \textbf{Full Ensemble}: 2 experiments
\end{itemize}

For each combination, we evaluated two forecast horizons and repeated each experiment with three different seeds, leading to:

\begin{equation*}
\begin{aligned}
\text{Total models} &= \text{Two-dataset models} + \text{Three-dataset models} + \text{Full ensemble models} \\
&= 120 + 6 + 6 \\
&= 132 \text{ models}
\end{aligned}
\end{equation*}
\fi

\paragraph{Strategies 5 and 6}
We reused the models trained under Strategy 2 and tested them with the test splits of the datasets that were used to train those models. In addition, we tested with the full ensemble (combining all 16 datasets). %The full ensemble models are considered under {Strategy 6} (Non-fine-tuned full ensemble models).

%\paragraph{Strategy 7 (Fine-tuning Two-Dataset Models, 240 models)}
\paragraph{Strategy 7 (Fine-tuning Two-Dataset Models)}
We further fine-tuned the models trained on two-dataset combinations with additional training on each target dataset alone. %For each of the 20 two-dataset combinations, we evaluated two forecast horizons and used three seeds, resulting in:
\iffalse
\begin{equation*}
\begin{aligned}
\text{Number of initial models} &= \text{Number of combinations} \times \text{Forecast horizons} \times \text{Seeds} \\
&= 20 \times 2 \times 3 \\
&= 120 \text{ models}
\end{aligned}
\end{equation*}

Since we fine-tuned on both datasets in each combination, the total number of fine-tuned models is:

\begin{equation*}
\begin{aligned}
\text{Total models} &= \text{Number of initial models} \times \text{Fine-tuning on datasets} \\
&= 120 \times 2 \\
&= 240 \text{ models}
\end{aligned}
\end{equation*}
\fi
\paragraph{Strategy 8 (Fine-tuning Three-Dataset Models and Full Ensemble Models)}

We fine-tuned the models trained on three-dataset combinations and the full ensemble model on each individual dataset.

During fine-tuning, we adjusted all parameters of the model without freezing any layers. The findings of Santos et al. \cite{LOPEZSANTOS2023113164} support this approach, which demonstrated the benefits of fine-tuning all parameters on Transformers instead of freezing layers in load forecasting contexts.

\subsection{Evaluation of Transformer Variants}

We extended our experiments to include advanced Transformer variants: Informer and PatchTST, to compare their performance under similar experimental conditions. For both Informer and PatchTST architectures, we conducted the following experiments: baseline experiments, Strategy 6 (full ensemble model) and Strategy 8 (further fine-tuning full ensemble model).

\section{Experiment Setup}
%This study employs a comprehensive and diverse collection of datasets from the BDGP2, focusing on building energy consumption across various climates, locations, and building types. The datasets represent a broad spectrum of energy usage patterns influenced by various factors, including weather conditions, building characteristics, and occupancy patterns. 

\subsection{Dataset Description}
This study employs a comprehensive and diverse collection of datasets from the BDGP2. The datasets include hourly energy consumption data for various buildings across multiple geographic locations and climate zones, such as educational institutions, offices, and public buildings. Each dataset exhibits a unique combination of features that influence energy consumption:

\begin{itemize}
    \item Load Type: All datasets focus on building energy consumption.
    \item Datetime Range and Number of Rows: The data spans two full years (01/01/2016 — 31/12/2017), with each dataset containing 17,544 hourly records, 53.6 million in total. This provides a substantial temporal range for analysing seasonal variations and trends in energy use.
    \item Granularity: Data is recorded hourly, allowing for detailed analysis of daily and seasonal peaks in energy demand.
    \item Actual Site Name and Location: Researchers collected data from 19 sites across North America and Europe, anonymising some datasets while specifying other locations. This variation allows researchers to explore how regional climate conditions and building practices influence energy consumption.
    \item Climate Conditions: The datasets are categorised using ASHRAE climate zones \cite{crawley2020ansi} (e.g., `3C - Warm and marine'), enabling the study of climate-specific energy usage patterns and the development of tailored forecasting models.
    \item Building and Weather Features: Detailed information on operational characteristics (e.g., building type and usage) and weather conditions (e.g., temperature, humidity) is included.
    \item Data Split: The datasets are divided into training, validation, and test sets using a 70\%/20\%/10\% split ratio, respectively. %This split ensures a robust evaluation of model performance, with a substantial portion for training, a sizeable validation set for hyperparameter tuning and model selection, and a held-out test set for final performance assessment.
\end{itemize}

% \subsection{Data Preprocessing}
% \su{this section has to be improved}
% Despite the initial cleaning, the BDGP2 datasets still contained outlier buildings and weather features—identified using statistical thresholds and manually checking the data—that necessitated further preprocessing to refine the data quality for load forecasting\su{sentence is too long and not clear}. Therefore, in this study, we utilised the ProEnFo package, as outlined in Wang et al. \cite{wang2023benchmarks}, to further clean the BDGP2 datasets as follows:

% We addressed this by removing buildings with 10\% or more missing values to maintain data integrity. Timestamp data were converted into a datetime index to align the BDGP2 datasets for time series analysis, and linear interpolation was employed to impute missing values. Post-interpolation, we eliminated columns with a count of zero values exceeding 3000 to ensure the datasets' quality. 16 datasets remained after processing. These processed datasets are delineated in Table~\ref{table:dataset_characteristics}. 

\subsection{Data Preprocessing and Dataset Combinations}
The BDGP2 datasets underwent initial cleaning by the original dataset authors. However, further preprocessing was necessary to refine the data quality for load forecasting. We identified outlier buildings and weather features using statistical thresholds. To address these issues, we employed the ProEnFo package, as described by Wang et al. \cite{wang2023benchmarks}, to further clean the BDGP2 datasets. We removed buildings with 10\% or more missing values to maintain data integrity. Then we used linear interpolation to impute missing values. After interpolation, we eliminated columns with more than 3000 zero values to ensure dataset quality. We also converted timestamp data into a date-time index to facilitate time series analysis.

Following this preprocessing, 16 datasets remained. In order to run experiments involving multiple sources, we combined some of these datasets. Note that we did not try out all the possible dataset combinations, as this results in a very large number of experiments. Instead, we combined datasets in such a way that we obtained combined datasets that have varying feature spaces. Figure~\ref{fig:colour_code} describes the variation in feature spaces. Note that in order to obtain a diverse set of feature spaces, we had to truncate some datasets. We employed three types of truncation. The first one is feature truncation, which was applied for weather feature-variant combinations - we simply removed specific weather feature columns from the dataset. Secondly, we carried out temporal truncation, where we removed a portion of the time series data and replaced it with zero-padding for temporal range-variant combinations. This approach allows us to simulate scenarios with limited historical data while maintaining the overall structure of the dataset. Finally, we carried out building data truncation, where we removed energy data related to entire buildings. Details of dataset truncation are mentioned in Table~\ref{table:clarified-temporal-range-variant}. The final set of datasets used in our experiments is detailed in Table~\ref{table:dataset_characteristics}.

\subsection{Model Hyper-parameters and Computational Setup}
Our implementation of  Vanilla Transformer, Informer, and PatchTST is based on Hertel et al. \cite{hertel2023transformer}. We reused their hyper-parameters, except for the following: Learning rate 0.0001 for strategies 1, 2, 5, and 6 and 0.00001 for strategies 7 and 8. 

To train the models, we used an NVIDIA GeForce RTX 3090 GPU. Models are implemented using PyTorch library running on Python 3.10. Following previous research, evaluation results are reported using MAE and MSE. We use MAE results for the discussion in the main text, and MSE results can be found in the Appendix. %We report results for both 24hr and 96hr horizons.

\subsection{Models and Experiments}
Table \ref{table} provides a comprehensive summary of the models trained across different strategies and architectures in our study. As shown, we conducted a total of 332 unique experiments. Each experiment was conducted 3 times using different seeds, to avoid any form of bias, resulting in 996 models altogether. Final result of each experiment is the average of these three runs. For each dataset, we evaluated two forecast horizons: 24 hours and 96 hours. This extensive set of experiments allows for a thorough comparison of performance across different approaches.
\begin{table}[h]
\centering
\caption{Summary of Models Trained Across Strategies and Architectures}
\vspace{2mm}
\label{table}
\resizebox{\textwidth}{!}{%
\begin{tabular}{lcc}
\toprule
\textbf{Strategy} & \textbf{Experiments} & \textbf{Total Models} \\
\midrule
\multicolumn{3}{l}{\textbf{Vanilla Transformer}} \\
Baseline & 38 & 114 \\
Strategy 2 (Multi-source Zero-shot) & 44 & 132 \\
Strategy 7 (Fine-tuning Two-Dataset Models) & 80 & 240 \\
Strategy 8 (Fine-tuning Three-Dataset Models) & 6 & 18 \\
Strategy 8 (Fine-tuning Full Ensemble Models) & 32 & 96 \\
\textbf{Subtotal} & \textbf{200} & \textbf{600} \\
\midrule
\multicolumn{3}{l}{\textbf{Informer}} \\
Baseline & 32 & 96 \\
Strategy 6 (Non-fine-tuned Full Ensemble Models) & 2 & 6 \\
Strategy 8 (Fine-tuning Full Ensemble Models) & 32 & 96 \\
\textbf{Subtotal} & \textbf{66} & \textbf{198} \\
\midrule
\multicolumn{3}{l}{\textbf{PatchTST}} \\
Baseline & 32 & 96 \\
Strategy 6 (Non-fine-tuned Full Ensemble Models) & 2 & 6 \\
Strategy 8 (Fine-tuning Full Ensemble Models) & 32 & 96 \\
\textbf{Subtotal} & \textbf{66} & \textbf{198} \\
\midrule
\textbf{Grand Total} & \textbf{332} & \textbf{996} \\
\bottomrule
\end{tabular}%
}

\end{table}

\section{Results}\label{sec:results}

\subsection{Results of Zero-Shot Testing}
\label{sec:result_zero}
These results correspond to Strategies 1 and 2. Tables~\ref{zero-shot_24h} and~\ref{zero-shot_96h} show the MAE results for 24hr and 96hr forecasting, respectively (Corresponding MSE results can be found it Tables~\ref{tab:24MSE} and~\ref{tab:96MSE} in Appendix). The top part of the table contains results for models trained with individual datasets. The bottom part shows the models trained with multiple datasets. In these tables, values highlighted in grey are the baseline results (i.e. train and test set belong to the same dataset). For models trained with multiple datasets, when a model is tested with a dataset that is included in the model, those columns are highlighted in purple. To reiterate, results highlighted in grey or purple are not zero-shot results - these were included in tables for comparison purposes.
%Zero-shot testing evaluates a model's ability to generalise to unseen data without fine-tuning. This subsection presents the results of zero-shot testing for both individual and combined dataset models trained on BDGP2 datasets and applied to other datasets not involved in their training.

\begin{sidewaystable}
\centering
\caption{Average MAE for 24-hour Horizon Zero-shot Forecasting}
\label{zero-shot_24h}
\vspace{0.2cm}
\tiny
\begin{adjustbox}{max width=1\textwidth,center}
\begin{tabular}{lrrrrrrrrrrrrrrrr}
\toprule
\textbf{Model} & \textbf{Bear} & \textbf{Bobcat} & \textbf{Bull} & \textbf{Cockatoo} & \textbf{Crow} & \textbf{Eagle} & \textbf{Fox} & \textbf{Gator} & \textbf{Hog} & \textbf{Lamb} & \textbf{Moose} & \textbf{Mouse} & \textbf{Peacock} & \textbf{Rat} & \textbf{Robin} & \textbf{Wolf} \\
\midrule
Bear & \cellcolor{gray!20}0.305 & 0.393 & 0.483 & 0.483 & 0.346 & 0.401 & 0.296 & 0.432 & 0.347 & 0.237 & 0.293 & 0.362 & 0.334 & 0.320 & 0.274 & 0.396 \\
Bobcat & 0.428 & \cellcolor{gray!20}0.415 & 0.509 & 0.471 & 0.368 & 0.498 & 0.393 & 0.438 & 0.360 & 0.335 & 0.337 & 0.404 & 0.414 & 0.391 & 0.382 & 0.464 \\
Bull & 0.442 & 0.478 & \cellcolor{gray!20}0.475 & 0.503 & 0.389 & 0.469 & 0.404 & 0.480 & 0.360 & 0.279 & 0.417 & 0.417 & 0.415 & 0.388 & 0.388 & 0.509 \\
Cockatoo & 0.678 & 0.634 & 0.582 & \cellcolor{gray!20}0.468 & 0.418 & 0.731 & 0.623 & 0.466 & 0.377 & 0.407 & 0.575 & 0.426 & 0.620 & 0.532 & 0.589 & 0.652 \\
Crow & 0.491 & 0.592 & 0.646 & 0.503 & \cellcolor{gray!20}0.359 & 0.717 & 0.488 & 0.452 & 0.440 & 0.440 & 0.475 & 0.413 & 0.516 & 0.476 & 0.456 & 0.526 \\
Eagle & 0.435 & 0.473 & 0.506 & 0.464 & 0.357 & \cellcolor{gray!20}\textbf{0.331} & 0.409 & 0.414 & 0.351 & 0.279 & 0.512 & 0.360 & 0.418 & 0.388 & 0.334 & 0.528 \\
Fox & 0.302 & 0.381 & \textbf{0.461} & 0.473 & 0.337 & 0.371 & \cellcolor{gray!20}0.268 & 0.433 & 0.329 & 0.207 & 0.281 & 0.355 & 0.322 & 0.302 & 0.265 & 0.402 \\
Gator & 0.838 & 0.698 & 0.581 & 0.523 & 0.492 & 0.795 & 0.758 & \cellcolor{gray!20}\textbf{0.292} & 0.391 & 0.406 & 0.626 & 0.490 & 0.747 & 0.578 & 0.724 & 0.803 \\
Hog & 0.499 & 0.521 & 0.515 & 0.470 & 0.362 & 0.610 & 0.468 & 0.432 & \cellcolor{gray!20}0.336 & 0.298 & 0.476 & 0.376 & 0.476 & 0.424 & 0.426 & 0.550 \\
Lamb & 0.383 & 0.461 & 0.530 & 0.497 & 0.336 & 0.496 & 0.371 & 0.440 & 0.353 & \cellcolor{gray!20}0.208 & 0.319 & 0.380 & 0.405 & 0.363 & 0.326 & 0.416 \\
Moose & 0.467 & 0.535 & 0.610 & 0.500 & 0.364 & 0.738 & 0.463 & 0.460 & 0.390 & 0.308 & \cellcolor{gray!20}0.294 & 0.411 & 0.498 & 0.436 & 0.427 & 0.446 \\
Mouse & 0.445 & 0.535 & 0.548 & 0.470 & 0.364 & 0.608 & 0.446 & 0.422 & 0.367 & 0.356 & 0.461 & \cellcolor{gray!20}0.365 & 0.470 & 0.441 & 0.374 & 0.499 \\
Peacock & 0.327 & 0.437 & 0.542 & 0.507 & 0.335 & 0.424 & 0.320 & 0.439 & 0.372 & 0.246 & 0.301 & 0.365 & \cellcolor{gray!20}\textbf{0.316} & 0.340 & 0.280 & 0.410 \\
Rat & 0.321 & 0.395 & 0.490 & 0.472 & 0.335 & 0.402 & 0.293 & 0.419 & 0.334 & 0.216 & 0.288 & 0.357 & 0.339 & \cellcolor{gray!20}\textbf{0.293} & 0.296 & 0.409 \\
Robin & 0.338 & 0.437 & 0.510 & 0.462 & 0.321 & 0.419 & 0.338 & 0.405 & 0.336 & 0.232 & 0.295 & 0.347 & 0.353 & 0.344 & \cellcolor{gray!20}0.254 & 0.411 \\
Wolf & 0.417 & 0.494 & 0.610 & 0.533 & 0.382 & 0.536 & 0.405 & 0.492 & 0.424 & 0.365 & 0.360 & 0.419 & 0.445 & 0.425 & 0.370 & \cellcolor{gray!20}0.387 \\
Hogtrunc2 & 0.545 & 0.533 & 0.516 & 0.457 & 0.377 & 0.607 & 0.501 & 0.420 & \cellcolor{gray!20}0.331 & 0.317 & 0.514 & 0.392 & 0.505 & 0.445 & 0.454 & 0.584 \\
Hogtrunc & 0.636 & 0.581 & 0.539 & 0.504 & 0.434 & 0.653 & 0.588 & 0.448 & \cellcolor{gray!20}0.382 & 0.346 & 0.546 & 0.435 & 0.575 & 0.484 & 0.545 & 0.621 \\
Moosetrunc2 & 0.629 & 0.631 & 0.592 & 0.533 & 0.445 & 0.772 & 0.613 & 0.473 & 0.408 & 0.360 & \cellcolor{gray!20}0.459 & 0.457 & 0.618 & 0.515 & 0.581 & 0.634 \\
Bear+Fox & \cellcolor{blue!10}\textbf{0.291} & \textbf{0.378} & 0.463 & 0.466 & 0.330 & 0.376 & \cellcolor{blue!10}\textbf{0.264} & 0.422 & 0.326 & 0.212 & 0.279 & 0.347 & 0.320 & 0.299 & 0.263 & 0.395 \\
Bobcat+Moose & 0.384 & \cellcolor{blue!10}0.399 & 0.522 & \textbf{0.450} & 0.331 & 0.503 & 0.365 & 0.420 & 0.337 & 0.268 & \cellcolor{blue!10}\textbf{0.278} & 0.365 & 0.392 & 0.364 & 0.330 & 0.419 \\
Bull+Cockatoo & 0.451 & 0.475 & \cellcolor{blue!10}0.476 & \cellcolor{blue!10}0.468 & 0.367 & 0.496 & 0.415 & 0.449 & 0.352 & 0.278 & 0.406 & 0.390 & 0.417 & 0.392 & 0.398 & 0.492 \\
Bull+Gator & 0.493 & 0.499 & \cellcolor{blue!10}0.480 & 0.491 & 0.379 & 0.565 & 0.457 & \cellcolor{blue!10}0.312 & 0.354 & 0.288 & 0.448 & 0.396 & 0.450 & 0.410 & 0.436 & 0.536 \\
Bull+Hog & 0.433 & 0.455 & \cellcolor{blue!10}0.468 & 0.468 & 0.357 & 0.501 & 0.396 & 0.440 & \cellcolor{blue!10}0.325 & 0.271 & 0.383 & 0.380 & 0.404 & 0.372 & 0.380 & 0.483 \\
Cockatoo+Hog & 0.488 & 0.512 & 0.521 & \cellcolor{blue!10}0.451 & 0.347 & 0.620 & 0.458 & 0.426 & \cellcolor{blue!10}0.323 & 0.301 & 0.447 & 0.375 & 0.473 & 0.416 & 0.426 & 0.533 \\
Crow+Robin & 0.337 & 0.441 & 0.543 & 0.460 & \cellcolor{blue!10}\textbf{0.313} & 0.432 & 0.343 & 0.400 & 0.337 & 0.233 & 0.301 & \textbf{0.344} & 0.350 & 0.347 & \cellcolor{blue!10}0.255 & 0.407 \\
Eagle+Robin & 0.463 & 0.488 & 0.506 & 0.464 & 0.374 & \cellcolor{blue!10}0.354 & 0.432 & 0.423 & 0.349 & 0.289 & 0.521 & 0.378 & 0.439 & 0.402 & \cellcolor{blue!10}0.368 & 0.547 \\
Hog+Moose & 0.423 & 0.467 & 0.530 & 0.464 & 0.343 & 0.608 & 0.410 & 0.408 & \cellcolor{blue!10}0.323 & 0.262 & \cellcolor{blue!10}0.297 & 0.366 & 0.437 & 0.384 & 0.368 & 0.435 \\
Lamb+Robin & 0.322 & 0.411 & 0.516 & 0.474 & 0.321 & 0.393 & 0.311 & 0.403 & 0.337 & \cellcolor{blue!10}\textbf{0.185} & 0.284 & 0.358 & 0.341 & 0.322 & \cellcolor{blue!10}\textbf{0.248} & 0.395 \\
Mouse+Rat & 0.319 & 0.395 & 0.493 & 0.467 & 0.330 & 0.397 & 0.292 & 0.412 & 0.332 & 0.216 & 0.286 & \cellcolor{blue!10}0.348 & 0.339 & \cellcolor{blue!10}0.294 & 0.289 & 0.407 \\
Peacock+Wolf & 0.341 & 0.422 & 0.537 & 0.493 & 0.339 & 0.424 & 0.327 & 0.435 & 0.365 & 0.242 & 0.305 & 0.361 & \cellcolor{blue!10}0.329 & 0.347 & 0.286 & \cellcolor{blue!10}0.379 \\
Bear+Foxtrunc & \cellcolor{blue!10}0.295 & \textbf{0.378} & 0.462 & 0.471 & 0.336 & 0.380 & \cellcolor{blue!10}0.273 & 0.425 & 0.331 & 0.212 & 0.279 & 0.352 & 0.322 & 0.304 & 0.263 & 0.394 \\
Bobcat+Moosetrunc & 0.388 & \cellcolor{blue!10}0.400 & 0.518 & \textbf{0.450} & 0.333 & 0.503 & 0.368 & 0.417 & 0.339 & 0.282 & \cellcolor{blue!10}0.282 & 0.364 & 0.397 & 0.368 & 0.336 & 0.420 \\
Bulltrunc+Gator & 0.490 & 0.505 & \cellcolor{blue!10}0.483 & 0.499 & 0.384 & 0.576 & 0.459 & \cellcolor{blue!10}0.308 & 0.357 & 0.279 & 0.454 & 0.393 & 0.451 & 0.413 & 0.433 & 0.524 \\
Eagletrunc+Robin & 0.463 & 0.488 & 0.506 & 0.464 & 0.374 & \cellcolor{blue!10}0.354 & 0.432 & 0.423 & 0.349 & 0.289 & 0.521 & 0.378 & 0.439 & 0.402 & \cellcolor{blue!10}0.368 & 0.547 \\
Hogtrunc2+Moosetrunc2 & 0.518 & 0.565 & 0.550 & 0.503 & 0.395 & 0.688 & 0.507 & 0.426 & \cellcolor{blue!10}0.371 & 0.305 & \cellcolor{blue!10}0.408 & 0.401 & 0.544 & 0.453 & 0.464 & 0.562 \\
Hogtrunc+Moosetrunc2 & 0.518 & 0.565 & 0.550 & 0.503 & 0.395 & 0.688 & 0.507 & 0.426 & \cellcolor{blue!10}0.371 & 0.305 & \cellcolor{blue!10}0.408 & 0.401 & 0.544 & 0.453 & 0.464 & 0.562 \\
Lamb+Robintrunc & 0.337 & 0.411 & 0.487 & 0.473 & 0.333 & 0.399 & 0.321 & 0.404 & 0.328 & \cellcolor{blue!10}0.218 & 0.300 & 0.365 & 0.353 & 0.331 & \cellcolor{blue!10}0.271 & 0.405 \\
Peacock+Wolftrunc & 0.340 & 0.424 & 0.522 & 0.492 & 0.348 & 0.417 & 0.327 & 0.427 & 0.368 & 0.248 & 0.318 & 0.370 & \cellcolor{blue!10}0.329 & 0.344 & 0.288 & \cellcolor{blue!10}\textbf{0.376} \\
Bull+Cockatoo+Hog & 0.440 & 0.460 & \cellcolor{blue!10}0.469 & \cellcolor{blue!10}0.455 & 0.351 & 0.526 & 0.409 & 0.417 & \cellcolor{blue!10}\textbf{0.322} & 0.276 & 0.382 & 0.374 & 0.414 & 0.379 & 0.391 & 0.487 \\
\bottomrule
\end{tabular}
\end{adjustbox}
\end{sidewaystable}

\begin{sidewaystable}
\centering
\caption{Average MAE for 96-hour Horizon Zero-shot Forecasting}
\label{zero-shot_96h}
\vspace{0.2cm}
\tiny
\begin{adjustbox}{max width=1\textwidth,center}
\begin{tabular}{lrrrrrrrrrrrrrrrr}
\toprule
\textbf{Model} & \textbf{Bear} & \textbf{Bobcat} & \textbf{Bull} & \textbf{Cockatoo} & \textbf{Crow} & \textbf{Eagle} & \textbf{Fox} & \textbf{Gator} & \textbf{Hog} & \textbf{Lamb} & \textbf{Moose} & \textbf{Mouse} & \textbf{Peacock} & \textbf{Rat} & \textbf{Robin} & \textbf{Wolf} \\
\midrule
Bear & \cellcolor{gray!20}0.375 & 0.495 & 0.600 & 0.608 & 0.470 & 0.496 & 0.365 & 0.617 & 0.484 & 0.403 & 0.385 & 0.505 & 0.419 & 0.446 & 0.370 & 0.480 \\
Bobcat & 0.554 & \cellcolor{gray!20}0.543 & 0.620 & 0.614 & 0.497 & 0.617 & 0.503 & 0.643 & 0.476 & 0.488 & 0.463 & 0.525 & 0.557 & 0.532 & 0.502 & 0.592 \\
Bull & 0.592 & 0.599 & \cellcolor{gray!20}0.602 & 0.640 & 0.506 & 0.607 & 0.530 & 0.650 & 0.472 & 0.382 & 0.552 & 0.542 & 0.553 & 0.525 & 0.540 & 0.631 \\
Cockatoo & 0.780 & 0.719 & 0.667 & \cellcolor{gray!20}0.600 & 0.545 & 0.803 & 0.708 & 0.661 & 0.488 & 0.519 & 0.662 & 0.561 & 0.726 & 0.638 & 0.686 & 0.718 \\
Crow & 0.610 & 0.727 & 0.814 & 0.658 & \cellcolor{gray!20}0.476 & 0.851 & 0.592 & 0.698 & 0.648 & 0.713 & 0.630 & 0.562 & 0.677 & 0.648 & 0.596 & 0.615 \\
Eagle & 0.569 & 0.581 & 0.653 & 0.602 & 0.483 & \cellcolor{gray!20}\textbf{0.392} & 0.518 & 0.605 & 0.499 & 0.394 & 0.656 & 0.475 & 0.546 & 0.518 & 0.446 & 0.644 \\
Fox & 0.382 & 0.480 & 0.581 & 0.591 & 0.435 & 0.454 & \cellcolor{gray!20}\textbf{0.325} & 0.595 & 0.448 & 0.339 & \textbf{0.362} & \textbf{0.463} & 0.411 & \textbf{0.413} & 0.360 & 0.484 \\
Gator & 0.843 & 0.727 & 0.677 & 0.638 & 0.568 & 0.794 & 0.753 & \cellcolor{gray!20}\textbf{0.503} & 0.503 & 0.492 & 0.703 & 0.579 & 0.751 & 0.631 & 0.748 & 0.806 \\
Hog & 0.631 & 0.635 & 0.619 & 0.606 & 0.480 & 0.703 & 0.571 & 0.623 & \cellcolor{gray!20}0.470 & 0.396 & 0.604 & 0.520 & 0.586 & 0.538 & 0.551 & 0.663 \\
Lamb & 0.476 & 0.567 & 0.674 & 0.637 & 0.460 & 0.613 & 0.464 & 0.630 & 0.496 & \cellcolor{gray!20}0.269 & 0.421 & 0.499 & 0.527 & 0.488 & 0.422 & 0.507 \\
Moose & 0.578 & 0.652 & 0.749 & 0.648 & 0.504 & 0.880 & 0.561 & 0.669 & 0.539 & 0.453 & \cellcolor{gray!20}0.377 & 0.549 & 0.646 & 0.574 & 0.551 & 0.524 \\
Mouse & 0.543 & 0.707 & 0.769 & 0.665 & 0.493 & 0.726 & 0.541 & 0.715 & 0.562 & 0.718 & 0.564 & \cellcolor{gray!20}0.532 & 0.646 & 0.626 & 0.505 & 0.599 \\
Peacock & 0.420 & 0.528 & 0.674 & 0.639 & 0.467 & 0.517 & 0.400 & 0.643 & 0.513 & 0.373 & 0.400 & 0.484 & \cellcolor{gray!20}0.411 & 0.467 & 0.374 & 0.492 \\
Rat & 0.419 & 0.487 & 0.612 & 0.611 & 0.458 & 0.504 & 0.380 & 0.621 & 0.454 & 0.289 & 0.377 & 0.478 & 0.442 & \cellcolor{gray!20}\textbf{0.413} & 0.402 & 0.505 \\
Robin & 0.427 & 0.550 & 0.650 & 0.611 & 0.447 & 0.519 & 0.419 & 0.613 & 0.474 & 0.321 & 0.398 & 0.484 & 0.454 & 0.462 & \cellcolor{gray!20}0.344 & 0.488 \\
Wolf & 0.517 & 0.625 & 0.745 & 0.688 & 0.515 & 0.686 & 0.518 & 0.703 & 0.608 & 0.668 & 0.473 & 0.610 & 0.579 & 0.611 & 0.499 & \cellcolor{gray!20}\textbf{0.453} \\
Hogtrunc2 & 0.743 & 0.673 & 0.635 & 0.606 & 0.520 & 0.742 & 0.660 & 0.630 & \cellcolor{gray!20}0.479 & 0.450 & 0.666 & 0.546 & 0.654 & 0.583 & 0.646 & 0.725 \\
Hogtrunc & 0.819 & 0.705 & 0.648 & 0.639 & 0.591 & 0.796 & 0.746 & 0.697 & \cellcolor{gray!20}0.502 & 0.462 & 0.712 & 0.584 & 0.702 & 0.615 & 0.725 & 0.749 \\
Moosetrunc2 & 0.680 & 0.739 & 0.748 & 0.723 & 0.573 & 0.880 & 0.668 & 0.712 & 0.528 & 0.493 & \cellcolor{gray!20}0.523 & 0.586 & 0.679 & 0.633 & 0.634 & 0.695 \\
Bear+Fox & \cellcolor{blue!10}\textbf{0.367} & 0.468 & \textbf{0.576} & \textbf{0.587} & 0.437 & 0.449 & \cellcolor{blue!10}0.329 & 0.580 & 0.447 & 0.311 & 0.363 & 0.467 & \textbf{0.400} & 0.414 & 0.351 & 0.483 \\
Bobcat+Moose & 0.492 & \cellcolor{blue!10}0.519 & 0.642 & 0.597 & 0.470 & 0.604 & 0.456 & 0.623 & 0.475 & 0.405 & \cellcolor{blue!10}0.364 & 0.503 & 0.510 & 0.501 & 0.441 & 0.513 \\
Bull+Cockatoo & 0.580 & 0.592 & \cellcolor{blue!10}0.604 & \cellcolor{blue!10}0.603 & 0.496 & 0.626 & 0.531 & 0.657 & 0.466 & 0.403 & 0.542 & 0.515 & 0.539 & 0.528 & 0.541 & 0.599 \\
Bull+Gator & 0.626 & 0.616 & \cellcolor{blue!10}0.602 & 0.618 & 0.506 & 0.668 & 0.566 & \cellcolor{blue!10}0.525 & 0.468 & 0.398 & 0.625 & 0.523 & 0.574 & 0.540 & 0.559 & 0.666 \\
Bull+Hog & 0.571 & 0.577 & \cellcolor{blue!10}0.598 & 0.618 & 0.487 & 0.633 & 0.510 & 0.637 & \cellcolor{blue!10}0.459 & 0.373 & 0.519 & 0.516 & 0.529 & 0.504 & 0.511 & 0.622 \\
Cockatoo+Hog & 0.687 & 0.661 & 0.650 & \cellcolor{blue!10}0.600 & 0.477 & 0.736 & 0.603 & 0.647 & \cellcolor{blue!10}0.474 & 0.440 & 0.632 & 0.518 & 0.627 & 0.559 & 0.586 & 0.695 \\
Crow+Robin & 0.422 & 0.530 & 0.662 & 0.594 & \cellcolor{blue!10}\textbf{0.423} & 0.517 & 0.413 & 0.592 & 0.459 & 0.331 & 0.393 & 0.465 & 0.449 & 0.458 & \cellcolor{blue!10}0.339 & 0.482 \\
Eagle+Robin & 0.628 & 0.610 & 0.639 & 0.595 & 0.498 & \cellcolor{blue!10}0.441 & 0.560 & 0.620 & 0.477 & 0.425 & 0.679 & 0.505 & 0.579 & 0.538 & \cellcolor{blue!10}0.502 & 0.688 \\
Hog+Moose & 0.526 & 0.566 & 0.632 & 0.603 & 0.465 & 0.719 & 0.499 & 0.626 & \cellcolor{blue!10}0.464 & 0.367 & \cellcolor{blue!10}0.402 & 0.491 & 0.527 & 0.497 & 0.485 & 0.528 \\
Lamb+Robin & 0.407 & 0.505 & 0.619 & 0.600 & 0.431 & 0.485 & 0.386 & 0.585 & 0.453 & \cellcolor{blue!10}\textbf{0.260} & 0.377 & 0.472 & 0.436 & 0.431 & \cellcolor{blue!10}\textbf{0.338} & 0.478 \\
Mouse+Rat & 0.414 & 0.485 & 0.614 & 0.610 & 0.448 & 0.492 & 0.378 & 0.608 & 0.464 & 0.293 & 0.378 & \cellcolor{blue!10}0.475 & 0.437 & \cellcolor{blue!10}0.414 & 0.397 & 0.504 \\
Peacock+Wolf & 0.437 & 0.549 & 0.669 & 0.615 & 0.454 & 0.516 & 0.418 & 0.637 & 0.507 & 0.433 & 0.413 & 0.492 & \cellcolor{blue!10}0.429 & 0.495 & 0.381 & \cellcolor{blue!10}0.455 \\
Bear+Foxtrunc & \cellcolor{blue!10}0.371 & \textbf{0.465} & 0.579 & 0.595 & 0.441 & 0.464 & \cellcolor{blue!10}0.336 & 0.588 & 0.457 & 0.328 & 0.373 & 0.468 & 0.401 & 0.414 & 0.360 & 0.489 \\
Bobcat+Moosetrunc & 0.496 & \cellcolor{blue!10}0.521 & 0.639 & 0.593 & 0.467 & 0.612 & 0.460 & 0.621 & 0.468 & 0.407 & \cellcolor{blue!10}0.370 & 0.500 & 0.523 & 0.504 & 0.448 & 0.516 \\
Bulltrunc+Gator & 0.610 & 0.609 & \cellcolor{blue!10}0.611 & 0.634 & 0.514 & 0.688 & 0.563 & \cellcolor{blue!10}0.524 & 0.482 & 0.383 & 0.629 & 0.513 & 0.562 & 0.532 & 0.552 & 0.637 \\
Eagletrunc+Robin & 0.628 & 0.610 & 0.639 & 0.595 & 0.498 & \cellcolor{blue!10}0.441 & 0.560 & 0.620 & 0.477 & 0.425 & 0.679 & 0.505 & 0.579 & 0.538 & \cellcolor{blue!10}0.502 & 0.688 \\
Hogtrunc2+Moosetrunc2 & 0.646 & 0.654 & 0.652 & 0.642 & 0.521 & 0.767 & 0.599 & 0.682 & \cellcolor{blue!10}0.489 & 0.403 & \cellcolor{blue!10}0.523 & 0.531 & 0.619 & 0.555 & 0.584 & 0.669 \\
Hogtrunc+Moosetrunc2 & 0.646 & 0.654 & 0.652 & 0.642 & 0.521 & 0.767 & 0.599 & 0.682 & \cellcolor{blue!10}0.489 & 0.403 & \cellcolor{blue!10}0.523 & 0.531 & 0.619 & 0.555 & 0.584 & 0.669 \\
Lamb+Robintrunc & 0.431 & 0.535 & 0.621 & 0.597 & 0.436 & 0.493 & 0.402 & 0.584 & \textbf{0.435} & \cellcolor{blue!10}0.304 & 0.384 & 0.471 & 0.474 & 0.452 & \cellcolor{blue!10}0.358 & 0.485 \\
Peacock+Wolftrunc & 0.442 & 0.570 & 0.688 & 0.639 & 0.473 & 0.526 & 0.424 & 0.660 & 0.536 & 0.465 & 0.436 & 0.514 & \cellcolor{blue!10}0.435 & 0.514 & 0.393 & \cellcolor{blue!10}0.462 \\
Bull+Cockatoo+Hog & 0.566 & 0.569 & \cellcolor{blue!10}0.604 & \cellcolor{blue!10}0.598 & 0.461 & 0.640 & 0.506 & 0.618 & \cellcolor{blue!10}0.454 & 0.367 & 0.513 & 0.489 & 0.520 & 0.495 & 0.509 & 0.604 \\
\bottomrule
\end{tabular}
\end{adjustbox}
\end{sidewaystable}

\paragraph{\textbf{Baseline results}}
Even in baseline experiments, a significant variation in the results can be observed. While datasets with more building data show better performance than those with a lesser number of buildings, it is difficult to say this observation always strictly holds. We believe this may be due to the quality of individual datasets.~\citet{YUAN2023126878} also mention that it is not possible to remove all the noise and errors only by pre-processing. We plan to further investigate this in the future.\\
\indent Result comparison between Hog, Hogtruncated1 and Hogtruncated2 suggests that if some data points are missing in the middle of the dataset, adding zero-padding to cover up for the missing data is beneficial.

\paragraph{\textbf{Zero-shot on individual dataset models (Strategy 1)}}
% Models trained on larger datasets (e.g., Fox with 127 buildings) generally performed\su{quantify} better in zero-shot scenarios compared to those trained on smaller datasets (e.g., Cockatoo, Crow, Mouse, and Moose with fewer than 10 buildings each). Some\su{quantify}  such models even outperformed their relevant baseline (e.g. results of Fox model on Bear, Bobcat, Bull, Crow, etc and Rat model on Crow and Hog for the 24-hour experiments).\\
% Another factor that has an impact on the zero-shot results is the availability of weather features. For example, Gator, which does not have any weather features shows the worst results for many datasets. Similarly, using a dataset from a similar climate zone has a positive impact. For example, Mouse that is similar to Crow except for the climate zone performs better for Rat and Robin than Crow. We believe this is due to Mouse, Rat and Robin being in the same climate zone.

Models trained on larger datasets (e.g., Fox with 127 buildings, Rat with 251 buildings) generally performed better in zero-shot scenarios compared to those trained on smaller datasets (e.g., Cockatoo with 1 building, Crow with 4 buildings, Mouse with 3 buildings, and Moose with 9 buildings), with datasets containing more than 70 buildings achieving an average MAE of 0.370 $\pm$ 0.075 compared to 0.497 $\pm$ 0.098 for datasets with fewer than 10 buildings, representing a 25.5\% improvement. Some models even outperformed their relevant baseline, with the Fox model showing improvements of 1--8.2\% over baseline for 8 different datasets (with the highest improvements for Bobcat at 8.2\% and Crow at 6.1\%), and the Rat model showing improvements of 0.6--6.7\% over baseline for 5 datasets (with the highest improvements for Crow at 6.7\% and Bobcat at 4.8\%) in the 24-hour experiments. These results align with observations of prior research, which also observed that higher amounts of source domain data are beneficial for TL~\cite{LI2022112461, FANG2023112968, LOPEZSANTOS2023113164}. The results also indicate that the difference between the weather features of source and target datasets has a large impact. For example, zero-shot results derived from the Gator model (the Gator dataset does not record any weather feature) show consistently worse results across datasets.

\paragraph{\textbf{Combined Dataset Models (Strategy 2)}}

Combined dataset models often outperformed individual dataset models in zero-shot scenarios by 15.9\% (average MAE of 0.367 vs 0.437), suggesting improved generalisation.  However, due to the observations discussed above, blindly combining multiple data sources will not provide better results over using a single source dataset. For example, Bull (separately) combined with Cockatoo, Gator and Hog shows dissimilar results - Bull+Cockatoo and Bull+Gator degrade performance by 0.2\% and 1.1\% respectively while Bull+Hog improves by 1.5\% compared to Bull's baseline MAE of 0.475.\\

The impact of the weather features and the dataset size is evident in these experiments as well. For example, when weather data of Wolf are removed to obtain Wolftruncated1 to match with the weather features of Peacock, the result of Peacock+Wolftruncated1 falls below the result of Peacock+Wolf (0.329 MAE). Similarly, when the building count of Eagle is reduced to obtain Eagletruncated1 to match with Robin, the result of Robin+Eagletruncated1 falls below Robin+Eagle (0.254 MAE).

\paragraph{\textbf{Comparison of 24-hour vs. 96-hour Forecasts}}
Our analysis reveals a consistent degradation in forecast accuracy when moving from 24-hour to 96-hour predictions. Looking at average MAE values across different model configurations: models related to Strategy 2 showed the most resilient performance, with mean MAE increasing from 0.367 to 0.463 (a 26.0\% degradation); models related to Strategy 1 showed slightly higher degradation, with mean MAE increasing from 0.430 to 0.554 (a 28.6\% increase). This systematic degradation in longer-horizon forecasts aligns with prior observations that shorter-term forecasting is typically more accurate in time series prediction~\cite{li2024improved}.

\subsection{Results Comparison of Strategies 5-8}
Table~\ref{tab:small_Scale_summary_transformer_mae} shows the results for models trained by combining 2 or 3 datasets.  Table~\ref{tab:large_scale_summary_transformer_mae} reports results for Strategies 6 and 8 considering the ensemble model trained with all the datasets. Note that in both these tables, the first six rows refer to strategies 6 and 8. All the other results refer to strategies 5 and 7. In both the tables, first `Imp' column shows the improvement of strategy 5/6 over baseline. Second `Imp' column shows the improvement of strategy 7/8 over the baseline.

\subsubsection{Strategy 5 and 6}
We make the following observations from our results for Strategy 5:
\begin{itemize}
    \item \textbf{Temporal alignment across the combined datasets:} Combining  two datasets representing different time periods hurt both datasets (e.g. Hogtruncated2+Moosetruncated2). However, zero-padding the datasets to align the temporal ranges of data results in positive gains for both datasets (Hogtruncated1+Moosetruncated2).

    \sloppy
    \item \textbf{Weather Feature Consistency:} When the recorded weather features have a large deviation, results can degrade. For example, Bull (with AT, DT, SLP) combined with Gator (no weather features) shows a degradation in results for both datasets. Similarly, despite having the same climate zone, the differing weather features of Peacock (with AT, DT, SLP) and Wolf (with AT, DT, SLP, WD, WS, CC) lead to a results degradation in the Peacock+Wolf dataset for Peacock. Artificially reducing the weather features to align the feature 
    spaces of the two datasets does not help either of the datasets and results are usually below the baseline (e.g. Lamb+Robin\-truncated, 
    Bulltruncated+Gator, Peacock+Wolf\-truncated).~\citet{XING2024123276} also reported that weather feature mismatch can lead to results degradation. 
    \fussy

    \item \textbf{Building count differences: }When combining datasets of drastically different sizes, the combined model underperforms or shows minimal gains against the baseline for the larger dataset. This observation is similar to the \textit{negative interference} experienced by high-resource languages when trained with low-resource languages in a multilingual setup~\cite{arivazhagan2019massively}. However, the smaller dataset gets a higher benefit (e.g. Mouse+Rat, Crow+Robin, Bear+Fox).% The same observation has been made by some past research as well~\cite{li2024comprehensive, YAN2023108127, FAN2020114499}

    \item \textbf{Climate Zone Considerations:} Combining datasets from different climate zones does not necessarily harm performance if other factors (i.e.~weather features and building counts) align. For example, despite Bobcat and Moose being from different climate zones (5B and 6A respectively), Bobcat+Moose show improvements over the baseline for both datasets. Similar observations hold for Bobcat+Moosetruncated1 and Lamb+Robin. We believe this is an important finding, because the previous research that claimed the high impact of climate zone has not considered the variation of other features~\cite{FANG2023112968, Gao, AHN2022111717}.

\end{itemize}
% As for Strategy 6, even combining three datasets does not provide consistent gains over the baseline.  For example, while Bull+Cockatoo+Hog model matches the Bull 96hr baseline (both 0.602), the full ensemble model actually performs better (0.565)\su{can't understand how this sentence tallies with the previous one}. Training an ensemble model with all available datasets outperforms baseline for 87.5\% of datasets in 24-hour forecasting, though this drops to just 43.8\% for 96-hour forecasting. When compared to smaller combinations, the full ensemble performs better in 75.0\% of cases for 24-hour and 50.0\% for 96-hour forecasts. This suggests that larger ensembles are generally better, as evidenced by previous research as well~\cite{LU2023109024}.  Performance of Bear+Fox is an example where the full ensemble model fails to derive the best result - Bear + Fox performs equally well as the full ensemble model for Bear in 24-hour forecasts (both 0.291) and better in 96-hour forecasts (0.367 vs 0.393). Moreover, for datasets with very different weather feature information like Gator (no weather features), the full ensemble model underperforms compared to the baseline at both horizons (0.323 vs 0.292 for 24hr, 0.529 vs 0.503 for 96hr). \

As for Strategy 6, combining three datasets provides mixed results compared to the baseline. For example, the Bull+Cockatoo+Hog model merely matches the Bull 96hr baseline (both 0.602). However, scaling up to include more datasets shows promise. Training an ensemble model with all available datasets outperforms baseline for 87.5\% of datasets in 24-hour forecasting, though this drops to just 43.8\% for 96-hour forecasting. When compared to smaller combinations, the full ensemble performs better in 75.0\% of cases for 24-hour and 50.0\% for 96-hour forecasts. This suggests that larger ensembles are generally better, as evidenced by previous research as well~\cite{LU2023109024}. However, there are exceptions - Bear+Fox is an example where the full ensemble model fails to derive the best result - Bear+Fox performs equally well as the full ensemble model for Bear in 24-hour forecasts (both 0.291) and better in 96-hour forecasts (0.367 vs 0.393). Moreover, for datasets with very different weather feature information like Gator (no weather features), the full ensemble model underperforms compared to the baseline at both horizons (0.323 vs 0.292 for 24hr, 0.529 vs 0.503 for 96hr). \\

\subsubsection{Strategy 7 and 8}
% After training a target dataset with source dataset(s), further fine-tuning that model with the target data is beneficial in most \su{quantify} of the cases. However, when there are discrepencies among the target and source datasets, this gain can be less. In particular, when the feature spaces differ with respect to weather features between two datasets, even further fine-tuning is not enough to beat the baseline (e.g.~Bulltruncated+Gator further fine-tuned on Gator result is still below the baseline). On the other hand, further fine-tuning the ensemble model trained with all the datasets with individual datasets always beat the baseline. However, in some\su{quantify}  experiments, further fine-tuning result falls below the result of Strategy 6, or just on-par (e.g. Bull 96hr, Cockatoo 24hr).  Such results question the utility of Strategy 8, given the fine-tuning overhead. 

Under Strategy 7, after training a target dataset with a source dataset, further fine-tuning that model with the target data is beneficial in  85.2\% of 24-hour cases and 77.8\% of 96-hour cases. However, when there are discrepancies among the target and source datasets, this gain can be less. In particular, when the feature spaces differ with respect to weather features between two datasets, even further fine-tuning is not enough to beat the baseline (e.g. Bulltruncated+Gator case).  On the other hand, further fine-tuning the ensemble model trained with all the datasets with individual datasets (Strategy 8) always beats the baseline. However, in 18.8\% of experiments (for both 24h and 96h horizons), further fine-tuning result falls below the result of Strategy 6, or just on-par (e.g. Bull 96hr: 0.571 vs 0.565, Cockatoo 24hr: both 0.432). Such results question the utility of Strategy 8, given the fine-tuning overhead. 

\subsection{Comparison of Different Transformer Models}
\begin{figure}[H]
    \centering
    \includegraphics[width=1\linewidth]{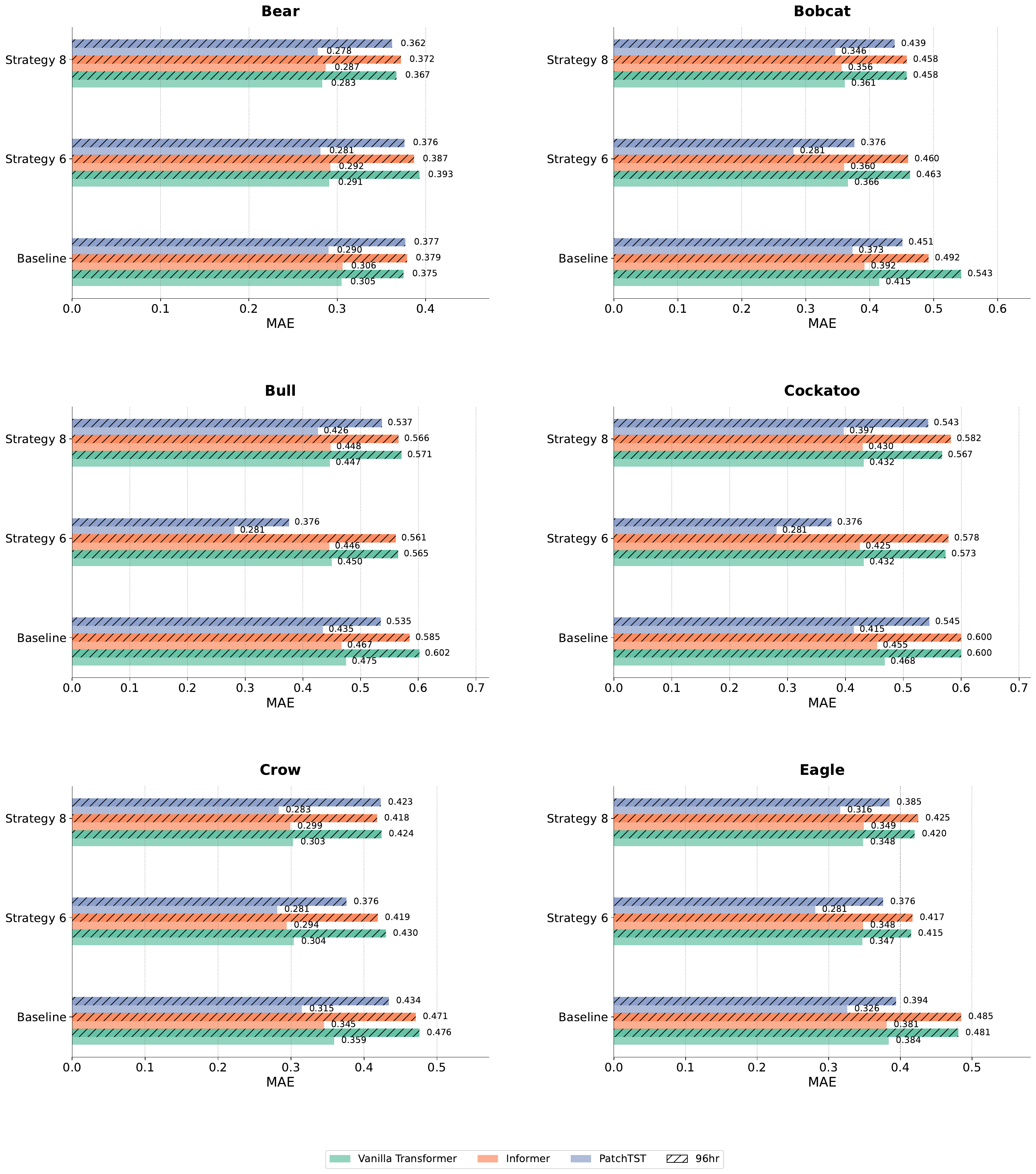}
    \caption{MAE Comparison for the Ensemble Model Across Different Transformer Architectures for the Following Datasets: Bear, Bobcat, Bull, Cockatoo, Crow and Eagle. }
    \label{fig:enter-label-1}
\end{figure}

The result comparison of models trained using all the datasets on vanilla Transformer, Informer and PatchTST are shown in Figure~\ref{fig:enter-label-1} for the following datasets:  Bear, Bobcat, Bull, Cockatoo, Crow and Eagle. Graphical results for other datasets, as well as the tabular version of the results, are in the Appendix. Results of Informer and PatchTST confirm the general trend we observed in the context of vanilla Transformers - results of Strategy 8 are the best, followed by those of Strategy 6. We also note that PatchTST overall shows the best performance. On the other hand, Informer results are on par with that of vanilla Transformer, and sometimes even worse than that of vanilla Transformer.

\newgeometry{bottom=2.6cm}  % Adjust the bottom margin
\begin{table}[p]
\caption{Performance Summary (reported in MAE) for two- and three-dataset combinations \textit{(Vanilla Transformer)}  }
\vspace{0.2cm}
\centering
\begin{adjustbox}{width=\textwidth,center}
\tiny
\label{tab:small_Scale_summary_transformer_mae}
\begin{tabular}{llllrrrrr}
\toprule
Combo & Test & Hor & Baseline & Strategy 5/6 & Imp & Strategy 7/8 & Imp \\
 & Set & & MAE & MAE & (\%) & MAE & (\%) \\
\midrule
Bull+Cockatoo+Hog & Bull & 24hr & 0.475 & 0.469 & \cellcolorgrad{1}\textcolor{black}{1.3\%} & 0.466 & \cellcolorgrad{2}\textcolor{black}{1.9\%} \\
Bull+Cockatoo+Hog & Bull & 96hr & 0.602 & 0.604 & \cellcolorgrad{0}\textcolor{black}{-0.3\%} & 0.602 & \cellcolorgrad{0}\textcolor{black}{0.0\%} \\
Bull+Cockatoo+Hog & Cockatoo & 24hr & 0.468 & 0.455 & \cellcolorgrad{3}\textcolor{black}{2.8\%} & 0.445 & \cellcolorgrad{5}\textcolor{black}{4.9\%} \\
Bull+Cockatoo+Hog & Cockatoo & 96hr & 0.600 & 0.598 & \cellcolorgrad{0}\textcolor{black}{0.3\%} & 0.587 & \cellcolorgrad{2}\textcolor{black}{2.2\%} \\
Bull+Cockatoo+Hog & Hog & 24hr & 0.336 & 0.322 & \cellcolorgrad{4}\textcolor{black}{4.2\%} & 0.318 & \cellcolorgrad{5}\textcolor{black}{5.4\%} \\
Bull+Cockatoo+Hog & Hog & 96hr & 0.470 & 0.454 & \cellcolorgrad{3}\textcolor{black}{3.4\%} & 0.458 & \cellcolorgrad{3}\textcolor{black}{2.6\%} \\
Bear+Fox & Bear & 24hr & 0.305 & 0.291 & \cellcolorgrad{5}\textcolor{black}{4.6\%} & 0.286 & \cellcolorgrad{6}\textcolor{black}{6.2\%} \\
Bear+Fox & Bear & 96hr & 0.375 & 0.367 & \cellcolorgrad{2}\textcolor{black}{2.1\%} & 0.363 & \cellcolorgrad{3}\textcolor{black}{3.2\%} \\
Bear+Fox & Fox & 24hr & 0.268 & 0.264 & \cellcolorgrad{1}\textcolor{black}{1.5\%} & 0.257 & \cellcolorgrad{4}\textcolor{black}{4.1\%} \\
Bear+Fox & Fox & 96hr & 0.325 & 0.329 & \cellcolorgrad{-1}\textcolor{black}{-1.2\%} & 0.320 & \cellcolorgrad{2}\textcolor{black}{1.5\%} \\
Bobcat+Moose & Bobcat & 24hr & 0.415 & 0.399 & \cellcolorgrad{4}\textcolor{black}{3.9\%} & 0.393 & \cellcolorgrad{5}\textcolor{black}{5.3\%} \\
Bobcat+Moose & Bobcat & 96hr & 0.543 & 0.519 & \cellcolorgrad{4}\textcolor{black}{4.4\%} & 0.507 & \cellcolorgrad{7}\textcolor{black}{6.6\%} \\
Bobcat+Moose & Moose & 24hr & 0.294 & 0.278 & \cellcolorgrad{5}\textcolor{black}{5.4\%} & 0.271 & \cellcolorgrad{8}\textcolor{black}{7.8\%} \\
Bobcat+Moose & Moose & 96hr & 0.377 & 0.364 & \cellcolorgrad{3}\textcolor{black}{3.4\%} & 0.355 & \cellcolorgrad{6}\textcolor{black}{5.8\%} \\
Bull+Gator & Bull & 24hr & 0.475 & 0.480 & \cellcolorgrad{-1}\textcolor{black}{-1.1\%} & 0.467 & \cellcolorgrad{2}\textcolor{black}{1.7\%} \\
Bull+Gator & Bull & 96hr & 0.602 & 0.602 & \cellcolorgrad{0}\textcolor{black}{0.0\%} & 0.596 & \cellcolorgrad{1}\textcolor{black}{1.0\%} \\
Bull+Gator & Gator & 24hr & 0.292 & 0.312 & \cellcolorgrad{-7}\textcolor{black}{-6.8\%} & 0.289 & \cellcolorgrad{1}\textcolor{black}{1.0\%} \\
Bull+Gator & Gator & 96hr & 0.503 & 0.525 & \cellcolorgrad{-4}\textcolor{black}{-4.4\%} & 0.512 & \cellcolorgrad{-2}\textcolor{black}{-1.8\%} \\
Crow+Robin & Crow & 24hr & 0.359 & 0.313 & \cellcolorgrad{13}\textcolor{black}{12.8\%} & 0.311 & \cellcolorgrad{13}\textcolor{black}{13.4\%} \\
Crow+Robin & Crow & 96hr & 0.476 & 0.423 & \cellcolorgrad{11}\textcolor{black}{11.1\%} & 0.424 & \cellcolorgrad{11}\textcolor{black}{10.9\%} \\
Crow+Robin & Robin & 24hr & 0.254 & 0.255 & \cellcolorgrad{0}\textcolor{black}{-0.4\%} & 0.253 & \cellcolorgrad{0}\textcolor{black}{0.4\%} \\
Crow+Robin & Robin & 96hr & 0.344 & 0.339 & \cellcolorgrad{1}\textcolor{black}{1.5\%} & 0.335 & \cellcolorgrad{3}\textcolor{black}{2.6\%} \\
Hog+Moose & Hog & 24hr & 0.336 & 0.323 & \cellcolorgrad{4}\textcolor{black}{3.9\%} & 0.322 & \cellcolorgrad{4}\textcolor{black}{4.2\%} \\
Hog+Moose & Hog & 96hr & 0.470 & 0.464 & \cellcolorgrad{1}\textcolor{black}{1.3\%} & 0.463 & \cellcolorgrad{1}\textcolor{black}{1.5\%} \\
Hog+Moose & Moose & 24hr & 0.294 & 0.297 & \cellcolorgrad{-1}\textcolor{black}{-1.0\%} & 0.280 & \cellcolorgrad{5}\textcolor{black}{4.8\%} \\
Hog+Moose & Moose & 96hr & 0.377 & 0.402 & \cellcolorgrad{-7}\textcolor{black}{-6.6\%} & 0.367 & \cellcolorgrad{3}\textcolor{black}{2.7\%} \\
Lamb+Robin & Lamb & 24hr & 0.208 & 0.185 & \cellcolorgrad{11}\textcolor{black}{11.1\%} & 0.191 & \cellcolorgrad{8}\textcolor{black}{8.2\%} \\
Lamb+Robin & Lamb & 96hr & 0.269 & 0.260 & \cellcolorgrad{3}\textcolor{black}{3.3\%} & 0.251 & \cellcolorgrad{7}\textcolor{black}{6.7\%} \\
Lamb+Robin & Robin & 24hr & 0.254 & 0.248 & \cellcolorgrad{2}\textcolor{black}{2.4\%} & 0.247 & \cellcolorgrad{3}\textcolor{black}{2.8\%} \\
Lamb+Robin & Robin & 96hr & 0.344 & 0.338 & \cellcolorgrad{2}\textcolor{black}{1.7\%} & 0.337 & \cellcolorgrad{2}\textcolor{black}{2.0\%} \\
Mouse+Rat & Mouse & 24hr & 0.365 & 0.348 & \cellcolorgrad{5}\textcolor{black}{4.7\%} & 0.338 & \cellcolorgrad{7}\textcolor{black}{7.4\%} \\
Mouse+Rat & Mouse & 96hr & 0.532 & 0.475 & \cellcolorgrad{11}\textcolor{black}{10.7\%} & 0.462 & \cellcolorgrad{13}\textcolor{black}{13.2\%} \\
Mouse+Rat & Rat & 24hr & 0.293 & 0.294 & \cellcolorgrad{0}\textcolor{black}{-0.3\%} & 0.294 & \cellcolorgrad{0}\textcolor{black}{-0.3\%} \\
Mouse+Rat & Rat & 96hr & 0.413 & 0.414 & \cellcolorgrad{0}\textcolor{black}{-0.2\%} & 0.413 & \cellcolorgrad{0}\textcolor{black}{0.0\%} \\
Peacock+Wolftruncated1 & Peacock & 24hr & 0.316 & 0.329 & \cellcolorgrad{-4}\textcolor{black}{-4.1\%} & 0.320 & \cellcolorgrad{-1}\textcolor{black}{-1.3\%} \\
Peacock+Wolftruncated1 & Peacock & 96hr & 0.411 & 0.435 & \cellcolorgrad{-6}\textcolor{black}{-5.8\%} & 0.414 & \cellcolorgrad{-1}\textcolor{black}{-0.7\%} \\
Peacock+Wolftruncated1 & Wolf & 24hr & 0.387 & 0.376 & \cellcolorgrad{3}\textcolor{black}{2.8\%} & 0.374 & \cellcolorgrad{3}\textcolor{black}{3.4\%} \\
Peacock+Wolftruncated1 & Wolf & 96hr & 0.453 & 0.462 & \cellcolorgrad{-2}\textcolor{black}{-2.0\%} & 0.456 & \cellcolorgrad{-1}\textcolor{black}{-0.7\%} \\
Eagletruncated1+Robin & Eagle & 24hr & 0.384 & 0.368 & \cellcolorgrad{4}\textcolor{black}{4.2\%} & 0.363 & \cellcolorgrad{5}\textcolor{black}{5.5\%} \\
Eagletruncated1+Robin & Eagle & 96hr & 0.481 & 0.428 & \cellcolorgrad{11}\textcolor{black}{11.0\%} & 0.427 & \cellcolorgrad{11}\textcolor{black}{11.2\%} \\
Eagletruncated1+Robin & Robin & 24hr & 0.254 & 0.368 & \cellcolorgrad{-45}\textcolor{black}{-44.9\%} & 0.254 & \cellcolorgrad{0}\textcolor{black}{0.0\%} \\
Eagletruncated1+Robin & Robin & 96hr & 0.344 & 0.502 & \cellcolorgrad{-46}\textcolor{black}{-45.9\%} & 0.336 & \cellcolorgrad{2}\textcolor{black}{2.3\%} \\
Bear+Foxtruncated1 & Bear & 24hr & 0.305 & 0.295 & \cellcolorgrad{3}\textcolor{black}{3.3\%} & 0.289 & \cellcolorgrad{5}\textcolor{black}{5.2\%} \\
Bear+Foxtruncated1 & Bear & 96hr & 0.375 & 0.371 & \cellcolorgrad{1}\textcolor{black}{1.1\%} & 0.363 & \cellcolorgrad{3}\textcolor{black}{3.2\%} \\
Bear+Foxtruncated1 & Foxtruncated1 & 24hr & 0.268 & 0.273 & \cellcolorgrad{-2}\textcolor{black}{-1.9\%} & 0.259 & \cellcolorgrad{3}\textcolor{black}{3.4\%} \\
Bear+Foxtruncated1 & Foxtruncated1 & 96hr & 0.325 & 0.336 & \cellcolorgrad{-3}\textcolor{black}{-3.4\%} & 0.320 & \cellcolorgrad{2}\textcolor{black}{1.5\%} \\
Bobcat+Moosetruncated1 & Bobcat & 24hr & 0.415 & 0.400 & \cellcolorgrad{4}\textcolor{black}{3.6\%} & 0.398 & \cellcolorgrad{4}\textcolor{black}{4.1\%} \\
Bobcat+Moosetruncated1 & Bobcat & 96hr & 0.543 & 0.521 & \cellcolorgrad{4}\textcolor{black}{4.1\%} & 0.512 & \cellcolorgrad{6}\textcolor{black}{5.7\%} \\
Bobcat+Moosetruncated1 & Moose & 24hr & 0.294 & 0.282 & \cellcolorgrad{4}\textcolor{black}{4.1\%} & 0.274 & \cellcolorgrad{7}\textcolor{black}{6.8\%} \\
Bobcat+Moosetruncated1 & Moose & 96hr & 0.377 & 0.370 & \cellcolorgrad{2}\textcolor{black}{1.9\%} & 0.356 & \cellcolorgrad{6}\textcolor{black}{5.6\%} \\
Eagle+Robin & Eagle & 24hr & 0.384 & 0.368 & \cellcolorgrad{4}\textcolor{black}{4.2\%} & 0.363 & \cellcolorgrad{5}\textcolor{black}{5.5\%} \\
Eagle+Robin & Eagle & 96hr & 0.481 & 0.428 & \cellcolorgrad{11}\textcolor{black}{11.0\%} & 0.427 & \cellcolorgrad{11}\textcolor{black}{11.2\%} \\
Eagle+Robin & Robin & 24hr & 0.254 & 0.368 & \cellcolorgrad{-45}\textcolor{black}{-44.9\%} & 0.254 & \cellcolorgrad{0}\textcolor{black}{0.0\%} \\
Eagle+Robin & Robin & 96hr & 0.344 & 0.502 & \cellcolorgrad{-46}\textcolor{black}{-45.9\%} & 0.336 & \cellcolorgrad{2}\textcolor{black}{2.3\%} \\
Bulltruncated1+Gator & Bull & 24hr & 0.475 & 0.483 & \cellcolorgrad{-2}\textcolor{black}{-1.7\%} & 0.466 & \cellcolorgrad{2}\textcolor{black}{1.9\%} \\
Bulltruncated1+Gator & Bull & 96hr & 0.602 & 0.611 & \cellcolorgrad{-1}\textcolor{black}{-1.5\%} & 0.591 & \cellcolorgrad{2}\textcolor{black}{1.8\%} \\
Bulltruncated1+Gator & Gator & 24hr & 0.292 & 0.308 & \cellcolorgrad{-5}\textcolor{black}{-5.5\%} & 0.287 & \cellcolorgrad{2}\textcolor{black}{1.7\%} \\
Bulltruncated1+Gator & Gator & 96hr & 0.503 & 0.524 & \cellcolorgrad{-4}\textcolor{black}{-4.2\%} & 0.512 & \cellcolorgrad{-2}\textcolor{black}{-1.8\%} \\
Peacock+Wolf & Peacock & 24hr & 0.316 & 0.329 & \cellcolorgrad{-4}\textcolor{black}{-4.1\%} & 0.320 & \cellcolorgrad{-1}\textcolor{black}{-1.3\%} \\
Peacock+Wolf & Peacock & 96hr & 0.411 & 0.429 & \cellcolorgrad{-4}\textcolor{black}{-4.4\%} & 0.412 & \cellcolorgrad{0}\textcolor{black}{-0.2\%} \\
Peacock+Wolf & Wolf & 24hr & 0.387 & 0.379 & \cellcolorgrad{2}\textcolor{black}{2.1\%} & 0.381 & \cellcolorgrad{2}\textcolor{black}{1.6\%} \\
Peacock+Wolf & Wolf & 96hr & 0.453 & 0.455 & \cellcolorgrad{0}\textcolor{black}{-0.4\%} & 0.458 & \cellcolorgrad{-1}\textcolor{black}{-1.1\%} \\
Lamb+Robintruncated1 & Lamb & 24hr & 0.208 & 0.218 & \cellcolorgrad{-5}\textcolor{black}{-4.8\%} & 0.197 & \cellcolorgrad{5}\textcolor{black}{5.3\%} \\
Lamb+Robintruncated1 & Lamb & 96hr & 0.269 & 0.304 & \cellcolorgrad{-13}\textcolor{black}{-13.0\%} & 0.273 & \cellcolorgrad{-1}\textcolor{black}{-1.5\%} \\
Lamb+Robintruncated1 & Robintruncated1 & 24hr & 0.254 & 0.271 & \cellcolorgrad{-7}\textcolor{black}{-6.7\%} & 0.252 & \cellcolorgrad{1}\textcolor{black}{0.8\%} \\
Lamb+Robintruncated1 & Robintruncated1 & 96hr & 0.344 & 0.358 & \cellcolorgrad{-4}\textcolor{black}{-4.1\%} & 0.338 & \cellcolorgrad{2}\textcolor{black}{1.7\%} \\
Hogtruncated1+Moosetruncated2 & Hogtruncated1 & 24hr & 0.417 & 0.364 & \cellcolorgrad{13}\textcolor{black}{12.7\%} & 0.367 & \cellcolorgrad{12}\textcolor{black}{12.0\%} \\
Hogtruncated1+Moosetruncated2 & Hogtruncated1 & 96hr & 0.606 & 0.485 & \cellcolorgrad{20}\textcolor{black}{20.0\%} & 0.492 & \cellcolorgrad{19}\textcolor{black}{18.8\%} \\
Hogtruncated1+Moosetruncated2 & Moosetruncated2 & 24hr & 0.254 & 0.242 & \cellcolorgrad{5}\textcolor{black}{4.7\%} & 0.239 & \cellcolorgrad{6}\textcolor{black}{5.9\%} \\
Hogtruncated1+Moosetruncated2 & Moosetruncated2 & 96hr & 0.329 & 0.331 & \cellcolorgrad{-1}\textcolor{black}{-0.6\%} & 0.327 & \cellcolorgrad{1}\textcolor{black}{0.6\%} \\
Hogtruncated2+Moosetruncated2 & Hogtruncated2 & 24hr & 0.387 & 0.429 & \cellcolorgrad{-11}\textcolor{black}{-10.9\%} & 0.390 & \cellcolorgrad{-1}\textcolor{black}{-0.8\%} \\
Hogtruncated2+Moosetruncated2 & Hogtruncated2 & 96hr & 0.539 & 0.562 & \cellcolorgrad{-4}\textcolor{black}{-4.3\%} & 0.520 & \cellcolorgrad{4}\textcolor{black}{3.5\%} \\
Hogtruncated2+Moosetruncated2 & Moosetruncated2 & 24hr & 0.254 & 0.242 & \cellcolorgrad{5}\textcolor{black}{4.7\%} & 0.239 & \cellcolorgrad{6}\textcolor{black}{5.9\%} \\
Hogtruncated2+Moosetruncated2 & Moosetruncated2 & 96hr & 0.329 & 0.331 & \cellcolorgrad{-1}\textcolor{black}{-0.6\%} & 0.327 & \cellcolorgrad{1}\textcolor{black}{0.6\%} \\
\bottomrule
\end{tabular}
\end{adjustbox}
\end{table}
\restoregeometry

\begin{table}[p]
\caption{Performance Summary (reported in MAE) for the model trained with all the datasets\\ \textit{(Vanilla Transformer)}}
\vspace{0.2cm}
\centering
\begin{adjustbox}{width=\textwidth,center}
\tiny
\label{tab:large_scale_summary_transformer_mae}
\begin{tabular}{llrrrrrr}
\toprule
Dataset & Horizon & Baseline & Strategy 6 & Imp & Strategy 8 & Imp \\
 & & MAE & MAE & (\%) & MAE & (\%) \\
\midrule
Bear & 24hr & 0.305 & 0.291 & \cellcolorgrad{5}\textcolor{black}{+4.6\%} & 0.283 & \cellcolorgrad{7}\textcolor{black}{+7.2\%} \\
Bear & 96hr & 0.375 & 0.393 & \cellcolorgrad{-5}\textcolor{black}{-4.8\%} & 0.367 & \cellcolorgrad{2}\textcolor{black}{+2.1\%} \\
Bobcat & 24hr & 0.415 & 0.366 & \cellcolorgrad{12}\textcolor{black}{+11.8\%} & 0.361 & \cellcolorgrad{13}\textcolor{black}{+13.0\%} \\
Bobcat & 96hr & 0.543 & 0.463 & \cellcolorgrad{15}\textcolor{black}{+14.7\%} & 0.458 & \cellcolorgrad{16}\textcolor{black}{+15.7\%} \\
Bull & 24hr & 0.475 & 0.450 & \cellcolorgrad{5}\textcolor{black}{+5.3\%} & 0.447 & \cellcolorgrad{6}\textcolor{black}{+5.9\%} \\
Bull & 96hr & 0.602 & 0.565 & \cellcolorgrad{6}\textcolor{black}{+6.1\%} & 0.571 & \cellcolorgrad{5}\textcolor{black}{+5.1\%} \\
Cockatoo & 24hr & 0.468 & 0.432 & \cellcolorgrad{8}\textcolor{black}{+7.7\%} & 0.432 & \cellcolorgrad{8}\textcolor{black}{+7.7\%} \\
Cockatoo & 96hr & 0.600 & 0.573 & \cellcolorgrad{5}\textcolor{black}{+4.5\%} & 0.567 & \cellcolorgrad{6}\textcolor{black}{+5.5\%} \\
Crow & 24hr & 0.359 & 0.304 & \cellcolorgrad{15}\textcolor{black}{+15.3\%} & 0.303 & \cellcolorgrad{16}\textcolor{black}{+15.6\%} \\
Crow & 96hr & 0.476 & 0.430 & \cellcolorgrad{10}\textcolor{black}{+9.7\%} & 0.424 & \cellcolorgrad{11}\textcolor{black}{+10.9\%} \\
Eagle & 24hr & 0.384 & 0.347 & \cellcolorgrad{10}\textcolor{black}{+9.6\%} & 0.348 & \cellcolorgrad{9}\textcolor{black}{+9.4\%} \\
Eagle & 96hr & 0.481 & 0.415 & \cellcolorgrad{14}\textcolor{black}{+13.7\%} & 0.420 & \cellcolorgrad{13}\textcolor{black}{+12.7\%} \\
Fox & 24hr & 0.268 & 0.263 & \cellcolorgrad{2}\textcolor{black}{+1.9\%} & 0.254 & \cellcolorgrad{5}\textcolor{black}{+5.2\%} \\
Fox & 96hr & 0.325 & 0.351 & \cellcolorgrad{-8}\textcolor{black}{-8.0\%} & 0.321 & \cellcolorgrad{1}\textcolor{black}{+1.2\%} \\
Gator & 24hr & 0.292 & 0.323 & \cellcolorgrad{-11}\textcolor{black}{-10.6\%} & 0.288 & \cellcolorgrad{1}\textcolor{black}{+1.4\%} \\
Gator & 96hr & 0.503 & 0.529 & \cellcolorgrad{-5}\textcolor{black}{-5.2\%} & 0.491 & \cellcolorgrad{2}\textcolor{black}{+2.4\%} \\
Hog & 24hr & 0.336 & 0.296 & \cellcolorgrad{12}\textcolor{black}{+11.9\%} & 0.294 & \cellcolorgrad{13}\textcolor{black}{+12.5\%} \\
Hog & 96hr & 0.470 & 0.414 & \cellcolorgrad{12}\textcolor{black}{+11.9\%} & 0.427 & \cellcolorgrad{9}\textcolor{black}{+9.1\%} \\
Lamb & 24hr & 0.208 & 0.182 & \cellcolorgrad{13}\textcolor{black}{+12.5\%} & 0.181 & \cellcolorgrad{13}\textcolor{black}{+13.0\%} \\
Lamb & 96hr & 0.269 & 0.278 & \cellcolorgrad{-3}\textcolor{black}{-3.3\%} & 0.251 & \cellcolorgrad{7}\textcolor{black}{+6.7\%} \\
Moose & 24hr & 0.294 & 0.265 & \cellcolorgrad{10}\textcolor{black}{+9.9\%} & 0.255 & \cellcolorgrad{13}\textcolor{black}{+13.3\%} \\
Moose & 96hr & 0.377 & 0.378 & \cellcolorgrad{0}\textcolor{black}{-0.3\%} & 0.342 & \cellcolorgrad{9}\textcolor{black}{+9.3\%} \\
Mouse & 24hr & 0.365 & 0.329 & \cellcolorgrad{10}\textcolor{black}{+9.9\%} & 0.325 & \cellcolorgrad{11}\textcolor{black}{+11.0\%} \\
Mouse & 96hr & 0.532 & 0.447 & \cellcolorgrad{16}\textcolor{black}{+16.0\%} & 0.441 & \cellcolorgrad{17}\textcolor{black}{+17.1\%} \\
Peacock & 24hr & 0.316 & 0.312 & \cellcolorgrad{1}\textcolor{black}{+1.3\%} & 0.303 & \cellcolorgrad{4}\textcolor{black}{+4.1\%} \\
Peacock & 96hr & 0.411 & 0.412 & \cellcolorgrad{0}\textcolor{black}{-0.2\%} & 0.388 & \cellcolorgrad{6}\textcolor{black}{+5.6\%} \\
Rat & 24hr & 0.293 & 0.280 & \cellcolorgrad{4}\textcolor{black}{+4.4\%} & 0.282 & \cellcolorgrad{4}\textcolor{black}{+3.8\%} \\
Rat & 96hr & 0.413 & 0.401 & \cellcolorgrad{3}\textcolor{black}{+2.9\%} & 0.394 & \cellcolorgrad{5}\textcolor{black}{+4.6\%} \\
Robin & 24hr & 0.254 & 0.247 & \cellcolorgrad{3}\textcolor{black}{+2.8\%} & 0.238 & \cellcolorgrad{6}\textcolor{black}{+6.3\%} \\
Robin & 96hr & 0.344 & 0.359 & \cellcolorgrad{-4}\textcolor{black}{-4.4\%} & 0.324 & \cellcolorgrad{6}\textcolor{black}{+5.8\%} \\
Wolf & 24hr & 0.387 & 0.379 & \cellcolorgrad{2}\textcolor{black}{+2.1\%} & 0.363 & \cellcolorgrad{6}\textcolor{black}{+6.2\%} \\
Wolf & 96hr & 0.453 & 0.472 & \cellcolorgrad{-4}\textcolor{black}{-4.2\%} & 0.448 & \cellcolorgrad{1}\textcolor{black}{+1.1\%} \\
\bottomrule
\end{tabular}
\end{adjustbox}
\end{table}

\newgeometry{bottom=4cm}  % Adjust the bottom margin
\restoregeometry

\section{Discussion}
\subsection{Answering the Research Questions}
With the insights derived from the results reported in Section~\ref{sec:results}, we now revisit the three research questions. 

\textit{RQ1: How do different data-centric TL strategies compare in performance when applied to the vanilla Transformer architecture for building energy consumption forecasting?}

If the target domain has no data, zero-shot TL can be employed. If multiple source datasets are available, training a single source model with all these data is generally more beneficial than using a single source. In other words, Strategy 2 should be favoured over Strategy 1. {We believe this is an important finding, given that none of the prior work we referred to experimented with Strategy 2.}

Similarly, if there is some target data available, training a model using that data along with the available source dataset(s), and then further fine-tuning with the target dataset produces the optimal result most of the time. In other words, strategies 6 and 8 are more beneficial than strategies 5 and 7 (respectively).  Similar to the observations in zero-shot TL, strategy 7 is generally better than strategy 5, and strategy 8 is generally better than strategy 6. In other words, using all the available source datasets to train a model is beneficial over relying on individual datasets. We believe this is an important finding, given that no related research used Strategy 7, and Strategy 8 was used less frequently than Strategy 6 (see Table~\ref{table:tl_techniques_summary}).  

Another decision to make is how many source datasets to use for training the initial model. While more datasets would be better, this causes a computational overhead. ~\citet{WEI2024108713} concluded that three source domains is the best. However, blindly combining the available source datasets can harm the target, and the exact benefit of these strategies depends on the feature spaces of source and target domains, as discussed under RQ2.

\textit{RQ2: What specific characteristics of building energy datasets influence the effectiveness of different data-centric TL strategies when implemented on the vanilla Transformer architecture?}
\\
Our results indicate that the nature of the ambient weather features has the greatest impact on TL. In other words, using source dataset(s) that have very distinct ambient weather features from the target does not help any of the TL strategies, except strategy 8 when all the datasets are used. The other factor is the number of building records in a dataset. For datasets with a large number of building records, TL using much smaller datasets is not productive. However, it is always beneficial to TL to use very large datasets to boost performance for smaller datasets. The climate zone does show some impact, however if weather features and the number of buildings are similar, the impact of climate zone is negligible. The difference in temporal change of source and target datasets also shows an adverse impact, which can be mitigated to a certain extent by zero-padding data augmentation. It is always possible to experiment with more advanced data interpolation techniques here.  

\textit{RQ3: How does the performance of various data-centric TL strategies differ when applied to advanced Transformer architectures specifically designed for time series forecasting, compared to the vanilla Transformer?}
\\
Our experiments with vanilla Transformer, PatchTST, and FEDformer indicate that TL strategies behave similarly despite the selected Transformer architecture. Out of the three architectures, PatchTST reported the best result.

\subsection{Recommendation}
To gain the maximum benefit of TL, we recommend researchers follow a step-wise approach. For a given target domain, the most suitable subset of source datasets can be selected by inspecting the weather features, number of building data, etc., as mentioned above. To further validate the efficacy of the selected datasets, Strategy 1 (if no target data is available for model training) or Strategy 5 (if there is target data available for model training) can be used on individual source datasets. This way, the best set of source datasets can be selected for model training. If there is no target data for model training, a model trained with the selected source datasets can be used in zero-shot TL. If there is target data for model training, this data can be combined with the source data sets to train a model, which can be further fine-tuned with target data. However, we recommend testing the combined model both before and after fine-tuning, because we observed a results degradation after fine-tuning for a small number of cases. Finally, we can recommend the use of PatchTST over FEDformer and vanilla Transformer.

\subsection{Limitations and Future Work}
Although we experimented with 16 datasets, all these datasets belong to North American and European countries. Our observations might not hold for datasets obtained from other parts of the world. In the future, we expect to find data from these under-represented regions and re-validate our observations.

To keep our experiments manageable, we tried only three combinations of datasets: 2, 3, and 16. However, for a given target dataset, the optimal dataset combination may involve a different number of datasets. As discussed in the recommendation, we invite researchers to experiment with these different combinations.

To keep the experiment space manageable, we only considered 24-hour and 96-hour forecast horizons. We plan to experiment with different forecast horizons in the future.

We followed a pragmatic approach to identify the impact of datasets - we prepared datasets with different feature combinations and compared them based on their performance. Moreover, our analysis was done at a higher granularity. In other words, we considered individual datasets as a single source domain, although a dataset may include data from different building types. It is always possible to do more fine-grained source selection, by considering building type. Building energy consumption profiles may vary even across buildings of the same type. There also can be other nuances such as the stability of energy consumption~\cite{li2024comprehensive}. Therefore, in the future, we plan to incorporate similarity measurement indexes discussed in Section~\ref{sec2} to select the most suitable source datasets. 

We considered only three Transformer architectures. However, there are other Transformer architectures designed for time series prediction. With the advancements of Foundation models such as TimeGPT~\cite{garza2023timegpt} and Lag-llama~\cite{rasul2023lag}, it is now possible to try TL on these already pre-trained models. However, fine-tuning these models requires significant computational resources, which we currently do not have. Thus, these experiments are kept for the future.

\section{Conclusion}
While TL has been explored for the task of building energy forecasting, a comparative study on different data-centric TL was missing. Many past research focused on identifying novel DL architectures, and consequently, less focus was given to determining the best use of available datasets. In response to this research gap, we carried out a large-scale empirical study on the effectiveness of different TL strategies on the Transformer architectures. Our results indicate that, while TL is generally beneficial, a clear understanding of the datasets is needed to determine which exact data-centric TL strategy to use. Based on our observations on the obtained results, we made some recommendations for researchers who intend to employ TL for building energy consumption forecasting. These recommendations contribute towards laying the foundation for a better understanding of TL for building energy consumption forecasting. This empirical study can be extended further by considering different forecasting horizons, building types, newer Transformer architectures, etc. We plan to focus on these extensions in the future and invite other researchers to contribute to enhance our understanding of the exact impact of TL for building energy forecasting.

\section{Acknowledgements}\label{sec7}

This research was funded by a research grant from the College of Sciences, Massey University, New Zealand.

We also sincerely thank Matthias Hertel for providing his codebase to initiate this research, and Clayton Miller for the productive discussions at the beginning of this project.

\section{Declaration of generative AI and AI-assisted technologies in the writing process}
 During the preparation of this work the authors used Grammarly in order to identify spelling and grammar issues in the manuscript. After using this tool/service, the authors reviewed and edited the content as needed and take full responsibility for the content of the published article.
                                        % BIBLIOGRAPHY
\clearpage
\addcontentsline{toc}{section}{References} % Manually add the bibliography to the TOC
\bibliographystyle{elsarticle-num-names} 
\bibliography{references}

\appendix
\clearpage

\section{ MSE Results}
Tables~\ref{tab:24MSE} and~\ref{tab:96MSE} report the zero-shot MSE results, respectively. These correspond to Tables~\ref{zero-shot_24h} and~\ref{zero-shot_96h} in section~\ref{sec:result_zero}. Table~\ref{tab:small_Scale_summary_transformer_mse} shows MSE results for two- and three-dataset combinations and Table~\ref{tab:large_scale_summary_transformer_mse} shows MSE results for the model trained with all the datasets (for vanilla Transformer). These correspond to Table~\ref{tab:small_Scale_summary_transformer_mae} and Table~\ref{tab:large_scale_summary_transformer_mae} (respectively).
MAE and MSE results for Informer and PatchTST results are shown in Tables~\ref{tab:large_scale_summary_informer_mae},~\ref{tab:large_scale_summary_informer_mse},~\ref{tab:large_scale_summary_patchtst_mae},~\ref{tab:large_scale_summary_patchtst_mse}.

%\raggedright

\begin{table}[!h]
\centering
\caption{Average MSE for 24-hour Horizon Zero-shot Forecasting}
\label{tab:24MSE}
%\vspace{0.2cm}
\tiny
\begin{adjustbox}{max width=1\textwidth,center}
\begin{tabular}{lrrrrrrrrrrrrrrrr}
\toprule
\textbf{Model} & \textbf{Bear} & \textbf{Bobcat} & \textbf{Bull} & \textbf{Cockatoo} & \textbf{Crow} & \textbf{Eagle} & \textbf{Fox} & \textbf{Gator} & \textbf{Hog} & \textbf{Lamb} & \textbf{Moose} & \textbf{Mouse} & \textbf{Peacock} & \textbf{Rat} & \textbf{Robin} & \textbf{Wolf} \\
\midrule
Bear & \cellcolor{gray!20}0.219 & 0.341 & 0.478 & 0.454 & 0.304 & 0.310 & 0.209 & 0.460 & 0.267 & 0.178 & 0.228 & 0.302 & 0.221 & 0.238 & 0.170 & 0.425 \\
Bobcat & 0.355 & \cellcolor{gray!20}0.353 & 0.504 & 0.435 & 0.309 & 0.440 & 0.312 & 0.452 & 0.273 & 0.253 & 0.267 & 0.345 & 0.310 & 0.322 & 0.274 & 0.485 \\
Bull & 0.377 & 0.445 & \cellcolor{gray!20}0.479 & 0.492 & 0.349 & 0.417 & 0.330 & 0.540 & 0.295 & 0.236 & 0.365 & 0.393 & 0.312 & 0.321 & 0.299 & 0.525 \\
Cockatoo & 0.761 & 0.711 & 0.647 & \cellcolor{gray!20}0.426 & 0.376 & 0.890 & 0.665 & 0.496 & 0.297 & 0.369 & 0.596 & 0.397 & 0.630 & 0.544 & 0.603 & 0.715 \\
Crow & 0.434 & 0.627 & 0.744 & 0.479 & \cellcolor{gray!20}0.293 & 0.847 & 0.451 & 0.457 & 0.393 & 0.354 & 0.420 & 0.365 & 0.476 & 0.449 & 0.374 & 0.501 \\
Eagle & 0.354 & 0.426 & 0.505 & 0.418 & 0.289 & \cellcolor{gray!20}\textbf{0.234} & 0.327 & 0.420 & 0.265 & 0.215 & 0.444 & 0.299 & 0.301 & 0.316 & 0.222 & 0.496 \\
Fox & 0.221 & 0.331 & 0.452 & 0.448 & 0.301 & 0.280 & \cellcolor{gray!20}0.181 & 0.460 & 0.254 & 0.167 & 0.222 & 0.309 & 0.211 & 0.221 & 0.166 & 0.432 \\
Gator & 1.249 & 0.920 & 0.680 & 0.559 & 0.560 & 1.152 & 1.045 & \cellcolor{gray!20}\textbf{0.270} & 0.336 & 0.467 & 0.836 & 0.585 & 0.966 & 0.699 & 0.995 & 1.168 \\
Hog & 0.450 & 0.503 & 0.519 & 0.432 & 0.297 & 0.621 & 0.408 & 0.455 & \cellcolor{gray!20}0.247 & 0.234 & 0.425 & 0.319 & 0.387 & 0.373 & 0.336 & 0.546 \\
Lamb & 0.345 & 0.445 & 0.578 & 0.474 & 0.290 & 0.474 & 0.324 & 0.474 & 0.278 & \cellcolor{gray!20}0.166 & 0.252 & 0.349 & 0.324 & 0.312 & 0.244 & 0.423 \\
Moose & 0.457 & 0.557 & 0.715 & 0.481 & 0.313 & 0.898 & 0.443 & 0.494 & 0.334 & 0.269 & \cellcolor{gray!20}0.227 & 0.394 & 0.480 & 0.419 & 0.381 & 0.459 \\
Mouse & 0.397 & 0.547 & 0.563 & 0.437 & 0.300 & 0.656 & 0.398 & 0.436 & 0.280 & 0.272 & 0.453 & \cellcolor{gray!20}0.309 & 0.400 & 0.402 & 0.274 & 0.509 \\
Peacock & 0.248 & 0.418 & 0.594 & 0.508 & 0.297 & 0.356 & 0.240 & 0.478 & 0.303 & 0.201 & 0.232 & 0.326 & \cellcolor{gray!20}\textbf{0.198} & 0.270 & 0.178 & 0.423 \\
Rat & 0.243 & 0.347 & 0.501 & 0.443 & 0.287 & 0.330 & 0.206 & 0.439 & 0.260 & 0.175 & 0.217 & 0.311 & 0.230 & \cellcolor{gray!20}\textbf{0.207} & 0.203 & 0.423 \\
Robin & 0.265 & 0.407 & 0.522 & 0.423 & 0.275 & 0.348 & 0.267 & 0.434 & 0.258 & 0.188 & 0.221 & 0.295 & 0.248 & 0.281 & \cellcolor{gray!20}0.155 & 0.432 \\
Wolf & 0.399 & 0.500 & 0.681 & 0.535 & 0.336 & 0.521 & 0.362 & 0.530 & 0.351 & 0.311 & 0.315 & 0.376 & 0.391 & 0.394 & 0.296 & \cellcolor{gray!20}0.399 \\
Hogtrunc2 & 0.510 & 0.515 & 0.510 & 0.411 & 0.309 & 0.613 & 0.448 & 0.426 & \cellcolor{gray!20}0.239 & 0.245 & 0.472 & 0.335 & 0.420 & 0.400 & 0.365 & 0.576 \\
Hogtrunc & 0.749 & 0.619 & 0.561 & 0.487 & 0.415 & 0.732 & 0.635 & 0.466 & \cellcolor{gray!20}0.309 & 0.324 & 0.584 & 0.427 & 0.565 & 0.486 & 0.590 & 0.737 \\
Moosetrunc2 & 0.696 & 0.694 & 0.664 & 0.554 & 0.427 & 0.986 & 0.666 & 0.492 & 0.340 & 0.325 & \cellcolor{gray!20}0.395 & 0.466 & 0.665 & 0.515 & 0.623 & 0.746 \\
Bear+Fox & \cellcolor{blue!10}\textbf{0.207} & 0.327 & 0.459 & 0.435 & 0.293 & 0.289 & \cellcolor{blue!10}\textbf{0.177} & 0.450 & 0.250 & 0.167 & 0.220 & 0.297 & 0.209 & 0.218 & 0.166 & 0.424 \\
Bobcat+Moose & 0.307 & \cellcolor{blue!10}0.333 & 0.525 & \textbf{0.396} & 0.264 & 0.453 & 0.282 & 0.426 & 0.250 & 0.203 & \cellcolor{blue!10}\textbf{0.197} & 0.305 & 0.285 & 0.291 & 0.222 & 0.415 \\
Bull+Cockatoo & 0.387 & 0.437 & \cellcolor{blue!10}0.468 & \cellcolor{blue!10}0.428 & 0.308 & 0.453 & 0.339 & 0.486 & 0.276 & 0.225 & 0.335 & 0.343 & 0.311 & 0.324 & 0.307 & 0.476 \\
Bull+Gator & 0.454 & 0.474 & \cellcolor{blue!10}0.480 & 0.482 & 0.335 & 0.567 & 0.401 & \cellcolor{blue!10}0.289 & 0.282 & 0.233 & 0.395 & 0.358 & 0.354 & 0.354 & 0.356 & 0.545 \\
Bull+Hog & 0.358 & 0.405 & \cellcolor{blue!10}0.453 & 0.431 & 0.297 & 0.448 & 0.312 & 0.469 & \cellcolor{blue!10}0.242 & 0.209 & 0.302 & 0.332 & 0.292 & 0.294 & 0.284 & 0.467 \\
Cockatoo+Hog & 0.428 & 0.486 & 0.537 & \cellcolor{blue!10}0.403 & 0.279 & 0.647 & 0.392 & 0.445 & \cellcolor{blue!10}0.234 & 0.236 & 0.377 & 0.317 & 0.382 & 0.359 & 0.334 & 0.515 \\
Crow+Robin & 0.265 & 0.410 & 0.569 & 0.417 & \cellcolor{blue!10}\textbf{0.257} & 0.365 & 0.275 & 0.417 & 0.255 & 0.186 & 0.218 & \textbf{0.288} & 0.248 & 0.286 & \cellcolor{blue!10}0.155 & 0.413 \\
Eagle+Robin & 0.401 & 0.452 & 0.509 & 0.422 & 0.316 & \cellcolor{blue!10}0.258 & 0.361 & 0.444 & 0.263 & 0.229 & 0.474 & 0.322 & 0.335 & 0.337 & \cellcolor{blue!10}0.264 & 0.548 \\
Hog+Moose & 0.368 & 0.436 & 0.559 & 0.426 & 0.283 & 0.643 & 0.348 & 0.421 & \cellcolor{blue!10}0.235 & 0.211 & \cellcolor{blue!10}0.215 & 0.319 & 0.357 & 0.329 & 0.274 & 0.444 \\
Lamb+Robin & 0.251 & 0.374 & 0.543 & 0.438 & 0.272 & 0.312 & 0.237 & 0.421 & 0.258 & \cellcolor{blue!10}\textbf{0.148} & 0.216 & 0.315 & 0.236 & 0.255 & \cellcolor{blue!10}\textbf{0.152} & 0.410 \\
Mouse+Rat & 0.239 & 0.347 & 0.504 & 0.432 & 0.281 & 0.319 & 0.204 & 0.432 & 0.254 & 0.175 & 0.214 & \cellcolor{blue!10}0.300 & 0.230 & \cellcolor{blue!10}0.208 & 0.192 & 0.421 \\
Peacock+Wolf & 0.275 & 0.386 & 0.566 & 0.476 & 0.292 & 0.352 & 0.252 & 0.454 & 0.287 & 0.199 & 0.236 & 0.319 & \cellcolor{blue!10}0.216 & 0.279 & 0.186 & \cellcolor{blue!10}0.392 \\
Bear+Foxtrunc & \cellcolor{blue!10}0.209 & \textbf{0.324} & 0.452 & 0.437 & 0.294 & 0.291 & \cellcolor{blue!10}0.184 & 0.455 & 0.253 & 0.166 & 0.216 & 0.297 & 0.209 & 0.222 & 0.162 & 0.418 \\
Bobcat+Moosetrunc & 0.313 & \cellcolor{blue!10}0.335 & 0.522 & 0.397 & 0.265 & 0.456 & 0.287 & 0.422 & 0.255 & 0.218 & \cellcolor{blue!10}0.200 & 0.307 & 0.292 & 0.296 & 0.230 & 0.422 \\
Bulltrunc+Gator & 0.460 & 0.484 & \cellcolor{blue!10}0.484 & 0.490 & 0.340 & 0.593 & 0.409 & \cellcolor{blue!10}0.290 & 0.285 & 0.230 & 0.415 & 0.357 & 0.359 & 0.362 & 0.361 & 0.538 \\
Eagletrunc+Robin & 0.401 & 0.452 & 0.509 & 0.422 & 0.316 & \cellcolor{blue!10}0.258 & 0.361 & 0.444 & 0.263 & 0.229 & 0.474 & 0.322 & 0.335 & 0.337 & \cellcolor{blue!10}0.264 & 0.548 \\
Hogtrunc2+Moosetrunc2 & 0.504 & 0.593 & 0.588 & 0.513 & 0.369 & 0.809 & 0.485 & 0.451 & \cellcolor{blue!10}0.297 & 0.261 & \cellcolor{blue!10}0.337 & 0.372 & 0.538 & 0.422 & 0.419 & 0.627 \\
Hogtrunc+Moosetrunc2 & 0.504 & 0.593 & 0.588 & 0.513 & 0.369 & 0.809 & 0.485 & 0.451 & \cellcolor{blue!10}0.297 & 0.261 & \cellcolor{blue!10}0.337 & 0.372 & 0.538 & 0.422 & 0.419 & 0.627 \\
Lamb+Robintrunc & 0.266 & 0.372 & 0.499 & 0.438 & 0.286 & 0.321 & 0.247 & 0.416 & 0.247 & \cellcolor{blue!10}0.163 & 0.231 & 0.312 & 0.249 & 0.261 & \cellcolor{blue!10}0.170 & 0.423 \\
Peacock+Wolftrunc & 0.272 & 0.395 & 0.543 & 0.483 & 0.308 & 0.346 & 0.249 & 0.455 & 0.291 & 0.202 & 0.251 & 0.326 & \cellcolor{blue!10}0.215 & 0.276 & 0.188 & \cellcolor{blue!10}\textbf{0.382} \\
Bull+Cockatoo+Hog & 0.357 & 0.409 & \cellcolor{blue!10}\textbf{0.450} & \cellcolor{blue!10}0.410 & 0.285 & 0.477 & 0.322 & 0.431 & \cellcolor{blue!10}\textbf{0.233} & 0.205 & 0.295 & 0.315 & 0.299 & 0.300 & 0.284 & 0.453 \\
\bottomrule
\end{tabular}
\end{adjustbox}
\end{table}

\newgeometry{bottom=2.6cm}  % Adjust the bottom margin

\begin{table}[!h]
\centering
\caption{Average MSE for 96-hour Horizon Zero-shot Forecasting}
\label{tab:96MSE}
%\vspace{0.2cm}
\tiny
\begin{adjustbox}{max width=1\textwidth,center}
\begin{tabular}{lrrrrrrrrrrrrrrrr}
\toprule
\textbf{Model} & \textbf{Bear} & \textbf{Bobcat} & \textbf{Bull} & \textbf{Cockatoo} & \textbf{Crow} & \textbf{Eagle} & \textbf{Fox} & \textbf{Gator} & \textbf{Hog} & \textbf{Lamb} & \textbf{Moose} & \textbf{Mouse} & \textbf{Peacock} & \textbf{Rat} & \textbf{Robin} & \textbf{Wolf} \\
\midrule
Bear & \cellcolor{gray!20}0.323 & 0.489 & 0.683 & 0.653 & 0.466 & 0.443 & 0.299 & 0.733 & 0.442 & 0.352 & 0.351 & 0.514 & 0.338 & 0.412 & 0.290 & 0.555 \\
Bobcat & 0.578 & \cellcolor{gray!20}0.541 & 0.697 & 0.649 & 0.473 & 0.643 & 0.480 & 0.751 & 0.431 & 0.477 & 0.414 & 0.553 & 0.535 & 0.547 & 0.457 & 0.718 \\
Bull & 0.634 & 0.643 & \cellcolor{gray!20}0.717 & 0.730 & 0.498 & 0.641 & 0.523 & 0.824 & 0.451 & 0.379 & 0.572 & 0.611 & 0.538 & 0.530 & 0.537 & 0.735 \\
Cockatoo & 0.976 & 0.849 & 0.796 & \cellcolor{gray!20}0.611 & 0.541 & 1.028 & 0.823 & 0.801 & 0.452 & 0.544 & 0.696 & 0.589 & 0.829 & 0.717 & 0.793 & 0.821 \\
Crow & 0.653 & 0.921 & 1.088 & 0.725 & \cellcolor{gray!20}0.446 & 1.128 & 0.624 & 0.864 & 0.764 & 0.758 & 0.664 & 0.559 & 0.799 & 0.745 & 0.608 & 0.632 \\
Eagle & 0.572 & 0.587 & 0.775 & 0.617 & 0.450 & \cellcolor{gray!20}\textbf{0.315} & 0.490 & 0.714 & 0.475 & 0.364 & 0.654 & 0.491 & 0.491 & 0.512 & 0.377 & 0.683 \\
Fox & 0.336 & 0.472 & 0.662 & 0.616 & 0.427 & 0.396 & \cellcolor{gray!20}\textbf{0.252} & 0.707 & 0.411 & 0.304 & 0.312 & 0.488 & 0.330 & 0.375 & 0.285 & 0.558 \\
Gator & 1.141 & 0.884 & 0.828 & 0.704 & 0.630 & 1.043 & 0.943 & \cellcolor{gray!20}\textbf{0.544} & 0.486 & 0.518 & 0.819 & 0.672 & 0.884 & 0.732 & 0.942 & 1.051 \\
Hog & 0.669 & 0.686 & 0.710 & 0.643 & 0.451 & 0.795 & 0.568 & 0.762 & \cellcolor{gray!20}0.426 & 0.367 & 0.605 & 0.518 & 0.557 & 0.556 & 0.526 & 0.719 \\
Lamb & 0.479 & 0.608 & 0.876 & 0.707 & 0.454 & 0.664 & 0.445 & 0.799 & 0.516 & \cellcolor{gray!20}0.249 & 0.375 & 0.538 & 0.525 & 0.493 & 0.363 & 0.564 \\
Moose & 0.683 & 0.808 & 1.025 & 0.708 & 0.490 & 1.230 & 0.613 & 0.837 & 0.556 & 0.467 & \cellcolor{gray!20}0.334 & 0.592 & 0.806 & 0.643 & 0.584 & 0.601 \\
Mouse & 0.575 & 0.903 & 0.938 & 0.738 & 0.485 & 0.887 & 0.557 & 0.907 & 0.549 & 0.860 & 0.595 & \cellcolor{gray!20}0.575 & 0.774 & 0.731 & 0.482 & 0.696 \\
Peacock & 0.387 & 0.534 & 0.831 & 0.714 & 0.467 & 0.489 & 0.342 & 0.806 & 0.491 & 0.328 & 0.358 & 0.499 & \cellcolor{gray!20}0.327 & 0.448 & 0.292 & 0.560 \\
Rat & 0.384 & 0.475 & 0.731 & 0.660 & 0.450 & 0.481 & 0.312 & 0.773 & 0.420 & 0.268 & 0.327 & 0.505 & 0.364 & \cellcolor{gray!20}0.371 & 0.335 & 0.573 \\
Robin & 0.400 & 0.592 & 0.766 & 0.646 & 0.428 & 0.493 & 0.378 & 0.743 & 0.453 & 0.299 & 0.347 & 0.515 & 0.405 & 0.450 & \cellcolor{gray!20}0.268 & 0.548 \\
Wolf & 0.553 & 0.735 & 0.927 & 0.782 & 0.512 & 0.791 & 0.528 & 0.883 & 0.640 & 0.837 & 0.454 & 0.676 & 0.630 & 0.701 & 0.489 & \cellcolor{gray!20}\textbf{0.477} \\
Hogtrunc2 & 0.919 & 0.753 & 0.739 & 0.630 & 0.521 & 0.891 & 0.746 & 0.759 & \cellcolor{gray!20}0.442 & 0.459 & 0.724 & 0.590 & 0.678 & 0.653 & 0.725 & 0.866 \\
Hogtrunc & 1.186 & 0.854 & 0.775 & 0.685 & 0.690 & 1.061 & 0.989 & 0.907 & \cellcolor{gray!20}0.483 & 0.521 & 0.825 & 0.710 & 0.802 & 0.746 & 0.967 & 0.998 \\
Moosetrunc2 & 0.786 & 0.901 & 1.004 & 0.919 & 0.662 & 1.184 & 0.763 & 0.923 & 0.548 & 0.540 & \cellcolor{gray!20}0.525 & 0.706 & 0.778 & 0.735 & 0.707 & 0.860 \\
Bear+Fox & \cellcolor{blue!10}\textbf{0.313} & 0.449 & \textbf{0.648} & 0.609 & 0.423 & 0.384 & \cellcolor{blue!10}0.255 & 0.687 & 0.406 & 0.277 & 0.317 & 0.479 & 0.311 & 0.372 & 0.271 & 0.562 \\
Bobcat+Moose & 0.488 & \cellcolor{blue!10}0.508 & 0.741 & 0.608 & 0.425 & 0.623 & 0.419 & 0.723 & 0.437 & 0.387 & \cellcolor{blue!10}\textbf{0.302} & 0.500 & 0.473 & 0.500 & 0.376 & 0.588 \\
Bull+Cockatoo & 0.610 & 0.624 & \cellcolor{blue!10}0.714 & \cellcolor{blue!10}0.649 & 0.485 & 0.691 & 0.522 & 0.837 & 0.436 & 0.386 & 0.531 & 0.545 & 0.507 & 0.536 & 0.537 & 0.651 \\
Bull+Gator & 0.650 & 0.654 & \cellcolor{blue!10}0.686 & 0.676 & 0.496 & 0.737 & 0.559 & \cellcolor{blue!10}0.580 & 0.431 & 0.374 & 0.616 & 0.549 & 0.540 & 0.553 & 0.530 & 0.706 \\
Bull+Hog & 0.593 & 0.602 & \cellcolor{blue!10}0.701 & 0.683 & 0.473 & 0.693 & 0.490 & 0.807 & \cellcolor{blue!10}0.424 & 0.354 & 0.490 & 0.543 & 0.484 & 0.506 & 0.483 & 0.698 \\
Cockatoo+Hog & 0.793 & 0.741 & 0.777 & \cellcolor{blue!10}0.623 & 0.455 & 0.882 & 0.629 & 0.806 & \cellcolor{blue!10}0.437 & 0.415 & 0.664 & 0.539 & 0.641 & 0.589 & 0.598 & 0.791 \\
Crow+Robin & 0.393 & 0.542 & 0.776 & 0.603 & \cellcolor{blue!10}\textbf{0.390} & 0.488 & 0.366 & 0.700 & 0.427 & 0.297 & 0.328 & \textbf{0.469} & 0.399 & 0.440 & \cellcolor{blue!10}\textbf{0.258} & 0.521 \\
Eagle+Robin & 0.657 & 0.632 & 0.740 & 0.608 & 0.483 & \cellcolor{blue!10}0.390 & 0.546 & 0.762 & 0.443 & 0.391 & 0.687 & 0.510 & 0.539 & 0.540 & \cellcolor{blue!10}0.453 & 0.729 \\
Hog+Moose & 0.540 & 0.590 & 0.744 & 0.637 & 0.437 & 0.841 & 0.480 & 0.768 & \cellcolor{blue!10}0.425 & 0.340 & \cellcolor{blue!10}0.332 & 0.502 & 0.492 & 0.495 & 0.445 & 0.599 \\
Lamb+Robin & 0.376 & 0.501 & 0.732 & 0.625 & 0.409 & 0.444 & 0.333 & 0.703 & 0.417 & \cellcolor{blue!10}\textbf{0.235} & 0.320 & 0.488 & 0.377 & 0.401 & \cellcolor{blue!10}0.259 & 0.533 \\
Mouse+Rat & 0.372 & 0.472 & 0.730 & 0.648 & 0.428 & 0.458 & 0.307 & 0.747 & 0.434 & 0.271 & 0.319 & \cellcolor{blue!10}0.498 & 0.356 & \cellcolor{blue!10}\textbf{0.369} & 0.325 & 0.564 \\
Peacock+Wolf & 0.418 & 0.576 & 0.805 & 0.655 & 0.431 & 0.487 & 0.373 & 0.768 & 0.486 & 0.395 & 0.360 & 0.495 & \cellcolor{blue!10}0.356 & 0.489 & 0.301 & \cellcolor{blue!10}0.486 \\
Bear+Foxtrunc & \cellcolor{blue!10}0.314 & \textbf{0.439} & 0.653 & 0.624 & 0.426 & 0.406 & \cellcolor{blue!10}0.261 & 0.695 & 0.420 & 0.301 & 0.331 & 0.481 & \textbf{0.309} & 0.377 & 0.284 & 0.568 \\
Bobcat+Moosetrunc & 0.494 & \cellcolor{blue!10}0.515 & 0.732 & \textbf{0.600} & 0.426 & 0.641 & 0.424 & 0.718 & 0.427 & 0.386 & \cellcolor{blue!10}0.304 & 0.501 & 0.494 & 0.502 & 0.385 & 0.597 \\
Bulltrunc+Gator & 0.637 & 0.642 & \cellcolor{blue!10}0.708 & 0.698 & 0.508 & 0.778 & 0.562 & \cellcolor{blue!10}0.574 & 0.456 & 0.359 & 0.620 & 0.535 & 0.523 & 0.545 & 0.526 & 0.672 \\
Eagletrunc+Robin & 0.657 & 0.632 & 0.740 & 0.608 & 0.483 & \cellcolor{blue!10}0.390 & 0.546 & 0.762 & 0.443 & 0.391 & 0.687 & 0.510 & 0.539 & 0.540 & \cellcolor{blue!10}0.453 & 0.729 \\
Hogtrunc2+Moosetrunc2 & 0.722 & 0.729 & 0.776 & 0.727 & 0.565 & 0.921 & 0.623 & 0.911 & \cellcolor{blue!10}0.461 & 0.388 & \cellcolor{blue!10}0.510 & 0.608 & 0.649 & 0.579 & 0.606 & 0.795 \\
Hogtrunc+Moosetrunc2 & 0.722 & 0.729 & 0.776 & 0.727 & 0.565 & 0.921 & 0.623 & 0.911 & \cellcolor{blue!10}0.461 & 0.388 & \cellcolor{blue!10}0.510 & 0.608 & 0.649 & 0.579 & 0.606 & 0.795 \\
Lamb+Robintrunc & 0.411 & 0.570 & 0.737 & 0.629 & 0.420 & 0.465 & 0.359 & 0.696 & \textbf{0.400} & \cellcolor{blue!10}0.294 & 0.334 & 0.501 & 0.446 & 0.435 & \cellcolor{blue!10}0.286 & 0.548 \\
Peacock+Wolftrunc & 0.428 & 0.625 & 0.856 & 0.711 & 0.465 & 0.501 & 0.380 & 0.807 & 0.545 & 0.434 & 0.397 & 0.527 & \cellcolor{blue!10}0.365 & 0.518 & 0.318 & \cellcolor{blue!10}0.502 \\
Bull+Cockatoo+Hog & 0.577 & 0.580 & \cellcolor{blue!10}0.706 & \cellcolor{blue!10}0.632 & 0.430 & 0.689 & 0.475 & 0.757 & \cellcolor{blue!10}0.405 & 0.345 & 0.474 & 0.496 & 0.461 & 0.484 & 0.471 & 0.646 \\
\bottomrule
\end{tabular}
\end{adjustbox}
\end{table}

\begin{table}[H]
\caption{MSE results for two- and three-dataset combinations \textit{(Vanilla Transformer)} }

\centering
\begin{adjustbox}{width=\textwidth,center}
\tiny
\label{tab:small_Scale_summary_transformer_mse}
\begin{tabular}{llllrrrrr}
\toprule
Combo & Test & Hor & Baseline & Strategy 5/6 & Imp & Strategy 7/8 & Imp \\

\midrule
Bull+Cockatoo+Hog & Bull & 24hr & 0.479 & 0.450 & \cellcolorgrad{6}\textcolor{black}{6.1\%} & 0.455 & \cellcolorgrad{5}\textcolor{black}{5.0\%} \\
Bull+Cockatoo+Hog & Bull & 96hr & 0.717 & 0.706 & \cellcolorgrad{2}\textcolor{black}{1.5\%} & 0.716 & \cellcolorgrad{0}\textcolor{black}{0.1\%} \\
Bull+Cockatoo+Hog & Cockatoo & 24hr & 0.426 & 0.403 & \cellcolorgrad{5}\textcolor{black}{5.4\%} & 0.410 & \cellcolorgrad{4}\textcolor{black}{3.8\%} \\
Bull+Cockatoo+Hog & Cockatoo & 96hr & 0.611 & 0.623 & \cellcolorgrad{-2}\textcolor{black}{-2.0\%} & 0.602 & \cellcolorgrad{1}\textcolor{black}{1.5\%} \\
Bull+Cockatoo+Hog & Hog & 24hr & 0.247 & 0.233 & \cellcolorgrad{6}\textcolor{black}{5.7\%} & 0.229 & \cellcolorgrad{7}\textcolor{black}{7.3\%} \\
Bull+Cockatoo+Hog & Hog & 96hr & 0.426 & 0.405 & \cellcolorgrad{5}\textcolor{black}{4.9\%} & 0.412 & \cellcolorgrad{3}\textcolor{black}{3.3\%} \\
Bear+Fox & Bear & 24hr & 0.219 & 0.207 & \cellcolorgrad{5}\textcolor{black}{5.5\%} & 0.200 & \cellcolorgrad{9}\textcolor{black}{8.7\%} \\
Bear+Fox & Bear & 96hr & 0.323 & 0.313 & \cellcolorgrad{3}\textcolor{black}{3.1\%} & 0.307 & \cellcolorgrad{5}\textcolor{black}{5.0\%} \\
Bear+Fox & Fox & 24hr & 0.181 & 0.177 & \cellcolorgrad{2}\textcolor{black}{2.2\%} & 0.170 & \cellcolorgrad{6}\textcolor{black}{6.1\%} \\
Bear+Fox & Fox & 96hr & 0.252 & 0.255 & \cellcolorgrad{-1}\textcolor{black}{-1.2\%} & 0.246 & \cellcolorgrad{2}\textcolor{black}{2.4\%} \\
Bobcat+Moose & Bobcat & 24hr & 0.353 & 0.333 & \cellcolorgrad{6}\textcolor{black}{5.7\%} & 0.328 & \cellcolorgrad{7}\textcolor{black}{7.1\%} \\
Bobcat+Moose & Bobcat & 96hr & 0.541 & 0.508 & \cellcolorgrad{6}\textcolor{black}{6.1\%} & 0.491 & \cellcolorgrad{9}\textcolor{black}{9.2\%} \\
Bobcat+Moose & Moose & 24hr & 0.227 & 0.197 & \cellcolorgrad{13}\textcolor{black}{13.2\%} & 0.192 & \cellcolorgrad{15}\textcolor{black}{15.4\%} \\
Bobcat+Moose & Moose & 96hr & 0.334 & 0.302 & \cellcolorgrad{10}\textcolor{black}{9.6\%} & 0.298 & \cellcolorgrad{11}\textcolor{black}{10.8\%} \\
Bull+Gator & Bull & 24hr & 0.479 & 0.480 & \cellcolorgrad{0}\textcolor{black}{-0.2\%} & 0.459 & \cellcolorgrad{4}\textcolor{black}{4.2\%} \\
Bull+Gator & Bull & 96hr & 0.717 & 0.686 & \cellcolorgrad{4}\textcolor{black}{4.3\%} & 0.691 & \cellcolorgrad{4}\textcolor{black}{3.6\%} \\
Bull+Gator & Gator & 24hr & 0.270 & 0.289 & \cellcolorgrad{-7}\textcolor{black}{-7.0\%} & 0.275 & \cellcolorgrad{-2}\textcolor{black}{-1.9\%} \\
Bull+Gator & Gator & 96hr & 0.544 & 0.580 & \cellcolorgrad{-7}\textcolor{black}{-6.6\%} & 0.563 & \cellcolorgrad{-4}\textcolor{black}{-3.5\%} \\
Crow+Robin & Crow & 24hr & 0.293 & 0.257 & \cellcolorgrad{12}\textcolor{black}{12.3\%} & 0.250 & \cellcolorgrad{15}\textcolor{black}{14.7\%} \\
Crow+Robin & Crow & 96hr & 0.446 & 0.390 & \cellcolorgrad{13}\textcolor{black}{12.6\%} & 0.387 & \cellcolorgrad{13}\textcolor{black}{13.2\%} \\
Crow+Robin & Robin & 24hr & 0.155 & 0.155 & \cellcolorgrad{0}\textcolor{black}{0.0\%} & 0.154 & \cellcolorgrad{1}\textcolor{black}{0.6\%} \\
Crow+Robin & Robin & 96hr & 0.268 & 0.258 & \cellcolorgrad{4}\textcolor{black}{3.7\%} & 0.256 & \cellcolorgrad{4}\textcolor{black}{4.5\%} \\
Hog+Moose & Hog & 24hr & 0.247 & 0.235 & \cellcolorgrad{5}\textcolor{black}{4.9\%} & 0.237 & \cellcolorgrad{4}\textcolor{black}{4.0\%} \\
Hog+Moose & Hog & 96hr & 0.426 & 0.425 & \cellcolorgrad{0}\textcolor{black}{0.2\%} & 0.428 & \cellcolorgrad{0}\textcolor{black}{-0.5\%} \\
Hog+Moose & Moose & 24hr & 0.227 & 0.215 & \cellcolorgrad{5}\textcolor{black}{5.3\%} & 0.204 & \cellcolorgrad{10}\textcolor{black}{10.1\%} \\
Hog+Moose & Moose & 96hr & 0.334 & 0.332 & \cellcolorgrad{1}\textcolor{black}{0.6\%} & 0.313 & \cellcolorgrad{6}\textcolor{black}{6.3\%} \\
Lamb+Robin & Lamb & 24hr & 0.166 & 0.148 & \cellcolorgrad{11}\textcolor{black}{10.8\%} & 0.150 & \cellcolorgrad{10}\textcolor{black}{9.6\%} \\
Lamb+Robin & Lamb & 96hr & 0.249 & 0.235 & \cellcolorgrad{6}\textcolor{black}{5.6\%} & 0.232 & \cellcolorgrad{7}\textcolor{black}{6.8\%} \\
Lamb+Robin & Robin & 24hr & 0.155 & 0.152 & \cellcolorgrad{2}\textcolor{black}{1.9\%} & 0.154 & \cellcolorgrad{1}\textcolor{black}{0.6\%} \\
Lamb+Robin & Robin & 96hr & 0.268 & 0.259 & \cellcolorgrad{3}\textcolor{black}{3.4\%} & 0.261 & \cellcolorgrad{3}\textcolor{black}{2.6\%} \\
Mouse+Rat & Mouse & 24hr & 0.309 & 0.300 & \cellcolorgrad{3}\textcolor{black}{2.9\%} & 0.284 & \cellcolorgrad{8}\textcolor{black}{8.1\%} \\
Mouse+Rat & Mouse & 96hr & 0.575 & 0.498 & \cellcolorgrad{13}\textcolor{black}{13.4\%} & 0.476 & \cellcolorgrad{17}\textcolor{black}{17.2\%} \\
Mouse+Rat & Rat & 24hr & 0.207 & 0.208 & \cellcolorgrad{0}\textcolor{black}{-0.5\%} & 0.210 & \cellcolorgrad{-1}\textcolor{black}{-1.4\%} \\
Mouse+Rat & Rat & 96hr & 0.371 & 0.369 & \cellcolorgrad{1}\textcolor{black}{0.5\%} & 0.373 & \cellcolorgrad{0}\textcolor{black}{-0.5\%} \\
Peacock+Wolftruncated1 & Peacock & 24hr & 0.198 & 0.215 & \cellcolorgrad{-9}\textcolor{black}{-8.6\%} & 0.204 & \cellcolorgrad{-3}\textcolor{black}{-3.0\%} \\
Peacock+Wolftruncated1 & Peacock & 96hr & 0.327 & 0.365 & \cellcolorgrad{-12}\textcolor{black}{-11.6\%} & 0.336 & \cellcolorgrad{-3}\textcolor{black}{-2.8\%} \\
Peacock+Wolftruncated1 & Wolf & 24hr & 0.399 & 0.382 & \cellcolorgrad{4}\textcolor{black}{4.3\%} & 0.383 & \cellcolorgrad{4}\textcolor{black}{4.0\%} \\
Peacock+Wolftruncated1 & Wolf & 96hr & 0.477 & 0.502 & \cellcolorgrad{-5}\textcolor{black}{-5.2\%} & 0.492 & \cellcolorgrad{-3}\textcolor{black}{-3.1\%} \\
Eagletruncated1+Robin & Eagle & 24hr & 0.311 & 0.276 & \cellcolorgrad{11}\textcolor{black}{11.3\%} & 0.271 & \cellcolorgrad{13}\textcolor{black}{12.9\%} \\
Eagletruncated1+Robin & Eagle & 96hr & 0.461 & 0.357 & \cellcolorgrad{23}\textcolor{black}{22.6\%} & 0.357 & \cellcolorgrad{23}\textcolor{black}{22.6\%} \\
Eagletruncated1+Robin & Robin & 24hr & 0.155 & 0.264 & \cellcolorgrad{-70}\textcolor{black}{-70.3\%} & 0.154 & \cellcolorgrad{1}\textcolor{black}{0.6\%} \\
Eagletruncated1+Robin & Robin & 96hr & 0.268 & 0.453 & \cellcolorgrad{-69}\textcolor{black}{-69.0\%} & 0.252 & \cellcolorgrad{6}\textcolor{black}{6.0\%} \\
Bear+Foxtruncated1 & Bear & 24hr & 0.219 & 0.209 & \cellcolorgrad{5}\textcolor{black}{4.6\%} & 0.203 & \cellcolorgrad{7}\textcolor{black}{7.3\%} \\
Bear+Foxtruncated1 & Bear & 96hr & 0.323 & 0.314 & \cellcolorgrad{3}\textcolor{black}{2.8\%} & 0.306 & \cellcolorgrad{5}\textcolor{black}{5.3\%} \\
Bear+Foxtruncated1 & Foxtruncated1 & 24hr & 0.181 & 0.184 & \cellcolorgrad{-2}\textcolor{black}{-1.7\%} & 0.172 & \cellcolorgrad{5}\textcolor{black}{5.0\%} \\
Bear+Foxtruncated1 & Foxtruncated1 & 96hr & 0.252 & 0.261 & \cellcolorgrad{-4}\textcolor{black}{-3.6\%} & 0.244 & \cellcolorgrad{3}\textcolor{black}{3.2\%} \\
Bobcat+Moosetruncated1 & Bobcat & 24hr & 0.353 & 0.335 & \cellcolorgrad{5}\textcolor{black}{5.1\%} & 0.333 & \cellcolorgrad{6}\textcolor{black}{5.7\%} \\
Bobcat+Moosetruncated1 & Bobcat & 96hr & 0.541 & 0.515 & \cellcolorgrad{5}\textcolor{black}{4.8\%} & 0.498 & \cellcolorgrad{8}\textcolor{black}{7.9\%} \\
Bobcat+Moosetruncated1 & Moose & 24hr & 0.227 & 0.200 & \cellcolorgrad{12}\textcolor{black}{11.9\%} & 0.196 & \cellcolorgrad{14}\textcolor{black}{13.7\%} \\
Bobcat+Moosetruncated1 & Moose & 96hr & 0.334 & 0.304 & \cellcolorgrad{9}\textcolor{black}{9.0\%} & 0.299 & \cellcolorgrad{10}\textcolor{black}{10.5\%} \\
Eagle+Robin & Eagle & 24hr & 0.311 & 0.276 & \cellcolorgrad{11}\textcolor{black}{11.3\%} & 0.271 & \cellcolorgrad{13}\textcolor{black}{12.9\%} \\
Eagle+Robin & Eagle & 96hr & 0.461 & 0.357 & \cellcolorgrad{23}\textcolor{black}{22.6\%} & 0.357 & \cellcolorgrad{23}\textcolor{black}{22.6\%} \\
Eagle+Robin & Robin & 24hr & 0.155 & 0.264 & \cellcolorgrad{-70}\textcolor{black}{-70.3\%} & 0.154 & \cellcolorgrad{1}\textcolor{black}{0.6\%} \\
Eagle+Robin & Robin & 96hr & 0.268 & 0.453 & \cellcolorgrad{-69}\textcolor{black}{-69.0\%} & 0.252 & \cellcolorgrad{6}\textcolor{black}{6.0\%} \\
Bulltruncated1+Gator & Bull & 24hr & 0.479 & 0.484 & \cellcolorgrad{-1}\textcolor{black}{-1.0\%} & 0.457 & \cellcolorgrad{5}\textcolor{black}{4.6\%} \\
Bulltruncated1+Gator & Bull & 96hr & 0.717 & 0.708 & \cellcolorgrad{1}\textcolor{black}{1.3\%} & 0.681 & \cellcolorgrad{5}\textcolor{black}{5.0\%} \\
Bulltruncated1+Gator & Gator & 24hr & 0.270 & 0.290 & \cellcolorgrad{-7}\textcolor{black}{-7.4\%} & 0.272 & \cellcolorgrad{-1}\textcolor{black}{-0.7\%} \\
Bulltruncated1+Gator & Gator & 96hr & 0.544 & 0.574 & \cellcolorgrad{-6}\textcolor{black}{-5.5\%} & 0.564 & \cellcolorgrad{-4}\textcolor{black}{-3.7\%} \\
Peacock+Wolf & Peacock & 24hr & 0.198 & 0.216 & \cellcolorgrad{-9}\textcolor{black}{-9.1\%} & 0.204 & \cellcolorgrad{-3}\textcolor{black}{-3.0\%} \\
Peacock+Wolf & Peacock & 96hr & 0.327 & 0.356 & \cellcolorgrad{-9}\textcolor{black}{-8.9\%} & 0.333 & \cellcolorgrad{-2}\textcolor{black}{-1.8\%} \\
Peacock+Wolf & Wolf & 24hr & 0.399 & 0.392 & \cellcolorgrad{2}\textcolor{black}{1.8\%} & 0.398 & \cellcolorgrad{0}\textcolor{black}{0.3\%} \\
Peacock+Wolf & Wolf & 96hr & 0.477 & 0.486 & \cellcolorgrad{-2}\textcolor{black}{-1.9\%} & 0.493 & \cellcolorgrad{-3}\textcolor{black}{-3.4\%} \\
Lamb+Robintruncated1 & Lamb & 24hr & 0.166 & 0.163 & \cellcolorgrad{2}\textcolor{black}{1.8\%} & 0.150 & \cellcolorgrad{10}\textcolor{black}{9.6\%} \\
Lamb+Robintruncated1 & Lamb & 96hr & 0.249 & 0.294 & \cellcolorgrad{-18}\textcolor{black}{-18.1\%} & 0.252 & \cellcolorgrad{-1}\textcolor{black}{-1.2\%} \\
Lamb+Robintruncated1 & Robintruncated1 & 24hr & 0.155 & 0.170 & \cellcolorgrad{-10}\textcolor{black}{-9.7\%} & 0.153 & \cellcolorgrad{1}\textcolor{black}{1.3\%} \\
Lamb+Robintruncated1 & Robintruncated1 & 96hr & 0.268 & 0.286 & \cellcolorgrad{-7}\textcolor{black}{-6.7\%} & 0.262 & \cellcolorgrad{2}\textcolor{black}{2.2\%} \\
Hogtruncated1+Moosetruncated2 & Hogtruncated1 & 24hr & 0.359 & 0.290 & \cellcolorgrad{19}\textcolor{black}{19.2\%} & 0.298 & \cellcolorgrad{17}\textcolor{black}{17.0\%} \\
Hogtruncated1+Moosetruncated2 & Hogtruncated1 & 96hr & 0.666 & 0.457 & \cellcolorgrad{31}\textcolor{black}{31.4\%} & 0.471 & \cellcolorgrad{29}\textcolor{black}{29.3\%} \\
Hogtruncated1+Moosetruncated2 & Moosetruncated2 & 24hr & 0.175 & 0.163 & \cellcolorgrad{7}\textcolor{black}{6.9\%} & 0.163 & \cellcolorgrad{7}\textcolor{black}{6.9\%} \\
Hogtruncated1+Moosetruncated2 & Moosetruncated2 & 96hr & 0.313 & 0.271 & \cellcolorgrad{13}\textcolor{black}{13.4\%} & 0.286 & \cellcolorgrad{9}\textcolor{black}{8.6\%} \\
Hogtruncated2+Moosetruncated2 & Hogtruncated2 & 24hr & 0.320 & 0.397 & \cellcolorgrad{-24}\textcolor{black}{-24.1\%} & 0.330 & \cellcolorgrad{-3}\textcolor{black}{-3.1\%} \\
Hogtruncated2+Moosetruncated2 & Hogtruncated2 & 96hr & 0.540 & 0.589 & \cellcolorgrad{-9}\textcolor{black}{-9.1\%} & 0.505 & \cellcolorgrad{6}\textcolor{black}{6.5\%} \\
Hogtruncated2+Moosetruncated2 & Moosetruncated2 & 24hr & 0.175 & 0.163 & \cellcolorgrad{7}\textcolor{black}{6.9\%} & 0.163 & \cellcolorgrad{7}\textcolor{black}{6.9\%} \\
Hogtruncated2+Moosetruncated2 & Moosetruncated2 & 96hr & 0.313 & 0.271 & \cellcolorgrad{13}\textcolor{black}{13.4\%} & 0.286 & \cellcolorgrad{9}\textcolor{black}{8.6\%} \\
\bottomrule
\end{tabular}
\end{adjustbox}
\end{table}
\restoregeometry

\begin{table}[H]
\caption{MSE Performance Summary for Large-Scale Modelling on Individual Datasets\\ \textit{(Vanilla Transformer)}}
\vspace{2mm}
\centering
\begin{adjustbox}{width=\textwidth,center}
\tiny
\label{tab:large_scale_summary_transformer_mse}
\begin{tabular}{llrrrrrr}
\toprule
Dataset & Horizon & Baseline & Strategy 6 & Imp & Strategy 8 & Imp \\
 & & MSE & MSE & (\%) & MSE & (\%) \\
\midrule
Bear & 24hr & 0.219 & 0.211 & \cellcolorgrad{4}\textcolor{black}{+3.7\%} & 0.201 & \cellcolorgrad{8}\textcolor{black}{+8.2\%} \\
Bear & 96hr & 0.323 & 0.347 & \cellcolorgrad{-7}\textcolor{black}{-7.4\%} & 0.314 & \cellcolorgrad{3}\textcolor{black}{+2.8\%} \\
Bobcat & 24hr & 0.353 & 0.311 & \cellcolorgrad{12}\textcolor{black}{+11.9\%} & 0.301 & \cellcolorgrad{15}\textcolor{black}{+14.7\%} \\
Bobcat & 96hr & 0.541 & 0.438 & \cellcolorgrad{19}\textcolor{black}{+19.0\%} & 0.426 & \cellcolorgrad{21}\textcolor{black}{+21.3\%} \\
Bull & 24hr & 0.479 & 0.436 & \cellcolorgrad{9}\textcolor{black}{+9.0\%} & 0.432 & \cellcolorgrad{10}\textcolor{black}{+9.8\%} \\
Bull & 96hr & 0.717 & 0.635 & \cellcolorgrad{11}\textcolor{black}{+11.4\%} & 0.655 & \cellcolorgrad{9}\textcolor{black}{+8.6\%} \\
Cockatoo & 24hr & 0.426 & 0.384 & \cellcolorgrad{10}\textcolor{black}{+9.9\%} & 0.379 & \cellcolorgrad{11}\textcolor{black}{+11.0\%} \\
Cockatoo & 96hr & 0.611 & 0.583 & \cellcolorgrad{5}\textcolor{black}{+4.6\%} & 0.566 & \cellcolorgrad{7}\textcolor{black}{+7.4\%} \\
Crow & 24hr & 0.293 & 0.254 & \cellcolorgrad{13}\textcolor{black}{+13.3\%} & 0.246 & \cellcolorgrad{16}\textcolor{black}{+16.0\%} \\
Crow & 96hr & 0.446 & 0.404 & \cellcolorgrad{9}\textcolor{black}{+9.4\%} & 0.389 & \cellcolorgrad{13}\textcolor{black}{+12.8\%} \\
Eagle & 24hr & 0.311 & 0.256 & \cellcolorgrad{18}\textcolor{black}{+17.7\%} & 0.261 & \cellcolorgrad{16}\textcolor{black}{+16.1\%} \\
Eagle & 96hr & 0.461 & 0.345 & \cellcolorgrad{25}\textcolor{black}{+25.2\%} & 0.351 & \cellcolorgrad{24}\textcolor{black}{+23.9\%} \\
Fox & 24hr & 0.181 & 0.178 & \cellcolorgrad{2}\textcolor{black}{+1.7\%} & 0.166 & \cellcolorgrad{8}\textcolor{black}{+8.3\%} \\
Fox & 96hr & 0.252 & 0.281 & \cellcolorgrad{-12}\textcolor{black}{-11.5\%} & 0.248 & \cellcolorgrad{2}\textcolor{black}{+1.6\%} \\
Gator & 24hr & 0.270 & 0.310 & \cellcolorgrad{-15}\textcolor{black}{-14.8\%} & 0.270 & \cellcolorgrad{0}\textcolor{black}{+0.0\%} \\
Gator & 96hr & 0.544 & 0.588 & \cellcolorgrad{-8}\textcolor{black}{-8.1\%} & 0.536 & \cellcolorgrad{2}\textcolor{black}{+1.5\%} \\
Hog & 24hr & 0.247 & 0.213 & \cellcolorgrad{14}\textcolor{black}{+13.8\%} & 0.209 & \cellcolorgrad{15}\textcolor{black}{+15.4\%} \\
Hog & 96hr & 0.426 & 0.356 & \cellcolorgrad{16}\textcolor{black}{+16.4\%} & 0.372 & \cellcolorgrad{13}\textcolor{black}{+12.7\%} \\
Lamb & 24hr & 0.166 & 0.150 & \cellcolorgrad{10}\textcolor{black}{+9.6\%} & 0.141 & \cellcolorgrad{15}\textcolor{black}{+15.1\%} \\
Lamb & 96hr & 0.249 & 0.257 & \cellcolorgrad{-3}\textcolor{black}{-3.2\%} & 0.223 & \cellcolorgrad{10}\textcolor{black}{+10.4\%} \\
Moose & 24hr & 0.227 & 0.198 & \cellcolorgrad{13}\textcolor{black}{+12.8\%} & 0.185 & \cellcolorgrad{19}\textcolor{black}{+18.5\%} \\
Moose & 96hr & 0.334 & 0.323 & \cellcolorgrad{3}\textcolor{black}{+3.3\%} & 0.285 & \cellcolorgrad{15}\textcolor{black}{+14.7\%} \\
Mouse & 24hr & 0.309 & 0.277 & \cellcolorgrad{10}\textcolor{black}{+10.4\%} & 0.270 & \cellcolorgrad{13}\textcolor{black}{+12.6\%} \\
Mouse & 96hr & 0.575 & 0.452 & \cellcolorgrad{21}\textcolor{black}{+21.4\%} & 0.433 & \cellcolorgrad{25}\textcolor{black}{+24.7\%} \\
Peacock & 24hr & 0.198 & 0.201 & \cellcolorgrad{-2}\textcolor{black}{-1.5\%} & 0.189 & \cellcolorgrad{5}\textcolor{black}{+4.5\%} \\
Peacock & 96hr & 0.327 & 0.326 & \cellcolorgrad{0}\textcolor{black}{+0.3\%} & 0.296 & \cellcolorgrad{10}\textcolor{black}{+9.5\%} \\
Rat & 24hr & 0.207 & 0.194 & \cellcolorgrad{6}\textcolor{black}{+6.3\%} & 0.194 & \cellcolorgrad{6}\textcolor{black}{+6.3\%} \\
Rat & 96hr & 0.371 & 0.352 & \cellcolorgrad{5}\textcolor{black}{+5.1\%} & 0.343 & \cellcolorgrad{8}\textcolor{black}{+7.5\%} \\
Robin & 24hr & 0.155 & 0.150 & \cellcolorgrad{3}\textcolor{black}{+3.2\%} & 0.142 & \cellcolorgrad{8}\textcolor{black}{+8.4\%} \\
Robin & 96hr & 0.268 & 0.280 & \cellcolorgrad{-5}\textcolor{black}{-4.5\%} & 0.240 & \cellcolorgrad{10}\textcolor{black}{+10.4\%} \\
Wolf & 24hr & 0.399 & 0.394 & \cellcolorgrad{1}\textcolor{black}{+1.3\%} & 0.362 & \cellcolorgrad{9}\textcolor{black}{+9.3\%} \\
Wolf & 96hr & 0.477 & 0.529 & \cellcolorgrad{-11}\textcolor{black}{-10.9\%} & 0.468 & \cellcolorgrad{2}\textcolor{black}{+1.9\%} \\
\bottomrule
\end{tabular}
\end{adjustbox}
\end{table}

\begin{table}[H]
\caption{MAE Performance Summary for Large-Scale Modelling on Individual Datasets\\ \textit{(Informer)}}
\vspace{2mm}
\centering
\begin{adjustbox}{width=\textwidth,center}
\tiny
\label{tab:large_scale_summary_informer_mae}
\begin{tabular}{llrrrrrr}
\toprule
Dataset & Horizon & Baseline & Strategy 6 & Imp & Strategy 8 & Imp \\
 & & MAE & MAE & (\%) & MAE & (\%) \\
\midrule
Bear & 24hr & 0.306 & 0.292 & \cellcolorgrad{5}\textcolor{black}{+4.6\%} & 0.287 & \cellcolorgrad{6}\textcolor{black}{+6.2\%} \\
Bear & 96hr & 0.379 & 0.387 & \cellcolorgrad{-2}\textcolor{black}{-2.1\%} & 0.372 & \cellcolorgrad{2}\textcolor{black}{+1.8\%} \\
Bobcat & 24hr & 0.392 & 0.360 & \cellcolorgrad{8}\textcolor{black}{+8.2\%} & 0.356 & \cellcolorgrad{9}\textcolor{black}{+9.2\%} \\
Bobcat & 96hr & 0.492 & 0.460 & \cellcolorgrad{7}\textcolor{black}{+6.5\%} & 0.458 & \cellcolorgrad{7}\textcolor{black}{+6.9\%} \\
Bull & 24hr & 0.467 & 0.446 & \cellcolorgrad{5}\textcolor{black}{+4.5\%} & 0.448 & \cellcolorgrad{4}\textcolor{black}{+4.1\%} \\
Bull & 96hr & 0.585 & 0.561 & \cellcolorgrad{4}\textcolor{black}{+4.1\%} & 0.566 & \cellcolorgrad{3}\textcolor{black}{+3.2\%} \\
Cockatoo & 24hr & 0.455 & 0.425 & \cellcolorgrad{7}\textcolor{black}{+6.6\%} & 0.430 & \cellcolorgrad{6}\textcolor{black}{+5.5\%} \\
Cockatoo & 96hr & 0.600 & 0.578 & \cellcolorgrad{4}\textcolor{black}{+3.7\%} & 0.582 & \cellcolorgrad{3}\textcolor{black}{+3.0\%} \\
Crow & 24hr & 0.345 & 0.294 & \cellcolorgrad{15}\textcolor{black}{+14.8\%} & 0.299 & \cellcolorgrad{13}\textcolor{black}{+13.3\%} \\
Crow & 96hr & 0.471 & 0.419 & \cellcolorgrad{11}\textcolor{black}{+11.0\%} & 0.418 & \cellcolorgrad{11}\textcolor{black}{+11.3\%} \\
Eagle & 24hr & 0.381 & 0.348 & \cellcolorgrad{9}\textcolor{black}{+8.7\%} & 0.349 & \cellcolorgrad{8}\textcolor{black}{+8.4\%} \\
Eagle & 96hr & 0.485 & 0.417 & \cellcolorgrad{14}\textcolor{black}{+14.0\%} & 0.425 & \cellcolorgrad{12}\textcolor{black}{+12.4\%} \\
Fox & 24hr & 0.262 & 0.266 & \cellcolorgrad{-2}\textcolor{black}{-1.5\%} & 0.257 & \cellcolorgrad{2}\textcolor{black}{+1.9\%} \\
Fox & 96hr & 0.330 & 0.347 & \cellcolorgrad{-5}\textcolor{black}{-5.2\%} & 0.324 & \cellcolorgrad{2}\textcolor{black}{+1.8\%} \\
Gator & 24hr & 0.300 & 0.291 & \cellcolorgrad{3}\textcolor{black}{+3.0\%} & 0.285 & \cellcolorgrad{5}\textcolor{black}{+5.0\%} \\
Gator & 96hr & 0.506 & 0.508 & \cellcolorgrad{0}\textcolor{black}{-0.4\%} & 0.494 & \cellcolorgrad{2}\textcolor{black}{+2.4\%} \\
Hog & 24hr & 0.320 & 0.293 & \cellcolorgrad{8}\textcolor{black}{+8.4\%} & 0.296 & \cellcolorgrad{8}\textcolor{black}{+7.5\%} \\
Hog & 96hr & 0.468 & 0.417 & \cellcolorgrad{11}\textcolor{black}{+10.9\%} & 0.424 & \cellcolorgrad{9}\textcolor{black}{+9.4\%} \\
Lamb & 24hr & 0.192 & 0.186 & \cellcolorgrad{3}\textcolor{black}{+3.1\%} & 0.179 & \cellcolorgrad{7}\textcolor{black}{+6.8\%} \\
Lamb & 96hr & 0.273 & 0.279 & \cellcolorgrad{-2}\textcolor{black}{-2.2\%} & 0.261 & \cellcolorgrad{4}\textcolor{black}{+4.4\%} \\
Moose & 24hr & 0.285 & 0.258 & \cellcolorgrad{9}\textcolor{black}{+9.5\%} & 0.255 & \cellcolorgrad{11}\textcolor{black}{+10.5\%} \\
Moose & 96hr & 0.368 & 0.369 & \cellcolorgrad{0}\textcolor{black}{-0.3\%} & 0.344 & \cellcolorgrad{7}\textcolor{black}{+6.5\%} \\
Mouse & 24hr & 0.351 & 0.320 & \cellcolorgrad{9}\textcolor{black}{+8.8\%} & 0.319 & \cellcolorgrad{9}\textcolor{black}{+9.1\%} \\
Mouse & 96hr & 0.497 & 0.444 & \cellcolorgrad{11}\textcolor{black}{+10.7\%} & 0.444 & \cellcolorgrad{11}\textcolor{black}{+10.7\%} \\
Peacock & 24hr & 0.315 & 0.310 & \cellcolorgrad{2}\textcolor{black}{+1.6\%} & 0.303 & \cellcolorgrad{4}\textcolor{black}{+3.8\%} \\
Peacock & 96hr & 0.395 & 0.403 & \cellcolorgrad{-2}\textcolor{black}{-2.0\%} & 0.392 & \cellcolorgrad{1}\textcolor{black}{+0.8\%} \\
Rat & 24hr & 0.291 & 0.282 & \cellcolorgrad{3}\textcolor{black}{+3.1\%} & 0.283 & \cellcolorgrad{3}\textcolor{black}{+2.7\%} \\
Rat & 96hr & 0.410 & 0.399 & \cellcolorgrad{3}\textcolor{black}{+2.7\%} & 0.397 & \cellcolorgrad{3}\textcolor{black}{+3.2\%} \\
Robin & 24hr & 0.259 & 0.247 & \cellcolorgrad{5}\textcolor{black}{+4.6\%} & 0.238 & \cellcolorgrad{8}\textcolor{black}{+8.1\%} \\
Robin & 96hr & 0.343 & 0.353 & \cellcolorgrad{-3}\textcolor{black}{-2.9\%} & 0.326 & \cellcolorgrad{5}\textcolor{black}{+5.0\%} \\
Wolf & 24hr & 0.379 & 0.369 & \cellcolorgrad{3}\textcolor{black}{+2.6\%} & 0.362 & \cellcolorgrad{5}\textcolor{black}{+4.5\%} \\
Wolf & 96hr & 0.451 & 0.463 & \cellcolorgrad{-3}\textcolor{black}{-2.7\%} & 0.455 & \cellcolorgrad{-1}\textcolor{black}{-0.9\%} \\
\bottomrule
\end{tabular}
\end{adjustbox}
\end{table}

\begin{table}[H]
\caption{MSE Performance Summary for Large-Scale Modelling on Individual Datasets\\ \textit{(Informer)}}
\vspace{2mm}
\centering
\begin{adjustbox}{width=\textwidth,center}
\tiny
\label{tab:large_scale_summary_informer_mse}
\begin{tabular}{llrrrrrr}
\toprule
Dataset & Horizon & Baseline & Strategy 6 & Imp & Strategy 8 & Imp \\
 & & MSE & MSE & (\%) & MSE & (\%) \\
\midrule
Bear & 24hr & 0.218 & 0.211 & \cellcolorgrad{3}\textcolor{black}{+3.2\%} & 0.202 & \cellcolorgrad{7}\textcolor{black}{+7.3\%} \\
Bear & 96hr & 0.323 & 0.340 & \cellcolorgrad{-5}\textcolor{black}{-5.3\%} & 0.320 & \cellcolorgrad{1}\textcolor{black}{+0.9\%} \\
Bobcat & 24hr & 0.326 & 0.300 & \cellcolorgrad{8}\textcolor{black}{+8.0\%} & 0.293 & \cellcolorgrad{10}\textcolor{black}{+10.1\%} \\
Bobcat & 96hr & 0.452 & 0.430 & \cellcolorgrad{5}\textcolor{black}{+4.9\%} & 0.423 & \cellcolorgrad{6}\textcolor{black}{+6.4\%} \\
Bull & 24hr & 0.461 & 0.425 & \cellcolorgrad{8}\textcolor{black}{+7.8\%} & 0.431 & \cellcolorgrad{7}\textcolor{black}{+6.5\%} \\
Bull & 96hr & 0.678 & 0.625 & \cellcolorgrad{8}\textcolor{black}{+7.8\%} & 0.635 & \cellcolorgrad{6}\textcolor{black}{+6.3\%} \\
Cockatoo & 24hr & 0.408 & 0.376 & \cellcolorgrad{8}\textcolor{black}{+7.8\%} & 0.382 & \cellcolorgrad{6}\textcolor{black}{+6.4\%} \\
Cockatoo & 96hr & 0.612 & 0.593 & \cellcolorgrad{3}\textcolor{black}{+3.1\%} & 0.597 & \cellcolorgrad{3}\textcolor{black}{+2.5\%} \\
Crow & 24hr & 0.273 & 0.242 & \cellcolorgrad{11}\textcolor{black}{+11.4\%} & 0.240 & \cellcolorgrad{12}\textcolor{black}{+12.1\%} \\
Crow & 96hr & 0.430 & 0.392 & \cellcolorgrad{9}\textcolor{black}{+8.8\%} & 0.384 & \cellcolorgrad{11}\textcolor{black}{+10.7\%} \\
Eagle & 24hr & 0.305 & 0.258 & \cellcolorgrad{15}\textcolor{black}{+15.4\%} & 0.262 & \cellcolorgrad{14}\textcolor{black}{+14.1\%} \\
Eagle & 96hr & 0.462 & 0.344 & \cellcolorgrad{26}\textcolor{black}{+25.5\%} & 0.358 & \cellcolorgrad{23}\textcolor{black}{+22.5\%} \\
Fox & 24hr & 0.173 & 0.178 & \cellcolorgrad{-3}\textcolor{black}{-2.9\%} & 0.168 & \cellcolorgrad{3}\textcolor{black}{+2.9\%} \\
Fox & 96hr & 0.255 & 0.278 & \cellcolorgrad{-9}\textcolor{black}{-9.0\%} & 0.251 & \cellcolorgrad{2}\textcolor{black}{+1.6\%} \\
Gator & 24hr & 0.273 & 0.279 & \cellcolorgrad{-2}\textcolor{black}{-2.2\%} & 0.271 & \cellcolorgrad{1}\textcolor{black}{+0.7\%} \\
Gator & 96hr & 0.541 & 0.561 & \cellcolorgrad{-4}\textcolor{black}{-3.7\%} & 0.540 & \cellcolorgrad{0}\textcolor{black}{+0.2\%} \\
Hog & 24hr & 0.227 & 0.209 & \cellcolorgrad{8}\textcolor{black}{+7.9\%} & 0.213 & \cellcolorgrad{6}\textcolor{black}{+6.2\%} \\
Hog & 96hr & 0.431 & 0.364 & \cellcolorgrad{16}\textcolor{black}{+15.5\%} & 0.376 & \cellcolorgrad{13}\textcolor{black}{+12.8\%} \\
Lamb & 24hr & 0.154 & 0.149 & \cellcolorgrad{3}\textcolor{black}{+3.2\%} & 0.140 & \cellcolorgrad{9}\textcolor{black}{+9.1\%} \\
Lamb & 96hr & 0.243 & 0.253 & \cellcolorgrad{-4}\textcolor{black}{-4.1\%} & 0.225 & \cellcolorgrad{7}\textcolor{black}{+7.4\%} \\
Moose & 24hr & 0.208 & 0.189 & \cellcolorgrad{9}\textcolor{black}{+9.1\%} & 0.186 & \cellcolorgrad{11}\textcolor{black}{+10.6\%} \\
Moose & 96hr & 0.321 & 0.310 & \cellcolorgrad{3}\textcolor{black}{+3.4\%} & 0.285 & \cellcolorgrad{11}\textcolor{black}{+11.2\%} \\
Mouse & 24hr & 0.293 & 0.267 & \cellcolorgrad{9}\textcolor{black}{+8.9\%} & 0.262 & \cellcolorgrad{11}\textcolor{black}{+10.6\%} \\
Mouse & 96hr & 0.496 & 0.452 & \cellcolorgrad{9}\textcolor{black}{+8.9\%} & 0.438 & \cellcolorgrad{12}\textcolor{black}{+11.7\%} \\
Peacock & 24hr & 0.197 & 0.197 & \cellcolorgrad{0}\textcolor{black}{+0.0\%} & 0.187 & \cellcolorgrad{5}\textcolor{black}{+5.1\%} \\
Peacock & 96hr & 0.300 & 0.311 & \cellcolorgrad{-4}\textcolor{black}{-3.7\%} & 0.298 & \cellcolorgrad{1}\textcolor{black}{+0.7\%} \\
Rat & 24hr & 0.206 & 0.198 & \cellcolorgrad{4}\textcolor{black}{+3.9\%} & 0.196 & \cellcolorgrad{5}\textcolor{black}{+4.9\%} \\
Rat & 96hr & 0.365 & 0.354 & \cellcolorgrad{3}\textcolor{black}{+3.0\%} & 0.349 & \cellcolorgrad{4}\textcolor{black}{+4.4\%} \\
Robin & 24hr & 0.159 & 0.150 & \cellcolorgrad{6}\textcolor{black}{+5.7\%} & 0.141 & \cellcolorgrad{11}\textcolor{black}{+11.3\%} \\
Robin & 96hr & 0.263 & 0.277 & \cellcolorgrad{-5}\textcolor{black}{-5.3\%} & 0.243 & \cellcolorgrad{8}\textcolor{black}{+7.6\%} \\
Wolf & 24hr & 0.362 & 0.379 & \cellcolorgrad{-5}\textcolor{black}{-4.7\%} & 0.356 & \cellcolorgrad{2}\textcolor{black}{+1.7\%} \\
Wolf & 96hr & 0.470 & 0.511 & \cellcolorgrad{-9}\textcolor{black}{-8.7\%} & 0.476 & \cellcolorgrad{-1}\textcolor{black}{-1.3\%} \\
\bottomrule
\end{tabular}
\end{adjustbox}
\end{table}

\begin{table}[H]
\caption{MAE Performance Summary for Large-Scale Modelling on Individual Datasets\\ \textit{(PatchTST)}}
\vspace{2mm}
\centering
\begin{adjustbox}{width=\textwidth,center}
\tiny
\label{tab:large_scale_summary_patchtst_mae}
\begin{tabular}{llrrrrrr}
\toprule
Dataset & Horizon & Baseline & Strategy 6 & Imp & Strategy 8 & Imp \\
 & & MAE & MAE & (\%) & MAE & (\%) \\
\midrule
Bear & 24hr & 0.290 & 0.281 & \cellcolorgrad{3}\textcolor{black}{+3.0\%} & 0.278 & \cellcolorgrad{4}\textcolor{black}{+4.0\%} \\
Bear & 96hr & 0.377 & 0.376 & \cellcolorgrad{0}\textcolor{black}{+0.3\%} & 0.362 & \cellcolorgrad{4}\textcolor{black}{+3.8\%} \\
Bobcat & 24hr & 0.373 & 0.281 & \cellcolorgrad{25}\textcolor{black}{+24.6\%} & 0.346 & \cellcolorgrad{7}\textcolor{black}{+7.1\%} \\
Bobcat & 96hr & 0.451 & 0.376 & \cellcolorgrad{17}\textcolor{black}{+16.8\%} & 0.439 & \cellcolorgrad{3}\textcolor{black}{+2.8\%} \\
Bull & 24hr & 0.435 & 0.281 & \cellcolorgrad{35}\textcolor{black}{+35.4\%} & 0.426 & \cellcolorgrad{2}\textcolor{black}{+2.1\%} \\
Bull & 96hr & 0.535 & 0.376 & \cellcolorgrad{30}\textcolor{black}{+29.8\%} & 0.537 & \cellcolorgrad{0}\textcolor{black}{-0.4\%} \\
Cockatoo & 24hr & 0.415 & 0.281 & \cellcolorgrad{32}\textcolor{black}{+32.3\%} & 0.397 & \cellcolorgrad{4}\textcolor{black}{+4.3\%} \\
Cockatoo & 96hr & 0.545 & 0.376 & \cellcolorgrad{31}\textcolor{black}{+31.1\%} & 0.543 & \cellcolorgrad{0}\textcolor{black}{+0.3\%} \\
Crow & 24hr & 0.315 & 0.281 & \cellcolorgrad{11}\textcolor{black}{+10.6\%} & 0.283 & \cellcolorgrad{10}\textcolor{black}{+9.9\%} \\
Crow & 96hr & 0.434 & 0.376 & \cellcolorgrad{13}\textcolor{black}{+13.5\%} & 0.423 & \cellcolorgrad{3}\textcolor{black}{+2.6\%} \\
Eagle & 24hr & 0.326 & 0.281 & \cellcolorgrad{14}\textcolor{black}{+13.7\%} & 0.316 & \cellcolorgrad{3}\textcolor{black}{+3.0\%} \\
Eagle & 96hr & 0.394 & 0.376 & \cellcolorgrad{5}\textcolor{black}{+4.6\%} & 0.385 & \cellcolorgrad{2}\textcolor{black}{+2.3\%} \\
Fox & 24hr & 0.259 & 0.281 & \cellcolorgrad{-9}\textcolor{black}{-8.6\%} & 0.250 & \cellcolorgrad{3}\textcolor{black}{+3.3\%} \\
Fox & 96hr & 0.324 & 0.376 & \cellcolorgrad{-16}\textcolor{black}{-15.9\%} & 0.317 & \cellcolorgrad{2}\textcolor{black}{+2.1\%} \\
Gator & 24hr & 0.286 & 0.281 & \cellcolorgrad{2}\textcolor{black}{+1.5\%} & 0.282 & \cellcolorgrad{1}\textcolor{black}{+1.3\%} \\
Gator & 96hr & 0.488 & 0.376 & \cellcolorgrad{23}\textcolor{black}{+23.0\%} & 0.486 & \cellcolorgrad{0}\textcolor{black}{+0.4\%} \\
Hog & 24hr & 0.291 & 0.281 & \cellcolorgrad{3}\textcolor{black}{+3.5\%} & 0.273 & \cellcolorgrad{6}\textcolor{black}{+6.3\%} \\
Hog & 96hr & 0.408 & 0.376 & \cellcolorgrad{8}\textcolor{black}{+7.9\%} & 0.390 & \cellcolorgrad{4}\textcolor{black}{+4.2\%} \\
Lamb & 24hr & 0.173 & 0.281 & \cellcolorgrad{-63}\textcolor{black}{-62.8\%} & 0.164 & \cellcolorgrad{5}\textcolor{black}{+5.3\%} \\
Lamb & 96hr & 0.222 & 0.376 & \cellcolorgrad{-69}\textcolor{black}{-69.4\%} & 0.216 & \cellcolorgrad{3}\textcolor{black}{+2.6\%} \\
Moose & 24hr & 0.267 & 0.281 & \cellcolorgrad{-5}\textcolor{black}{-5.4\%} & 0.234 & \cellcolorgrad{12}\textcolor{black}{+12.3\%} \\
Moose & 96hr & 0.350 & 0.376 & \cellcolorgrad{-7}\textcolor{black}{-7.4\%} & 0.332 & \cellcolorgrad{5}\textcolor{black}{+5.1\%} \\
Mouse & 24hr & 0.331 & 0.281 & \cellcolorgrad{15}\textcolor{black}{+15.0\%} & 0.312 & \cellcolorgrad{6}\textcolor{black}{+5.6\%} \\
Mouse & 96hr & 0.435 & 0.376 & \cellcolorgrad{14}\textcolor{black}{+13.6\%} & 0.439 & \cellcolorgrad{-1}\textcolor{black}{-1.0\%} \\
Peacock & 24hr & 0.302 & 0.281 & \cellcolorgrad{7}\textcolor{black}{+7.0\%} & 0.290 & \cellcolorgrad{4}\textcolor{black}{+3.9\%} \\
Peacock & 96hr & 0.375 & 0.376 & \cellcolorgrad{0}\textcolor{black}{-0.2\%} & 0.366 & \cellcolorgrad{2}\textcolor{black}{+2.4\%} \\
Rat & 24hr & 0.275 & 0.281 & \cellcolorgrad{-2}\textcolor{black}{-2.2\%} & 0.266 & \cellcolorgrad{3}\textcolor{black}{+3.5\%} \\
Rat & 96hr & 0.378 & 0.376 & \cellcolorgrad{1}\textcolor{black}{+0.8\%} & 0.371 & \cellcolorgrad{2}\textcolor{black}{+2.0\%} \\
Robin & 24hr & 0.247 & 0.281 & \cellcolorgrad{-14}\textcolor{black}{-13.9\%} & 0.235 & \cellcolorgrad{5}\textcolor{black}{+4.6\%} \\
Robin & 96hr & 0.333 & 0.376 & \cellcolorgrad{-13}\textcolor{black}{-12.9\%} & 0.320 & \cellcolorgrad{4}\textcolor{black}{+3.8\%} \\
Wolf & 24hr & 0.396 & 0.281 & \cellcolorgrad{29}\textcolor{black}{+29.0\%} & 0.371 & \cellcolorgrad{6}\textcolor{black}{+6.3\%} \\
Wolf & 96hr & 0.495 & 0.376 & \cellcolorgrad{24}\textcolor{black}{+24.1\%} & 0.489 & \cellcolorgrad{1}\textcolor{black}{+1.1\%} \\
\bottomrule
\end{tabular}
\end{adjustbox}
\end{table}

\begin{table}[H]
\caption{MSE Performance Summary for Large-Scale Modelling on Individual Datasets\\ \textit{(PatchTST)}}
\vspace{2mm}
\centering
\begin{adjustbox}{width=\textwidth,center}
\tiny
\label{tab:large_scale_summary_patchtst_mse}
\begin{tabular}{llrrrrrr}
\toprule
Dataset & Horizon & Baseline & Strategy 6 & Imp & Strategy 8 & Imp \\
 & & MSE & MSE & (\%) & MSE & (\%) \\
\midrule
Bear & 24hr & 0.201 & 0.203 & \cellcolorgrad{-1}\textcolor{black}{-1.4\%} & 0.191 & \cellcolorgrad{5}\textcolor{black}{+4.7\%} \\
Bear & 96hr & 0.321 & 0.327 & \cellcolorgrad{-2}\textcolor{black}{-2.0\%} & 0.307 & \cellcolorgrad{4}\textcolor{black}{+4.4\%} \\
Bobcat & 24hr & 0.305 & 0.203 & \cellcolorgrad{33}\textcolor{black}{+33.2\%} & 0.284 & \cellcolorgrad{7}\textcolor{black}{+6.9\%} \\
Bobcat & 96hr & 0.406 & 0.327 & \cellcolorgrad{19}\textcolor{black}{+19.3\%} & 0.402 & \cellcolorgrad{1}\textcolor{black}{+1.1\%} \\
Bull & 24hr & 0.407 & 0.203 & \cellcolorgrad{50}\textcolor{black}{+49.9\%} & 0.396 & \cellcolorgrad{3}\textcolor{black}{+2.5\%} \\
Bull & 96hr & 0.582 & 0.327 & \cellcolorgrad{44}\textcolor{black}{+43.7\%} & 0.589 & \cellcolorgrad{-1}\textcolor{black}{-1.3\%} \\
Cockatoo & 24hr & 0.351 & 0.203 & \cellcolorgrad{42}\textcolor{black}{+42.0\%} & 0.335 & \cellcolorgrad{5}\textcolor{black}{+4.6\%} \\
Cockatoo & 96hr & 0.536 & 0.327 & \cellcolorgrad{39}\textcolor{black}{+38.9\%} & 0.539 & \cellcolorgrad{-1}\textcolor{black}{-0.7\%} \\
Crow & 24hr & 0.249 & 0.203 & \cellcolorgrad{18}\textcolor{black}{+18.1\%} & 0.226 & \cellcolorgrad{9}\textcolor{black}{+9.2\%} \\
Crow & 96hr & 0.404 & 0.327 & \cellcolorgrad{19}\textcolor{black}{+19.0\%} & 0.413 & \cellcolorgrad{-2}\textcolor{black}{-2.1\%} \\
Eagle & 24hr & 0.226 & 0.203 & \cellcolorgrad{10}\textcolor{black}{+9.9\%} & 0.219 & \cellcolorgrad{3}\textcolor{black}{+3.1\%} \\
Eagle & 96hr & 0.322 & 0.327 & \cellcolorgrad{-2}\textcolor{black}{-1.6\%} & 0.314 & \cellcolorgrad{2}\textcolor{black}{+2.5\%} \\
Fox & 24hr & 0.169 & 0.203 & \cellcolorgrad{-20}\textcolor{black}{-20.1\%} & 0.164 & \cellcolorgrad{3}\textcolor{black}{+3.4\%} \\
Fox & 96hr & 0.249 & 0.327 & \cellcolorgrad{-31}\textcolor{black}{-31.4\%} & 0.243 & \cellcolorgrad{2}\textcolor{black}{+2.5\%} \\
Gator & 24hr & 0.274 & 0.203 & \cellcolorgrad{26}\textcolor{black}{+25.8\%} & 0.275 & \cellcolorgrad{0}\textcolor{black}{-0.3\%} \\
Gator & 96hr & 0.547 & 0.327 & \cellcolorgrad{40}\textcolor{black}{+40.1\%} & 0.543 & \cellcolorgrad{1}\textcolor{black}{+0.7\%} \\
Hog & 24hr & 0.205 & 0.203 & \cellcolorgrad{1}\textcolor{black}{+0.7\%} & 0.190 & \cellcolorgrad{7}\textcolor{black}{+7.0\%} \\
Hog & 96hr & 0.350 & 0.327 & \cellcolorgrad{7}\textcolor{black}{+6.6\%} & 0.336 & \cellcolorgrad{4}\textcolor{black}{+4.1\%} \\
Lamb & 24hr & 0.143 & 0.203 & \cellcolorgrad{-42}\textcolor{black}{-41.9\%} & 0.137 & \cellcolorgrad{4}\textcolor{black}{+4.5\%} \\
Lamb & 96hr & 0.219 & 0.327 & \cellcolorgrad{-50}\textcolor{black}{-49.8\%} & 0.214 & \cellcolorgrad{2}\textcolor{black}{+2.3\%} \\
Moose & 24hr & 0.188 & 0.203 & \cellcolorgrad{-8}\textcolor{black}{-8.3\%} & 0.167 & \cellcolorgrad{11}\textcolor{black}{+11.3\%} \\
Moose & 96hr & 0.293 & 0.327 & \cellcolorgrad{-12}\textcolor{black}{-11.7\%} & 0.289 & \cellcolorgrad{1}\textcolor{black}{+1.5\%} \\
Mouse & 24hr & 0.272 & 0.203 & \cellcolorgrad{25}\textcolor{black}{+25.2\%} & 0.257 & \cellcolorgrad{5}\textcolor{black}{+5.3\%} \\
Mouse & 96hr & 0.427 & 0.327 & \cellcolorgrad{23}\textcolor{black}{+23.4\%} & 0.457 & \cellcolorgrad{-7}\textcolor{black}{-7.0\%} \\
Peacock & 24hr & 0.189 & 0.203 & \cellcolorgrad{-8}\textcolor{black}{-7.8\%} & 0.179 & \cellcolorgrad{5}\textcolor{black}{+5.3\%} \\
Peacock & 96hr & 0.283 & 0.327 & \cellcolorgrad{-16}\textcolor{black}{-15.9\%} & 0.273 & \cellcolorgrad{3}\textcolor{black}{+3.3\%} \\
Rat & 24hr & 0.183 & 0.203 & \cellcolorgrad{-11}\textcolor{black}{-11.3\%} & 0.176 & \cellcolorgrad{4}\textcolor{black}{+4.0\%} \\
Rat & 96hr & 0.316 & 0.327 & \cellcolorgrad{-3}\textcolor{black}{-3.5\%} & 0.308 & \cellcolorgrad{3}\textcolor{black}{+2.6\%} \\
Robin & 24hr & 0.150 & 0.203 & \cellcolorgrad{-36}\textcolor{black}{-36.0\%} & 0.140 & \cellcolorgrad{6}\textcolor{black}{+6.4\%} \\
Robin & 96hr & 0.251 & 0.327 & \cellcolorgrad{-30}\textcolor{black}{-30.4\%} & 0.240 & \cellcolorgrad{5}\textcolor{black}{+4.6\%} \\
Wolf & 24hr & 0.342 & 0.203 & \cellcolorgrad{41}\textcolor{black}{+40.6\%} & 0.326 & \cellcolorgrad{5}\textcolor{black}{+4.8\%} \\
Wolf & 96hr & 0.494 & 0.327 & \cellcolorgrad{34}\textcolor{black}{+33.7\%} & 0.492 & \cellcolorgrad{0}\textcolor{black}{+0.4\%} \\
\bottomrule
\end{tabular}
\end{adjustbox}
\end{table}
\section{Graphical Representation of MAE Results for Strategies 6 and 8}
\begin{figure}[H]
    \centering
    \includegraphics[width=1\linewidth]{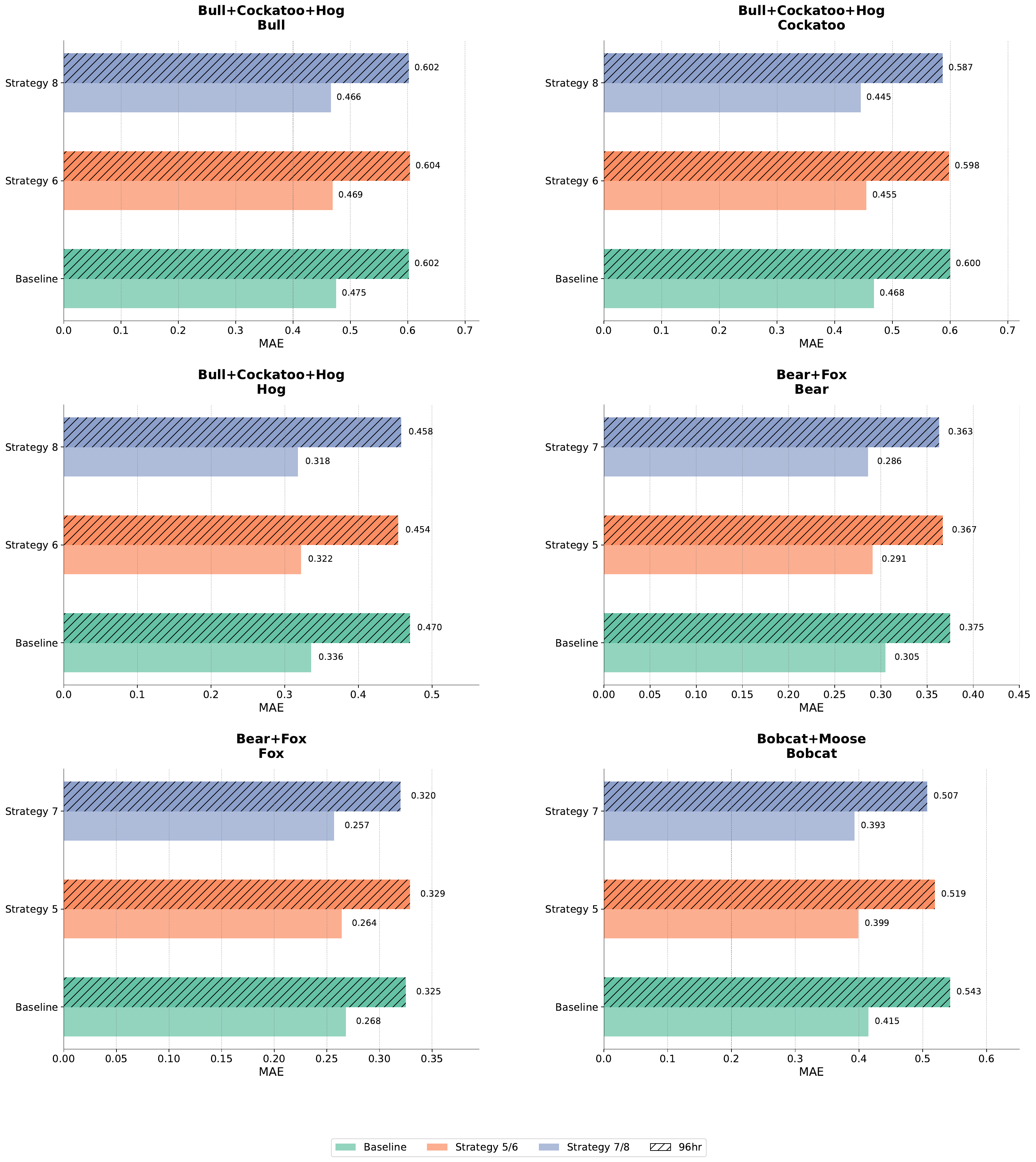}
    \caption{Mean Absolute Error (MAE) Comparison for Small-Scale Transfer Learning Scenarios (Part 1). This figure shows the performance of base, combined, and fine-tuned models for the first 20 dataset combinations and test scenarios.}
    \label{fig:enter-label}
\end{figure}

\begin{figure}[H]
    \centering
    \includegraphics[width=1\linewidth]{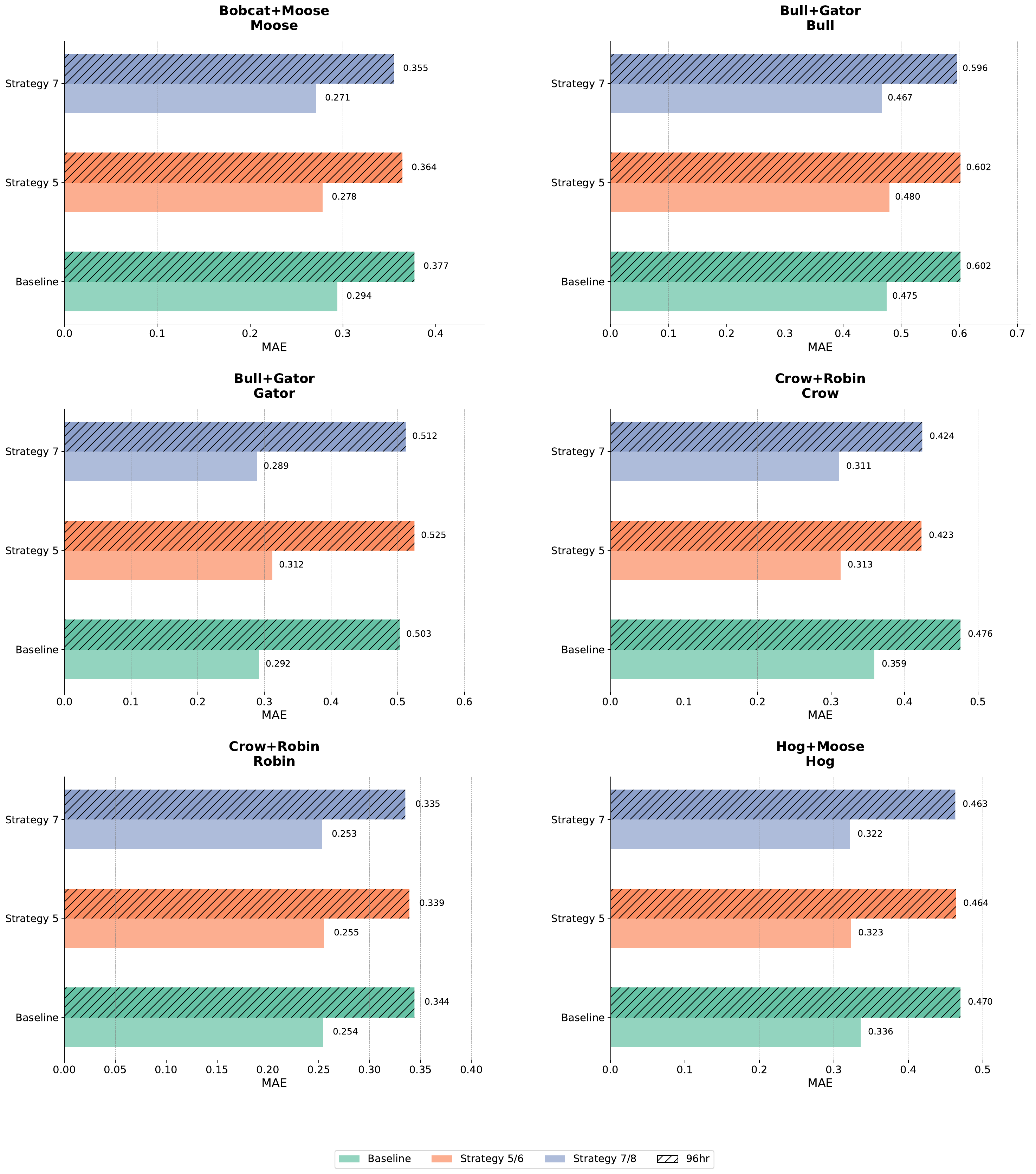}
    \caption{Mean Absolute Error (MAE) Comparison for Small-Scale Transfer Learning Scenarios (Part 2). This figure presents the performance of base, combined, and fine-tuned models for the remaining 14 dataset combinations and test scenarios.}
    \label{fig:enter-label}
\end{figure}

\begin{figure}[H]
    \centering
    \includegraphics[width=1\linewidth]{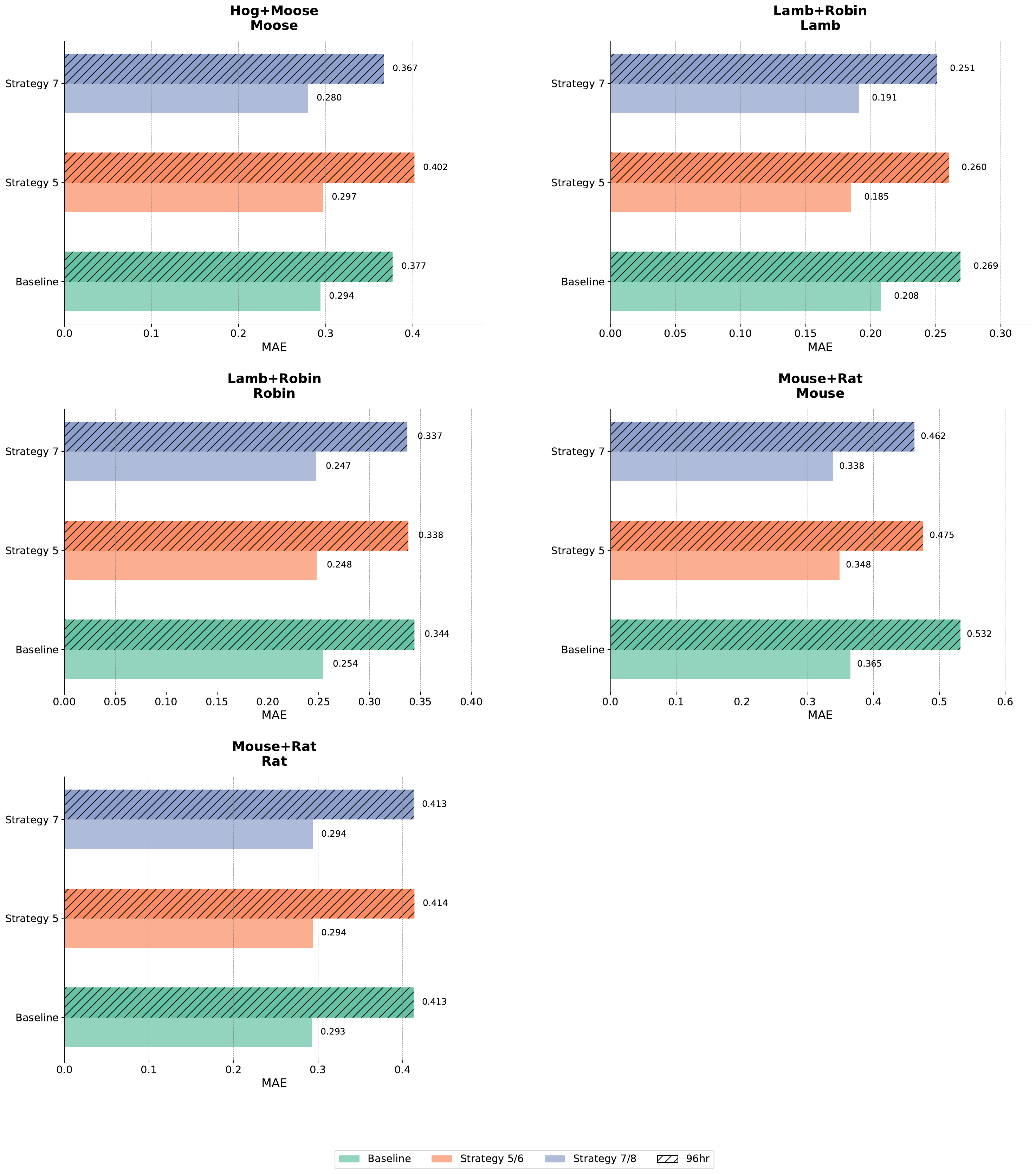}
    \caption{Mean Absolute Error (MAE) Comparison for Small-Scale Transfer Learning Scenarios (Part 3). This figure presents the performance of base, combined, and fine-tuned models for the remaining 14 dataset combinations and test scenarios.}
    \label{fig:enter-label}
\end{figure}
\iffalse
\begin{figure}[H]
    \centering
    \includegraphics[width=1\linewidth]{large_scale_comparison_figure2.pdf}
    \caption{Mean Absolute Error (MAE) Comparison for Large-Scale Modelling Across Different Transformer Architectures (Part 2). This figure compares the performance of Vanilla Transformer, Informer, and PatchTST models for the remaining 6 datasets, showing base, zero-shot, and fine-tuned results for both 24-hour and 96-hour forecasting horizons.}
    \label{fig:enter-label}
\end{figure}

\begin{figure}[H]
    \centering
    \includegraphics[width=1\linewidth]{large_scale_comparison_figure3.pdf}
    \caption{Mean Absolute Error (MAE) Comparison for Large-Scale Modelling Across Different Transformer Architectures (Part 3). This figure compares the performance of Vanilla Transformer, Informer, and PatchTST models for the remaining 6 datasets, showing base, zero-shot, and fine-tuned results for both 24-hour and 96-hour forecasting horizons.}
    \label{fig:enter-label}
\end{figure}
\fi
\restoregeometry

\end{document}